\documentclass[10pt,twocolumn,letterpaper]{article}

\usepackage[pagenumbers]{cvpr} 

\definecolor{cvprblue}{rgb}{0.21,0.49,0.74}
\usepackage[pagebackref,breaklinks,colorlinks,allcolors=cvprblue]{hyperref}
\usepackage[utf8]{inputenc} 
\usepackage[T1]{fontenc}    
\usepackage{hyperref}       
\usepackage{url}            
\usepackage{booktabs}       
\usepackage{amsfonts}       
\usepackage{nicefrac}       
\usepackage{microtype}      
\usepackage{xcolor}
\usepackage{graphicx}
\usepackage{svg}
\usepackage{wrapfig}
\usepackage{subcaption}
\usepackage{multirow}
\usepackage{makecell}
\usepackage{amssymb}
\usepackage{amsmath}
\usepackage{placeins}
\usepackage{listings}
\usepackage{bbm}
\usepackage{rotating}
\usepackage{pifont}       
\usepackage{array}        
\usepackage{bm}       
\usepackage{tabularx}  

\let\oldaddcontentsline\addcontentsline  
\renewcommand{\addcontentsline}[3]{}    

\definecolor{darkorange}{RGB}{242,140,29}
\definecolor{lightorange}{RGB}{255,211,159}
\definecolor{darkgreen}{RGB}{91,196,166}
\definecolor{lightgreen}{RGB}{170,229,214}

\newcommand{\cbest}{\cellcolor{darkgreen!70}}
\newcommand{\cbetter}{\cellcolor{lightgreen!50}}
\newcommand{\csame}{\cellcolor{white}}
\newcommand{\cworse}{\cellcolor{lightorange!50}}
\newcommand{\cworst}{\cellcolor{darkorange!50}}

\renewcommand\cbest{}
\renewcommand\cbetter{}
\renewcommand\csame{}
\renewcommand\cworse{}
\renewcommand\cworst{}

\newcommand{\subfiglabel}[1]{%
  \refstepcounter{subfigure}%
  \label{#1}%
}

\newcommand{\scaledmath}[2][0.9]{
  \scalebox{#1}{$\smash{#2}$}%
}

\newcolumntype{L}[1]{>{\raggedright\arraybackslash}m{#1}}  
\newcolumntype{C}[1]{>{\centering\arraybackslash}m{#1}}    
\newcolumntype{R}[1]{>{\raggedleft\arraybackslash}m{#1}}   

\newcommand{\Yes}{\ding{51}}  
\newcommand{\No}{\ding{55}}   
\newcommand{\yes}{\textcolor{gray!50}{\ding{51}}}  
\newcommand{\no}{\textcolor{gray!50}{\ding{55}}}   
\newcommand{\BlackNum}[1]{\textbf{\textit{\textsf{#1}}}}
\newcommand{\GrayNum}[1]{\textcolor{gray!50}{\textbf{\textit{\textsf{#1}}}}}
\newcommand{\rhead}[1]{\rotatebox{90}{#1}} 

\newcommand{\CNN}{\textcolor{black}{\ding{104}} }         
\newcommand{\Transformer}{\textcolor{red}{\ding{169}}}  
\newcommand{\GNN}{\textcolor{blue}{\ding{72}}}       
\newcommand{\HGNN}{\textcolor{orange}{\ding{110}}}    
\newcommand{\HGVT}{\textcolor{green!60!black}{\ding{115}}}    

\newcommand{\adj}[0]{\mathrm{adj}}

\newcommand{\V}[0]{{(V)}}
\newcommand{\E}[0]{{(E)}}

\newcommand{\sublabel}[1]{\noindent\textbf{#1}}

\def\paperID{12021} 
\def\confName{CVPR}
\def\confYear{2025}

\title{Hypergraph Vision Transformers: Images are More than Nodes, More than Edges}

\author{Joshua Fixelle\\
University of Virginia\\
{\tt\small jf9fk@virginia.edu}
}

\begin{document}
\maketitle



\begin{abstract}
Recent advancements in computer vision have highlighted the scalability of Vision Transformers (ViTs) across various tasks, yet challenges remain in balancing adaptability, computational efficiency, and the ability to model higher-order relationships. Vision Graph Neural Networks (ViGs) offer an alternative by leveraging graph-based methodologies but are hindered by the computational bottlenecks of clustering algorithms used for edge generation. To address these issues, we propose the \textbf{H}yper\textbf{g}raph \textbf{V}ision \textbf{T}ransformer (\textbf{HgVT}), which incorporates a hierarchical bipartite hypergraph structure into the vision transformer framework to capture higher-order semantic relationships while maintaining computational efficiency. HgVT leverages population and diversity regularization for dynamic hypergraph construction without clustering, and expert edge pooling to enhance semantic extraction and facilitate graph-based image retrieval. Empirical results demonstrate that HgVT achieves strong performance on image classification and retrieval, positioning it as an efficient framework for semantic-based vision tasks.
\\
\end{abstract}

\section{Introduction}

\begin{figure*}[ht]
  \centering
  \includegraphics[width=0.75\textwidth]{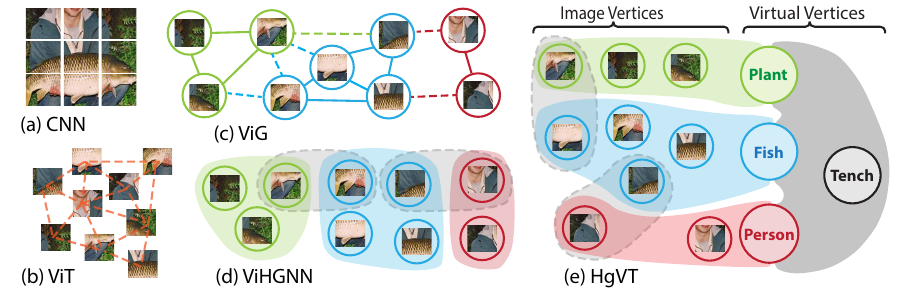}
  \vspace{-0.5em}
  \caption{
  Comparison of Graph Structures for different methods. Showing (a) CNNs, (b) Vision Transformers, (c) ViG with a KNN clustered GNN, (d) ViHGNN with clustered hyperedges, and (e) our proposed HgVT method. Strong group edges shown with solid lines; weak edges with dashed lines. Hyperedges shown with shaded regions; less dominant hyperedges with gray dashed regions.
  }
  \label{fig:hyper_tree}
  \subfiglabel{fig:hyper_tree_cnn}
  \subfiglabel{fig:hyper_tree_vit}
  \subfiglabel{fig:hyper_tree_vig}
  \subfiglabel{fig:hyper_tree_vihgnn}
  \subfiglabel{fig:hyper_tree_hgvt}
  \vspace{-1em}
\end{figure*}


Computer vision has recently transitioned from the historically dominant Convolutional Neural Networks (CNNs)~\cite{lenet_paper, alexnet_paper, resnet_paper} to the increasingly prominent Vision Transformers (ViTs), which have quickly embedded themselves as the new de facto standard~\cite{vit16x16words, oquab2024dinov2}. This shift reflects the broader success of transformers in natural language processing~\cite{attention_paper, bert_paper, touvron2023llama} and is driven by the remarkable scalability of ViTs across various tasks such as image classification~\cite{zhai2022scalingvisiontransformers, deit_paper}, semantic segmentation~\cite{segvit_paper, sam_paper}, and image retrieval~\cite{dinov1_paper, Kusupati2022MatryoshkaRL}. While hybrid models like hierarchical attention and CNN-ViT methods~\cite{nattentransformer, swintransformer, hybrid_cnn_vit_survey} have emerged to balance computational load and flexibility, challenges remain, particularly with ViTs focusing on salient features rather than comprehensive image understanding~\cite{vit16x16words, dinov1_paper, Demidov2023SalientMV, puigcerver2023sparse}. This underscores the ongoing need for approaches that enhance computational efficiency alongside semantic accuracy.

Within the spectrum of novel architectures, Vision Graph Neural Networks (ViGs) \cite{vig_paper} and Vision Hypergraph Neural Networks (ViHGNNs) \cite{vihgnn_paper} leverage graph-based topologies to advance image processing. Unlike CNNs, which harness locality and translation-invariance through densely connected pixel grids and repeated convolutions, both ViTs and ViGs represent images as sets of patches. In ViTs, each patch acts as a vertex within a maximally connected graph, creating semantically weak connections through self-attention. ViGs enhance this by using clustering algorithms to detect edge groupings and applying graph convolutions to these clusters, forming meaningful patch relationships. ViHGNNs extend these capabilities by employing hyperedges that capture complex, higher-order relationships, enriching understanding of the images. These methodologies are depicted in Figure~\ref{fig:hyper_tree}.

While graph-based models like ViG and ViHGNN have introduced notable advancements in visual perception, two critical observations emerge regarding these architectures:
\begin{enumerate}
    \item In existing vision GNN models~\cite{vig_paper, vihgnn_paper, mobilevig2023, GreedyViG_2024_CVPR}, edge features are primarily used for basic vertex-to-vertex communication and are not integrated across successive layers:~a strategy that could enhance cumulative learning and improve classification accuracy.
    \item The computational complexities associated with clustering algorithms used for edge generation, such as KNN in ViG and Fuzzy C-Means in ViHGNN, pose significant computational bottlenecks. Approaches like MobileViG~\cite{mobilevig2023} and GreedyViG~\cite{GreedyViG_2024_CVPR} attempt to mitigate these challenges with static graph structures and adding dynamic masking, but do so by trading adaptability for efficiency, failing to achieve a well-balanced solution.
\end{enumerate}
In response to limitations in existing graph-based models, we propose the Hypergraph Vision Transformer (HgVT), which advances the hypergraph concept with a bipartite representation where hyperedge features and image patches (vertices) are continuously processed. Unlike traditional models that use graph convolutions, HgVT employs structured multi-head attention for efficient vertex-hyperedge message passing and incorporates a dynamic querying mechanism that constructs graph structures in $\mathcal{O}(|V|\cdot E)$ time complexity, where 
$E < |V|$. This graph structure is then utilized in attention masking to balance structural adaptability with computational efficiency. 
Furthermore, HgVT integrates virtual elements into vertices and hyperedges to enable restricted message passing via attention masking, facilitating a hierarchical semantic structure that leverages virtual hyperedge features for classification, as illustrated in \cref{fig:hyper_tree_hgvt}. Our contributions are thereby summarized as follows:

\begin{itemize}
\item We propose the Hypergraph Vision Transformer (HgVT), which integrates a hierarchical bipartite hypergraph structure within a vision transformer framework. Our isotropic HgVT-Ti model achieves a top-1 accuracy of 76.2\% on the ImageNet-1k classification task, surpassing the prior state-of-the-art by 1.9\%, demonstrating the efficacy of hypergraph-based learning within vision transformers.
\item We introduce population and diversity regularization strategies that enable dynamic hypergraph structure construction in HgVT, allowing the model to self-sparsify connections without relying on traditional clustering techniques. 
\item We implement expert edge pooling, a pooling approach that selects edges based on learned confidence scores, facilitating efficient representation pruning and graph extraction. This approach demonstrates strong semantic clustering behavior across macro-classes and achieves competitive image retrieval performance compared to other feature extractors, while maintaining a more compact model size.
\end{itemize}
\section{Related Work}

\sublabel{Vision Transformers.} Vision Transformers (ViTs) proposed by~\cite{vit16x16words} and refined by~\cite{oquab2024dinov2, deit_paper, dinov1_paper} use self-attention to process image patches as sequences, scaling to complex datasets and tasks. Recent ViTs have reintroduced spatial hierarchies by leveraging local attention~\cite{swintransformer, nattentransformer}, integrating sparse global summaries~\cite{zhu2023biformer}, and employing biomimetic modeling to focus on key regions within images~\cite{shi2023transnext}. However, current models tend to focus on the most salient objects and patch-level similarities, ignoring global structure. HgVT addresses this by introducing bipartite hypergraphs to model higher-order relationships for improved semantic understanding.
\vspace{0.5em}
\\
\sublabel{Graph-Based Vision Models and Clustering.} Graph Neural Networks~(GNNs), initially conceptualized by~\cite{gnn_paper}, have been applied to vision tasks through Vision Graph Neural Networks (ViGs)~\cite{vig_paper}, which exhibit improved accuracy over ViTs on common vision tasks. ViGs use graph convolutions to model image patch relationships on a graph structure, typically constructed by iterative clustering algorithms such as KNN and Fuzzy C-Means, which introduce computational overhead. Recent methods avoid clustering inefficiencies with static graph structures~\cite{mobilevig2023, GreedyViG_2024_CVPR}, sacrificing adaptability. HgVT instead introduces a dynamic graph construction method, relying on cosine similarity from learned features to enable efficient, non-iterative, adaptive clustering.
%
%
\vspace{0.5em}
\\
\sublabel{Hypergraph-Based Methods.} While previously used in many computer vision tasks~\cite{video_hyper_cuts_paper, 3dview_hyper_classification_paper, hyper_image_retrieval_paper}, hypergraphs have recently been incorporated into vision GNNs~\cite{vihgnn_paper, micrograph_hgnn}, improving their ability to model complex multi-way relationships. However, these methods treat hypergraphs as an intermediate tool rather than producing a hypergraph to represent underlying images, preventing their use in downstream tasks. HgVT instead iteratively refines a hypergraph through subsequent network layers to produce structured representations.

\section{Hierarchical Hypergraphs}
\sublabel{Graphs and Basic Notations.} Graphs are powerful mathematical tools for representing structured information, applicable across diverse disciplines.
A graph $\mathcal{G}$ is defined as a pair $\left(\mathcal{V},\mathcal{E}\right)$, where $\mathcal{V}=\{v_1, v_2, \dots, v_N\}$ is a set of vertices, and $\mathcal{E}=\{e_{ij}|(v_i,v_j)\}$ is a set of edges, for directed graphs, or with $e_{ij}=e_{ji}$ for undirected graphs. Each edge~$e_{ij}$ connects a pair of vertices~$v_i$~and~$v_j$, where~$v_i,v_j\in \mathcal{V}$. The adjacency matrix \scaledmath{\hat{\mathbf{A}}} is a binary matrix~\scaledmath{\{0,1\}^{|V|\times|V|}}, representing the presence~(1) or absence~(0) of an edge between each pair of vertices. Similarly, an edge weight matrix can be defined as \scaledmath{\mathbf{A}\in\mathbb{R}^{|V|\times|V|}} to quantify the strength or capacity of these connections.
\\

\vspace{-0.5em}
\sublabel{Graph Convolution Networks (GCNs).} Building on this foundation, GCNs utilize vertex feature matrices~\scaledmath{\mathbf{X}\in\mathbb{R}^{|V|\times d}} to encode vertex properties. Their core mechanism, message passing, updates vertex features through a convolution with a learned projection matrix \scaledmath{\mathbf{W}\in\mathbb{R}^{d\times d}} and a non-linear activation, guided by the adjacency matrix \scaledmath{\hat{\mathbf{A}}}, which specifies neighboring vertices Adjacency features~\scaledmath{\mathbf{X}_\mathrm{adj}\in\mathbb{R}^{|V|\times d_a}}, typically set as \scaledmath{\mathbf{X}_\mathrm{adj} = \mathbf{X}}, enable dynamic updates to \scaledmath{\hat{\mathbf{A}}}, and the edge weight matrix $\mathbf{A}$, allowing the graph structure to evolve based on learned interactions. However, GCNs are inherently limited by the pairwise edges in $\mathcal{E}$, unable to capture multi-vertex relationships.

\subsection{Hypergraphs and Bipartite Representations}

\begin{figure}[h]
  \centering
  \vspace{-1em}
  \includegraphics[width=0.4\textwidth]{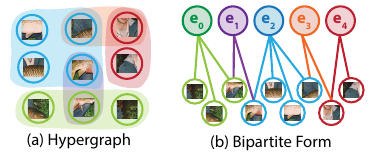}
  \vspace{-0.5em}
  \caption{Comparison of (a) hypergraph and (b) equivalent bipartite representation from \cref{fig:hyper_tree_vihgnn}, showing five hyperedges.}
  \vspace{-0.5em}
  \label{fig:bipartite}
\end{figure}

To overcome the pairwise limitation inherent in traditional graphs, hypergraphs offer a robust solution by extending the concept of edges to hyperedges, which connect multiple vertices simultaneously.
In a hypergraph~$\mathcal{H}=(\mathcal{V},\mathcal{E})$, hyperedges~$e_j \in \mathcal{E}$ each connect a subset of vertices, defined as~$e_j = \{ v_i \;| \; v_i\!\in\!\mathcal{V} \; \mathrm{and} \; i\!\in\!I_j\}$, where $I_j$ is the set of indices for vertices that are included in hyperedge~$e_j$.
The set $I_j$ directly corresponds to the nonzero entries of the $j$-th column of the incidence matrix~$\mathbf{H} \in \{0,1\}^{|V|\times|E|}$, where~$\mathbf{H}_{ij} = 1$ if vertex~$v_i$ is included in the hyperedge~$e_j$. 
This structure effectively captures complex inter-vertex relationships, making hypergraphs especially valuable in applications that require a deep understanding of networked systems or grouped interactions.

Hypergraphs can alternatively be described using a bipartite representation, where the vertex set~$\mathcal{V}$ and hyperedge set~$\mathcal{E}$ form distinct groups linked by the incidence matrix~$\mathbf{H}$~(refer to \cref{fig:hyper_tree_vihgnn}). 
This representation results in a new graph~\mbox{$\mathcal{G}_B=(\mathcal{V}, \mathcal{E}, \mathcal{E}_B)$}, where $\mathcal{V}$ represents the original vertices of the hypergraph, and the elements in $\mathcal{E}$ correspond to hyperedges. The edges in $\mathcal{E}_B$, denoted as~$\epsilon_{ve}=(\nu_v,\nu_e)$ exist if~$\mathbf{H}_{ve}=1$, with $\nu_v \in \mathcal{V}$ and  $\nu_e\in \mathcal{E}$, linking $\mathcal{V}$ and $\mathcal{E}$.

In the bipartite graph $\mathcal{G}_B$, the corresponding adjacency matrix can be simplified as \scaledmath{\hat{\mathbf{A}} = \mathbf{H}} for $\mathcal{E}\rightarrow\mathcal{V}$ interactions, and \scaledmath{\hat{\mathbf{A}}=\mathbf{H}^T} for $\mathcal{V}\rightarrow\mathcal{E}$. 
Drawing on principles similar to those in ViHGNN \cite{vihgnn_paper}, the edge weight matrix $\mathbf{A}$ can be interpreted as fuzzy membership weights, enabling graded interactions and supporting various communication strategies across GNN layers. 
Complementing this setup, the feature matrices are split into~$\mathbf{X}^\V\in\mathbb{R}^{|V|\times d_v}$ and~$\smash{\mathbf{X}^\E\in\mathbb{R}^{|E|\times d_e}}$, along with their correspond adjacency feature matrices~\scaledmath{\mathbf{X}^\V_\mathrm{adj}} and~\scaledmath{\mathbf{X}^\E_\mathrm{adj}}, mirroring traditional GNNs.




\subsection{Imposing Hierarchical Structure in Images}
\label{sec:imposing_hierarchical_structure}

To enhance the capability of hypergraphs in image analysis, we draw inspiration from the register tokens introduced in~\cite{darcet2023vision}, which act to summarize information that otherwise manifests as noise in areas of low visual significance.
Similarly, this work integrates virtual vertices~($v\mathcal{V}$), alongside typical image patch vertices~($i\mathcal{V}$), and introduces virtual hyperedges~($v\mathcal{E}$),  alongside primary hyperedges~($p\mathcal{E}$), to provide layers of semantic feature aggregation and relational abstraction. 
These virtual elements, illustrated in Figure \ref{fig:hyper_tree_hgvt}, do not correspond to specific image patches; instead, they are learned embeddings used for semantic summarization and capturing high-level abstract information.

Our proposed hypergraph, constructed from image~$I$ as $\mathcal{G}_B(I)$, integrates primary and virtual sets, forming~\mbox{$\mathcal{V}=i\mathcal{V}\cup v\mathcal{V}$} and~\mbox{$\mathcal{E}=p\mathcal{E}\cup v\mathcal{E}$}, with statically masked communication pathways to enforce a hierarchical structure.
Primary~hyperedges~($p\mathcal{E}$) interact with all vertices to support unrestricted semantic aggregation, whereas virtual~hyperedges~($v\mathcal{E}$), designated for class predictions, connect solely with virtual~vertices~($v\mathcal{V}$).
These restrictions separate visual and abstract information, thereby producing a graph structure suitable for use in downstream applications.



\vspace{-0.5em}
\section{A Hypergraph Vision Transformer}

\begin{figure*}[ht!]
  \centering
  \includegraphics[width=0.85\textwidth]{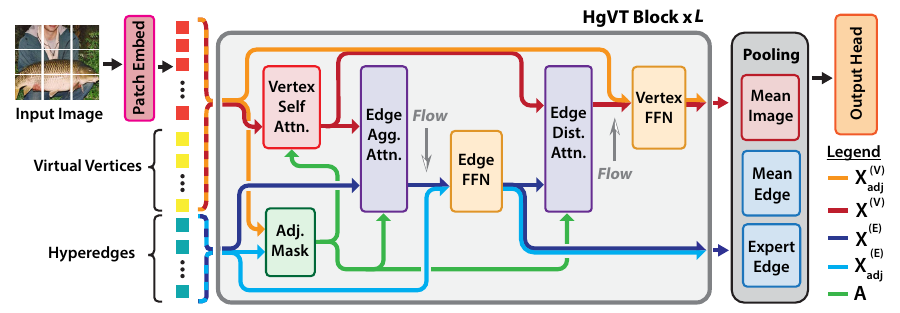}
  \caption{HgVT Architecture Diagram, composed of stacked HgVT blocks with adjacency matrix $\mathbf{A}$, vertex features \scaledmath{\mathbf{X}^\V}, and hyperedge features $\mathbf{X}^\E$. Edge attention flow is shown with gray arrows; input norms and residual adds are omitted for clarity. }
  \label{fig:hgvt_arch}
  \vspace{-1em}
\end{figure*}

The Hypergraph Vision Transformer (HgVT) adapts the architecture of standard Vision Transformers by incorporating bipartite hypergraph features for enhanced image analysis capabilities. 
Like Vision Transformers, HgVT begins with a patch embedding layer, followed by an isotropic stack of \mbox{$L\times$ HgVT} blocks, culminating in feature pooling and a classifier head. 
The bipartite hypergraph is represented by four principal feature matrices -- \scaledmath{\mathbf{X}^\V},~\scaledmath{\mathbf{X}^\V_\adj},~\scaledmath{\mathbf{X}^\E},~and~\scaledmath{\mathbf{X}^\E_\adj} -- which are updated iteratively and in an interleaved fashion within each block. 
Each block constructs a new adjacency matrix~$\mathbf{A}$ from~\scaledmath{\mathbf{X}^\V_\adj} and~\scaledmath{\mathbf{X}^\E_\adj}, enabling flexible adjustments to the hypergraph structure.
As illustrated in \cref{fig:hgvt_arch}, this modular process allows for the continuous integration and processing of these matrices within each HgVT block.


\subsection{Hyperedges as Communication Pools}

Each HgVT block processes both vertex and edge information, refining them from the previous block based on a newly constructed graph structure.
Initially, adjacency mask computation (detailed in the next section) determines the connectivity for the subsequent processing steps within each block, dynamically adjusting to the updated feature matrices from the previous block.
Three attention layers -- vertex self-attention, edge aggregate attention, and edge distribution attention -- operate sequentially to enhance feature integration and facilitate effective message passing along the hyperedges formed in the adjacency computation step. 
Finally, separate feed-forward networks process vertex and edge features independently, ensuring specialized treatment for the two distinct sets within the bipartite hypergraph, preserving the unique properties of each set.
Operational details of these components are further described in \cref{app:app_arch_details}.



\vspace{0.5em}
\sublabel{Hypergraph Feature Processing.} Within each HgVT block, two distinct point-wise feed-forward networks (FFNs) independently process vertex and hyperedge features, aligning with the bipartite structure of the hypergraph. Each FFN integrates both the element features and their corresponding adjacency features through a fully connected layer, improving the model’s ability to synthesize relationships. Processing both feature types within the same FFN layer allows adjacency information to be handled directly, bypassing the need for graph-based message passing and improving computational efficiency. Moreover, parameter overhead can be reduced by optionally tying edge and vertex FFN weights.



\vspace{0.5em}

\begin{figure}[h] 
   \vspace{1.5em}
    \centering
    \begin{subfigure}[b]{0.22\textwidth}
        \includegraphics[width=\textwidth]{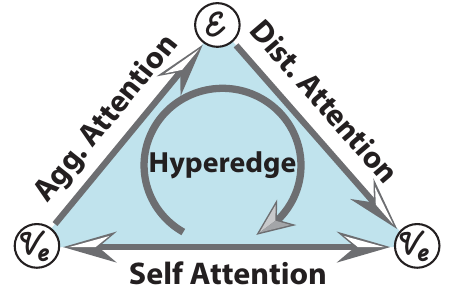}
        \caption{Communication Pool.}
        \label{fig:comm_pool}
    \end{subfigure}
    \hfill 
    \begin{subfigure}[b]{0.23\textwidth}
        \includegraphics[width=\textwidth]{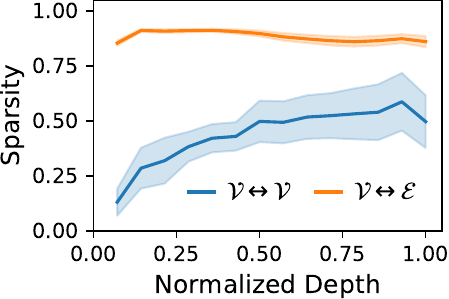}
        \caption{Attention Sparsity.}
        \label{fig:sparse_depth}
    \end{subfigure}
    \vspace{-0.5em}
    \caption{(a) Hyperedge Communication Pool Flow with edges~$\mathcal{E}$ and member vertices~$\mathcal{V}_e$; (b) Attention Sparsity (Mean and std) for HgVT-S on the ImageNet-1k Validation set.}
  \vspace{-1.5em}
  \label{fig:comm_and_sparse}
\end{figure}

\sublabel{Hyperedges as Communication Pools.} Hypergraph GNNs typically employ a gather$\rightarrow$scatter mechanism for processing vertex-hyperedge interactions, whereas HgVT reconceptualizes hyperedges as communication pools that facilitate information flow among vertices and their associated hyperedges. 
Specifically, vertex self-attention manages vertex-to-vertex~($\mathcal{V}\!\rightarrow\!\mathcal{V}$) interactions within hyperedges, edge aggregate attention orchestrates the flow from vertices to hyperedges~($\mathcal{V}\!\rightarrow\!\mathcal{E}$), and edge distribution attention handles the reverse, from hyperedges back to vertices ($\mathcal{E}\!\rightarrow\!\mathcal{V}$). By segmenting the attention operations, HgVT efficiently approximates an all-to-all feature transfer within hyperedges, as illustrated in \cref{fig:comm_pool},
significantly reducing the quadratic complexity associated with full attention mechanisms. 

\vspace{0.5em}
\sublabel{Sparse and Fuzzy Attention.} Building upon the dynamic communication pools concept, HgVT employs both sparse and fuzzy attention mechanisms to further optimize computational efficiency.
Vertex self-attention is applied selectively to pairs of vertices connected by common hyperedges, as defined by the adjacency matrix \scaledmath{\hat{\mathbf{A}}}, resulting in a sparse attention pattern. 
As sparsity increases with network depth -- demonstrated in \cref{fig:sparse_depth} -- computational load decreases, while still maintaining compatibility with dense attention during training. 
Conversely, the edge aggregate and distribution attention mechanisms utilize cross-attention between the vertex and edge feature matrices, \scaledmath{\mathbf{X}^\V} and \scaledmath{\mathbf{X}^\E}, modulated by the soft adjacency matrix $\mathbf{A}$.
This modulation, akin to Fuzzy C-Means in ViHGNNs \cite{vihgnn_paper}, adjusts attention logits based on soft memberships to the individual hyperedges, dynamically adapting to the hypergraph structure and providing a mechanism for gradient flow into the adjacency matrix generation.
Furthermore, by thresholding the soft adjacency matrix during inference, the edge attention mechanisms can be converted into a sparse cross-attention mechanism, thereby reducing computational overhead.


\subsection{Dynamic Adjacency Formation}

HgVT dynamically establishes its hypergraph structure to adapt to the varying semantic and spatial structures of different image inputs. It employs cosine similarity, akin to query-key interactions in attention mechanisms, to evaluate the alignment between vertex and hyperedge adjacency features. This approach allows hyperedges to ``query'' vertices for relevant features, providing a scale-invariant assessment that emphasizes the directionality of embedding vectors. The cosine similarity is subsequently transformed into adjacency membership using a sharpened sigmoid function:
\begin{equation}
\!\!\mathbf{A} = \sigma\left(\alpha \cdot \mathbf{\tilde{X}}^\V_\adj \left[\mathbf{\tilde{X}}^\E_\adj\right]^T\right),\; \mathbf{\tilde{X}}^{(*)}_\adj = \frac{\mathbf{X}^{(*)}_\adj}{||\mathbf{X}^{(*)}_\adj||_2 + \epsilon}
\end{equation}

Here, $\sigma$ denotes the sigmoid function and $\alpha=4$ is a sharpening factor, which pushes values away from zero to establish binary-like membership values in matrix $\mathbf{A}$. This soft adjacency matrix is further thresholded to create the hard adjacency matrix \(\mathbf{\hat{A}} = [\mathbf{A}\!>\!0.5]\), which provides binary memberships to facilitate sparse attention masking.



\subsection{Architecture Scaling}
\begin{table}[h]
    \vspace{-1em}
    \caption{Scaling variants of our HgVT architecture. 
    All models are trained at 224x224 resolution, except lite variant (HgVT-Lt), trained at 160x160. 
    Showing count for vertices ($i\mathcal{V}$, $v\mathcal{V}$), hyperedges ($p\mathcal{E}$, $v\mathcal{E}$), dim for adj. ($d_a$) and features ($d_f$), depth ($L$), and heads ($h$).
    }
    \centering
    \vspace{-0.5em}
    \resizebox{0.48\textwidth}{!}{
    \begin{tabular}{l | 
    C{0.5cm} C{0.5cm} C{0.5cm} C{0.6cm} | 
    c C{0.2cm} C{0.3cm} | R{0.8cm} c }
        \toprule
        \textbf{Model}  & \bm{$|i\mathcal{V}|$} & \bm{$|v\mathcal{V}|$} & \bm{$|p\mathcal{E}|$} & \bm{$|v\mathcal{E}|$} & \bm{$d_f+d_a$} & \bm{$L$} & \bm{$h$} & \textbf{Params} & \textbf{FLOPS} \\
        \midrule
        HgVT-Lt   & 100 & 12 & 32 & 6 & $128+64$   & 12 &  4 & 6.8M & 0.92B \\
        HgVT-Ti   & 196 & 16 & 50 & 8 & $128+64$   & 12 &  4 & 7.7M & 1.80B \\
        HgVT-S    & 196 & 16 & 50 & 8 & $224+96$  & 14 &  7 & 23M & 5.48B \\
        \bottomrule
    \end{tabular}
    }
    \label{tab:scaling_variants}
\end{table}


\noindent Building upon a hybrid scaling strategy inspired by DeiT~\cite{deit_paper} and ViG~\cite{vig_paper}, HgVT achieves a balanced computational footprint across various model sizes. 
Table \ref{tab:scaling_variants} specifies transformer scaling hyperparameters and delineates allocations for different vertex and edge types, where non-image vertices ($i\mathcal{V}$) are assigned fixed capacities as proposed by ViHGNN \cite{vihgnn_paper}. 
Additionally, we introduce a Ti-Lite variant~\mbox{(HgVT-Lt)} aimed at facilitating computationally efficient ablations within a constrained training budget.


\section{Enforcing Semantic Structure}

Feature matrices for virtual vertices and hyperedges, lacking direct input-based initialization, risk converging to homogeneous solutions and collapsed representations. Additionally, the dynamic adjacency calculation fails to naturally promote semantic grouping, in contrast to clustering-based approaches commonly used in vision GNNs. To address these issues, we introduce diversity regularization to enforce orthogonal embeddings and population regularization to encourage a structured, sparse hypergraph. For enhanced semantic differentiation of virtual hyperedge features in classification, we incorporate an expert-based pooling strategy as a more robust alternative to mean pooling.

\subsection{Diversity-Driven Feature Differentiation}

To prevent homogenization of learned feature matrices and to encourage distinct, semantically rich embeddings, we implement a diversity-driven regularization approach. This method, designed to maintain maximum orthogonality among the embeddings of virtual vertices and hyperedges, penalizes the absolute value of the cosine similarity between different feature vectors, aiming for values close to zero. By using normalized embeddings and masking off diagonal elements to preserve self-similarity, the approach prevents the model from converging to homogeneous solutions or driving individual vectors towards zero magnitude.
We then individually penalize $v\mathcal{V}$, $\mathcal{E}$, and their adjacency features.

\resizebox{0.49\textwidth}{!}{
\begin{minipage}{1.08\linewidth}
\begin{flalign}
\!\!\!\!\!\!&\mathrm{D}_\mathrm{L}\left(\mathbf{X}\right) = \frac{1}{2}\sum_{ij}\left(1 - \delta_{ij}\right) \cdot \left|\mathbf{\tilde{X}} \mathbf{\tilde{X}}^T\right|_{ij}, \; \mathbf{\tilde{X}} = \frac{\mathbf{X}}{||\mathbf{X}||_2 + \epsilon}&\!\!\!\!\!\! \\
\!\!\!\!\!\!&\mathcal{L}_\mathrm{DIV} = \sum_x \mathrm{D}_\mathrm{L}\left(x\right),\; x\in\{\mathbf{X}^{(:vV)},\mathbf{X}^{(:vV)}_\mathrm{adj}, \mathbf{X}^\E, \mathbf{X}^\E_\mathrm{adj} \}&\!\!\!\!\!\!
\end{flalign}
\end{minipage}

}
\vspace{0.2em}

%
\noindent Where $(:vV)$ represents the subset of $\mathcal{V}$ containing only the virtual nodes, $\delta_{ij}$ is the Kronecker delta function, ensuring that self-similarity is not penalized, and $|\cdot|_{ij}$ denotes the element-wise absolute value, applied to calculate the penalty for non-orthogonal relationships between embeddings.

\subsection{Population Regularization: Learned Sparsity}

Unlike clustering methods like KNN or Fuzzy C-Means, which enforce fixed cluster sizes, our model’s dynamic adjacency calculation allows for flexible, self-adjusting hyperedge populations. To prevent the associated risks of overly sparse or densely connected hypergraphs, we introduce population regularization. This method applies penalties based on the computed soft membership density of each hyperedge derived from the soft adjacency matrix $\mathbf{A}$, ensuring each maintains a vertex population within appropriate bounds to avoid overgeneralization and preserve hypergraph integrity.
\vspace{-0.2em}
\begin{align}
    \textstyle 
    &P_j = 2 \cdot \sum_i \mathrm{max}(A_{ij} - 0.5, 0)& \\
    &\mathcal{L}_\mathrm{POP} = \sum_j \mathrm{max}(P_j - \beta,0) + \mathrm{max}(\gamma - P_j,0)&
\end{align} 
\vspace{-1em}

\noindent Here, $P_j$ represents a soft density estimate of vertex connections for the \scaledmath{j^{th}} hyperedge, only considering non-zero entries of \scaledmath{\hat{\mathbf{A}}}. $\beta$ and $\gamma$ set the upper and lower density limits, ensuring that hyperedges maintain an optimal balance of connections. Penalties are applied if $P_j$ either exceeds $\beta$ or falls below $\gamma$, maintaining the desired sparsity and ensuring the structural efficacy of the hypergraph.

\subsection{Expert Pooling for Semantic Specialization}

To effectively combine features from multiple virtual hyperedges for classification, our approach utilizes a method akin to expert-choice, where each virtual hyperedge acts as an~\mbox{``expert''} generating a confidence score. Unlike mean pooling, which risks collapsing distinct features into an average that may dilute individual contributions, this strategy encourages virtual hyperedges to develop unique, semantically meaningful representations. The normalized confidence scores,~$P(e)$, determine the relevance of each hyperedge $e$'s contribution to the classification task, with only the top-k most confident scores selected for creating a weighted average and subsequent class prediction.
\begin{equation}
    \textstyle 
    P(e) = \mathrm{softmax}\left(\mathbf{X}^{(:vE)}\mathbf{W}_e + b_e\right)
\end{equation}
%
\noindent Here, \mbox{$(:\!vE)$} denotes the subset of $\E$ containing only the virtual hyperedges, and the softmax is computed across the expert gate set $\{e\}$ after projection by \scaledmath{\mathbf{W}_e\in \mathbb{R}^{d\times |:vE|}}. During training, $P(e)$ guides the weighted averaging of hyperedge features. For inference, a binary threshold enforces top-k routing, selectively integrating the most relevant hyperedge outputs based on their confidence. 
To prevent underutilization of any single virtual hyperedge, a density loss function \cite{fedus2022switch, chen2023sparse} is applied, complemented by a cross-entropy term with label smoothing to increase expert confidence. 





\section{Empirical Evaluation and Performance}

This section presents the evaluation of the Hypergraph Vision Transformer using two specific model configurations as detailed in \cref{tab:scaling_variants}: the HgVT-Ti-Lite for targeted ablation studies and scaled variants for benchmarks against comparable image classifiers. We apply standard augmentation techniques established by DeiT~\cite{deit_paper} across all datasets using the Timm library~\cite{rw2019timm}. Specifically, we use:  RandAugment~\cite{randaug_paper}, Mixup~\cite{mixup_paper}, Cutmix~\cite{cutmix_paper}, Random Erasing~\cite{random_erasing_paper}, and Repeated Augment~\cite{repeataug_paper}. 
\\

\vspace{-0.5em}
\sublabel{Datasets.} For classification in computer vision, we follow standard practices and use the \mbox{ImageNet-1k} dataset~\cite{imagenet_cvpr09} at a resolution of 224x224 pixels for scaled model evaluations. For ablation studies, we employ \mbox{ImageNet-100}~\cite{tian2020contrastive}, a 100-class subset of \mbox{ImageNet-1k} with images scaled to~160x160 pixels. This selection provides a computationally manageable dataset while maintaining sufficient class variation and larger image sizes compared to datasets like \mbox{CIFAR-100 (32x32 pixels)}\cite{cifar100_ref}. Nevertheless, we find \mbox{CIFAR-100} useful for assessing the effects of regularization on the hypergraph structure, as detailed in \cref{app:more_ablations}.
\\

\vspace{-0.5em}
\sublabel{Training Hyperparameters.} Consistent with DeiT, we use the AdamW optimizer with a weight decay of 0.05. Training is conducted on the \mbox{ImageNet-1k} dataset with a batch size of 1024 for 300 epochs following~DeiT. For ablations, we train on \mbox{ImageNet-100} with a batch size of 512 for with a shorter duration of 200 epochs as proposed~by~\cite{Lee_2022_CVPR}. Learning rates follow a cosine-annealing schedule peaking at 1e-3 for both datasets following scaling from~DeiT. Furthermore, we omit the use of Exponential Moving Average (EMA) due to its minimal performance improvement (0.1\% in DeiT) relative to its overhead per training step. All models were trained with bfloat16 mixed precision using PyTorch on local NVIDIA RTX A6000 GPUs, detailed further in \cref{app:implement}.
\\

\vspace{-0.5em}
\sublabel{Evaluation Metrics.} Following standard protocols, we measure the Top-1 and Top-5 class prediction accuracy to assess overall performance. Additionally, we take advantage of the learned graph structure (extracted from the last layer) on each image, and measure: Hyperedge Entropy~(HE), Intra-Cluster Similarity~(ICS), Inter-Cluster Distance~(ICD), and Silhouette Score~(SIL)~\cite{sil_score}; further details on graph structure measurements can be found in \cref{app:graph_quality}.


\subsection{Evaluation on ImageNet}

\vspace{-1em}

\begin{table}[h!]
    \caption{\mbox{ImageNet-1k} results for HgVT and other isotropic networks. \CNN CNN, \Transformer Transformer, 
    \GNN GNN, \HGNN HGNN, and \HGVT HgVT.
    }\label{tab:short_sota}
    \vspace{-0.5em}
    \centering
    \resizebox{0.48\textwidth}{!}{
    \begin{tabular}{l|c|c|c|c|c}
        \toprule
        \textbf{Model} & \textbf{Params} & \textbf{FLOPs} & \shortstack{ \textbf{ImNet} \\ \textbf{Top-1}} & \shortstack{ \textbf{ReaL} \\ \textbf{Top-1}} & \shortstack{ \textbf{V2} \\ \textbf{Top-1}}\\
        \midrule
        \CNN ResMLP-S12 conv3x3 \cite{res_mpl_paper} 
                                & 16.7M & 3.2B & 77.0 & 84.0 & 65.5 \\
        \CNN ConvMixer-768/32 \cite{convmixer} 
                                & 21.1M & 20.9B & 80.2 & -- & --\\
        \CNN ConvMixer-1536/20 \cite{convmixer} 
                                & 51.6M & 51.1B & 81.4 & -- & -- \\
        \midrule
        \Transformer DINOv1-S  \cite{dinov1_paper}
                                & 21.7M & 4.6B & 77.0 & -- & -- \\
        \Transformer ViT-B/16 \cite{vit16x16words} 
                                & 86.4M & 55.5B & 77.9 & 83.6 & --\\
        \Transformer DeiT-Ti  \cite{deit_paper}
                                & 5.7M & 1.3B & 72.2 & 80.1 & 60.4\\
        \Transformer DeiT-S  \cite{deit_paper}
                                & 22.1M & 4.6B & 79.8 & 85.7 & 68.5\\
        \Transformer DeiT-B  \cite{deit_paper}
                                & 86.4M & 17.6B & 81.8 & 86.7 & 71.5\\
        \midrule
        \GNN ViG-Ti \cite{vig_paper}
                                & 7.1M & 1.3B & 73.9 & -- & --\\
        \GNN ViG-S \cite{vig_paper}
                                & 22.7M & 4.5B & 80.4 & -- & --\\
        \GNN ViG-B \cite{vig_paper}
                                & 86.8M & 17.7B & 82.3 & -- & --\\
        \midrule
        \HGNN ViHGNN-Ti \cite{vihgnn_paper}
                                & 8.2M & 1.8B & 74.3 & -- & --\\
        \HGNN ViHGNN-S \cite{vihgnn_paper}
                                & 23.2M & 5.6B & 81.5 & -- & --\\
        \HGNN ViHGNN-B \cite{vihgnn_paper}
                                & 88.1M & 19.4B & 82.9 & -- & --\\
        \midrule
        \HGVT HgVT-Ti (ours) & 7.7M & 1.8B & 
        \textbf{76.2} & \textbf{83.2} & \textbf{64.3}\\
        \HGVT HgVT-S (ours) & 22.9M & 5.5B & 81.2 & \textbf{86.7} & \textbf{70.1}\\
        \bottomrule
    \end{tabular}
    }
\end{table}

\noindent \cref{tab:short_sota} presents the ImageNet-1k top-1 accuracy of HgVT, benchmarked against comparable isotropic models. Due to the complexities associated with downscaling virtual tokens lacking spatial alignment, we limit our analysis to isotropic architectures, excluding pyramidal models which generally exhibit superior performance due to hierarchical feature extraction \cite{vig_paper, vihgnn_paper}. Among the evaluated models, HgVT-Ti demonstrates a notable advantage, surpassing ViHGNN-Ti by 1.9\% in accuracy with 6\% fewer parameters and equivalent FLOPs. The HgVT-S model achieves accuracy comparable to ViHGNN-S, due to reduced layer count when matching parameters and FLOPs, constrained by scaling factors such as integer head counts in attention. Additionally, HgVT-S matches DieT-B’s accuracy on the ImageNet ReaL \cite{imagenet_reaL} and achieves competitive performance on ImageNet V2 \cite{imagenet_v2},  all while operating at nearly a quarter of DieT-B’s model size. Overall, these results underscore the efficiency of integrating hypergraph structures within a vision transformer framework, suggesting that HgVT provides a resource-efficient alternative for complex vision tasks without sacrificing performance.

\subsection{Ablation Studies}

We conducted a series of ablations on the ImageNet-100 dataset using the HgVT-Lt model, reporting Top-1 classification accuracy alongside mean hyperedge entropy and silhouette scores to assess the quality of the hypergraph’s structure. Notably, we observe a weak anti-correlation between the graph quality metrics and Top-1 accuracy (see Appendix \ref{app:more_ablations}), indicating opposing objectives. Overall, the ablations are grouped into three categories: regularization, architecture, and pooling methods, with results in \cref{tab:ablations}.

\begin{table}[h!]
    \caption{ImageNet-100 ablations with HgVT-Lt. Indicating used~(\Yes) or not used~(\No), and pooling methods: Average~(\BlackNum{A}), Image~(\BlackNum{I}), Expert~(\BlackNum{E}), Expert+Image~(\BlackNum{EI}),
    and \BlackNum{EI} dropping \BlackNum{I} (\BlackNum{DI}).
    } \label{tab:ablations}
    \centering
    \vspace{-0.5em}
    \resizebox{0.48\textwidth}{!}{
    \begin{tabular}{c| p{0.1cm} p{0.1cm} p{0.1cm} c| 
    p{0.1cm} p{0.05cm} p{0.15cm} c |
     p{0.5cm} c | p{0.7cm} c}
        Ablation on $\downarrow$ & 
        \rhead{CLS Dropout} &
        \rhead{Stoch. Decay} &
        \rhead{Diversity} &
        \rhead{Population} &
        \rhead{Tied FFN} &
        \rhead{$\mathbf{X}_\adj = \mathbf{X}$} &
        \rhead{$d_f$ Mult.} &
        \rhead{Pooling Op} & 
        \hspace{0.01cm}
        \rhead{Edge Entropy} &
        \rhead{Silhouette} &
        \hspace{0.1cm}
        \rhead{Top-1 Acc.} & 
        \rhead{Params (M)} 
        \\\toprule
        none: HgVT-Lt
        & \No & \Yes & \Yes & \Yes & 
        \Yes & \Yes & \BlackNum{1} & \BlackNum{EI} &
        \cworse 3.32 & \cbest 0.780 & \cbest 84.36 & \csame 6.75 \\
        \midrule
        \multirow{5}{*}{Regularization}

        & \Yes & \yes & \yes & \yes & 
        \yes & \yes & \GrayNum{1} & \GrayNum{E} &
        \csame 3.13 & \csame 0.751 &  \csame 82.23 & \csame 6.62  \\

        & \no & \No & \yes & \yes & 
        \yes & \yes & \GrayNum{1} & \GrayNum{E} &
        \csame 3.12 & \csame 0.745 &  \cworse 81.89 & \csame 6.62 \\

        & \no & \yes & \No & \yes & 
        \yes & \yes & \GrayNum{1} & \GrayNum{E} &
        \cbest 1.99 & \cworse 0.723 &  \cworse 80.79 & \csame 6.62  \\

        & \no & \yes & \yes & \No & 
        \yes & \yes & \GrayNum{1} & \GrayNum{E} &
        \cworst 3.89 & \cworst 0.639 & \cworse 81.79 & \csame 6.62  \\

        & \no & \yes & \No & \No & 
        \yes & \yes & \GrayNum{1} & \GrayNum{E} &
        \cworse 3.58 & \cworst 0.610 & \cworse 81.99 & \csame 6.62  \\
        
        \midrule
        \multirow{4}{*}{Architecture} 

        & \no & \yes & \yes & \yes & 
        \No & \yes & \GrayNum{1} & \GrayNum{E} &
        \csame 3.09 & \csame 0.741 & \csame 82.89 & \cworst 9.86  \\

        & \no & \yes & \yes & \yes & 
        \No & \No & \GrayNum{1} & \GrayNum{E} &
        \cbetter 2.27 & \cbest 0.808 &  \cworst 78.62 & \cbetter 5.84   \\

        & \no & \yes & \yes & \yes & 
        \yes & \No & \GrayNum{1} & \GrayNum{E} &
        \cbetter 2.12 & \cbest 0.780 &  \cworst 76.95 & \cbest 4.40   \\

        & \no & \yes & \yes & \yes & 
        \yes & \No  & \BlackNum{1.5} & \GrayNum{E} &
        \cbest 2.05 & \cbetter 0.770 &  \cworst 77.46 & \cworst 9.62  \\

        \midrule
        \multirow{4}{*}{Pooling} 

        & \no & \yes & \yes & \yes & 
        \yes & \yes & \GrayNum{1} & \BlackNum{A} &
        \csame 3.07 & \csame 0.747 &  \csame 82.06 & \csame 6.62  \\

        & \no & \yes & \yes & \yes & 
        \yes & \yes & \GrayNum{1} & \BlackNum{I} &
        \cbetter 2.93 & \cbetter 0.760 & \cbest 84.08 & \csame 6.62  \\


        & \no & \yes & \yes & \yes & 
        \yes & \yes & \GrayNum{1} & \BlackNum{E} &
        \csame 3.13 & \csame 0.749 &  \csame 82.52 & \csame 6.62 \\

        & \no & \yes & \yes & \yes & 
        \yes & \yes & \GrayNum{1} & \BlackNum{DI} &
        \cworse 3.32 & \cbest 0.780 & \cworse 80.94 & \csame 6.75 \\

        \bottomrule
    \end{tabular}
    }
\end{table}

As shown in \cref{tab:ablations}, the regularization ablations demonstrate that \textit{stochastic path dropout decay} \cite{stoch_depth_decay} improves both~\mbox{Top-1} accuracy and silhouette scores, consistent with ViG and~\mbox{ViHGNN}~\cite{vig_paper, vihgnn_paper}. Omitting \textit{Class dropout} also boosts accuracy, aligning with DeiT~\cite{deit_paper}. Futhermore, our proposed \textit{diversity} and \textit{population} regularization are essential for preserving graph structure; removing diversity leads to partial representation collapse, while removing population regularization results in near-zero sparsity, effectively turning HgVT into a ViT with increased network complexity.
%

In the \textit{architecture} ablations, untying the FFN improves accuracy but significantly increases parameter count, making it an unfavorable tradeoff. Tying adjacency and embedding features ($\mathbf{X}_\adj=\mathbf{X}$) reduces parameters and FLOPs but degrades performance, and while untying the FFN or increasing feature dimensionality partially mitigates this, the parameter increase remains suboptimal. This indicates that adjacency and embedding features ($\mathbf{X}^\E$ and $\mathbf{X}^\V$) are similar, yet require dedicated feature spaces to avoid crowding.

For \textit{pooling methods}, \textit{expert edge} pooling outperforms \textit{average edge} pooling in accuracy, while \textit{image} pooling achieves the highest accuracy at the cost of degraded graph structure. Combining image and expert pooling recovers lost structure and improves accuracy, with each input focusing on different semantic levels (see \cref{app:representations}). Additionally, dropping the pooled image embedding prior to the classifier head maintains moderate performance, indicating that both paths meaningfully contribute to the final prediction.


\begin{table}[t]
    \caption{Top-1 accuracy of HgVT-Lt on ImageNet-100 with~(\Yes) or without~(\No) vertex self-attention. Contrasting pooling operations and patch embedding versions:~Conv. Stem or Patch Projection. }\label{tab:nodesa}
    \vspace{-0.5em}
    \centering
    \resizebox{0.44\textwidth}{!}{
    \begin{tabular}{l|c c|c c|c c}
        \toprule
        Pooling Op. $\rightarrow$  & \multicolumn{2}{c|}{Average} & \multicolumn{2}{c|}{Image} & \multicolumn{2}{c}{Expert} \\
        \midrule
        Vertex SA $\rightarrow$ & \No & \Yes & \No & \Yes & \No & \Yes \\
        \midrule
        Conv. Stem & 78.02 & 82.06 & 82.05 & 84.08 & 78.87 & 82.52 \\
        Patch Project & 64.30 & 72.76 & 70.76 & 76.17 & 62.62 & 71.43 \\
        \bottomrule
    \end{tabular}
    }
    \vspace{-1em}
\end{table}

\vspace{1em}
\sublabel{Impact of Vertex Self-Attention and Patch Embedding.} 
\noindent We evaluated the impact of \textit{patch embedding} methods and \textit{vertex self-attention}, comparing a convolutional stem (Conv2D-BN-GELU layers \cite{vig_paper,vihgnn_paper}) and a simpler patch projection (pixel-shuffled patches with affine projection \cite{vit16x16words,dinov1_paper}), across various pooling strategies (average, image, and expert), as shown in~\cref{tab:nodesa}. The patch projection consistently underperforms the convolutional stem, likely due to the model's small size limiting its effectiveness. Omitting vertex self-attention leads to further degradation, especially without the convolutional stem, suggesting it is crucial for effectively separating features in low-dimensional space. 
PCA shows that~71/128 channels are needed to explain~95\% of the variance for the convolutional stem, compared to~19/128 for the patch projection, indicating the richer representation captured by the convolutional stem. Notably, image pooling shows the least degradation, likely due to its more direct gradient flow compared to the edge pooling methods.

\subsection{Pooling Methods and Graph Structure}

\vspace{-0.5em}
\begin{table}[ht]
    \vspace{-0.5em}
    \caption{Impact of graph structure on pooling operation for \mbox{HgVT-Lt} on ImageNet-100. Measuring graph metrics for image vertices ($i\mathcal{V}$) and all vertices ($\mathcal{V}$), along with features from DINOv2.
    }\label{tab:pooling_structure}
    \vspace{-0.5em}
    \centering
    \resizebox{0.48\textwidth}{!}{
    \begin{tabular}{l|
    p{0.4cm} p{0.4cm} c| 
    p{0.4cm} p{0.4cm} c|
    p{0.4cm} p{0.4cm} c}
        \toprule
        Pooling Op. $\rightarrow$  & \multicolumn{3}{c|}{Image} & \multicolumn{3}{c|}{Expert} & \multicolumn{3}{c}{Img. \& Expert} \\
        \midrule
        Feature Model $\downarrow$ & 
        HE & ICS & ICD & 
        HE & ICS & ICD & 
        HE & ICS & ICD 
        \\
        \midrule
        HgVT-Lt ($i\mathcal{V}$) &
        \csame 3.03 & \csame 0.50 & \csame 0.31 & 
        \csame 3.24 & \csame 0.43 & \csame 0.37 & 
        \csame 3.23 & \csame 0.42 & \csame 0.36 \\
        HgVT-Lt ($\mathcal{V}$) &
        \cworse 3.20 & \cworst 0.25 & \cbest 0.72 & 
        \cworse 3.39 & \cworse 0.36 & \cbetter 0.45 & 
        \cworse 3.39 & \cworse 0.28 & \cbetter 0.58\\
        \midrule
        DINO2-S ($i\mathcal{V}$)& 
        3.04 & 0.84 & 0.08 & 
        3.25 & 0.85 & 0.05 &
        3.25 & 0.83 & 0.06 \\
        DINO2-G ($i\mathcal{V}$)& 
        3.04 & 0.69 & 0.15 & 
        3.25 & 0.68 & 0.12 &
        3.25 & 0.68 & 0.13 \\
        \bottomrule
    \end{tabular}
    }
    \vspace{-0.5em}
\end{table}




\noindent To evaluate the impact of pooling methods on graph structure, we measured HE, ICS, and ICD using the ImageNet-100 validation set with three strategies: image pooling, expert pooling, and a combined image + expert pooling approach. Metrics used either the image vertex subset~($i\mathcal{V}$) or the full vertex set~($\mathcal{V}$), with features derived from \mbox{DINOv2}~(S and G)~\cite{oquab2024dinov2} and HgVT-Lt using the HgVT adjacency matrix~($A$). Results in \cref{tab:pooling_structure} show that while all methods achieve similar graph quality using $i\mathcal{V}$, image pooling slightly improves similarity. However, including all vertices~($\mathcal{V}$) consistently increases ICD and entropy while reducing ICS, indicating decreased graph coherence. This effect is more severe with image pooling methods, suggesting that virtual vertices~($v\mathcal{V}$) act as noisy elements, rather than summarization points.


Comparing DINOv2 models, all pooling methods align more closely with DINOv2-G, where achieving a balance between ICS and ICD is preferable to maximizing either individually. 
This trend, along with consistent HE, suggests a focus on higher-level detail, irrespective of the smaller HgVT model size or pooling method. 
Image pooling shows slightly stronger alignment with both DINO models, indicating that both high- and low-level semantics are encoded within a single feature space, unlike methods that can use edge channels for high-level concepts. Notably, all expert pooling methods exhibit an emergent macro-class prediction behavior, where each virtual edge ($v\mathcal{E}$) consistently captures broader taxonomic groups (e.g., dogs, birds). Further representation and macro-class analysis are provided in \cref{app:representations,app:macro_prediction}.

\subsection{Performance on Image Retrieval}

\noindent To evaluate the capability of HgVT in capturing semantic structures,  we perform image retrieval experiments comparing four methods: pooling similarity~(PS), volumetric similarity~(VS), adaptive pooling similarity~(APS), and adaptive volumetric similarity~(AVS). PS ranks the pooled embeddings by cosine similarity (vector search), while the other methods enhance retrieval by leveraging graph structure. Volumetric similarity calculates ellipsoid overlap using an approximate Mahalanobis distance, with pruned hyperedges defining the distribution spread around the pooled embedding (centroid). Adaptive methods further refine retrieval via a graph similarity measure on the pruned hyperedges, re-ranking from a shortlist of $R=100$.  Computational efficiency in adaptive retrieval is ensured through centroid hash binning and limiting comparisons to the \mbox{top-$C=4$} most significant query hyperedges, resulting in a complexity of~$\mathcal{O}(RC)$. Notably, we prune to 12 hyperedges and use 10 centroid bins; additional details provided in \cref{app:retrieval}.

\begin{table}[h!]
    \caption{Image Retrieval on \mbox{ImageNet-1k} with proposed search methods:~PS, APS, VS, and AVS. Reporting Top-1 accuracy, mAP@10~(\%), and 1-NN-hit@10~(\%) for pooling methods. 
    } \label{tab:imnet_ret}
    \centering
    \vspace{-0.5em}
    \resizebox{0.48\textwidth}{!}{
    \begin{tabular}{l | c | c | c c | c c }
    \toprule
    \multicolumn{7}{c}{\textbf{mAP@10~(\%)}} \\
    \midrule
    Model & Top-1 & Pooling & PS & APS & VS & AVS \\

    \midrule
    MRL-128 \cite{Kusupati2022MatryoshkaRL} & 70.52 & Image & 64.94 & 65.20 & -- & -- \\
     MRL-256 \cite{Kusupati2022MatryoshkaRL} & 70.62 & Image & 65.04 & 65.20 & -- & -- \\
     \midrule
HgVT-Ti (ours) & 76.18 & Expert & 70.56 & 69.59 & 70.53 & 69.40 \\
HgVT-Ti (ours) & 76.18 & Im\&Ex & 73.23 & 69.59 & 73.08 & 69.49 \\
\midrule
\multicolumn{7}{c}{\textbf{1-NN-hit@10~(\%)}} \\
\midrule
HgVT-Ti (ours) & 76.18 & Expert & 21.22 & 19.10 & 21.25 & 19.56 \\
HgVT-Ti (ours) & 76.18 & Im\&Ex & 25.13 & 19.10 & 25.22 & 19.17 \\
     
    \bottomrule
    \end{tabular}
    }
\end{table}

\sublabel{ImageNet Retrieval.}
We evaluate retrieval performance on the ImageNet-1K dataset to assess HgVT’s ability to capture semantic relationships and compare four retrieval methods: PS, VS, APS, and AVS. The primary metric is mAP@10, with MRL~\cite{Kusupati2022MatryoshkaRL} serving as the baseline due to its adaptive re-ranking approach. We also report the 1-NN-CLIP-L hit-rate@10, which measures how often the top-1 result ranked by CLIP-L~\cite{clip_paper} appears in the top-10 retrieved results, offering additional insight into semantic alignment. Results in \cref{tab:imnet_ret} show that HgVT-Ti surpasses MRL by over 8\% in retrieval performance, despite being significantly smaller than MRL (ResNet-50) and using a comparably compact embedding size ($d=2\times 128$). Among our methods, PS and VS achieve similar results, while APS and AVS underperform, likely due to their focus on exact-feature similarity and ambiguous-class features, which limits alignment with the high-level semantics required by ImageNet’s diverse dataset.

\begin{table}[h!]
    \caption{Image Retrieval on revisited Oxford and Paris; reporting mAP (\%). Showing training set and method used. $^*$mAP@100.
    }\label{tab:oxpar_ret}
    \vspace{-0.75em}
    \centering
    \resizebox{0.48\textwidth}{!}{
    \begin{tabular}{l | c | c | C{0.5cm} c | C{0.5cm} C{0.4cm} }
    \multicolumn{3}{r|} {Dataset $\rightarrow$} & \multicolumn{2}{c|}{$\mathcal{R}$Oxford} & \multicolumn{2}{c}{$\mathcal{R}$Paris} \\
    \toprule
    Model & Train Set & Method & M & H & M & H \\
    \midrule
    ALEX+GeM \cite{rn101_max_oxpar_paper} & ImNet-1k &
     PS & 33.8 & 10.4 & 52.7 &  26.0 \\
    RN101+R-MAC \cite{rn101_max_oxpar_paper} & ImNet-1k &
     PS & 49.8 & 18.5 & 74.0 &  52.1 \\
    DINOv1-S/16 \cite{dinov1_paper} & ImNet-1k & 
     PS & 41.8 & 13.7 & 63.1 & 34.4 \\
    DINOv1-S/16 \cite{dinov1_paper} & GLDv2 &
     PS & 51.5 & 24.3 & 75.3 & 51.6 \\
    \midrule
    \multirow{3}{*}{HgVT-Ti (ours)} &  \multirow{3}{*}{ImNet-1k} & 
    VS & 26.8 & 10.5 & 55.4 & 28.7 \\
    & & 
    VS$^*$ & 25.8 & 9.0 & 65.6 & 28.0 \\ 
    & & 
    AVS$^*$ & 27.0 & 6.8 & 64.1 & 26.7 \\ 
    \midrule
    \multirow{3}{*}{HgVT-S (ours)} &  \multirow{3}{*}{ImNet-1k} & 
    VS & 28.0 & 12.1 & 56.7 & 31.1 \\
    & & 
    VS$^*$ & 26.3 & 10.2 & 65.0 & 28.4 \\ 
    & & 
    AVS$^*$ & 27.4 & 10.3 & 65.0 & 27.1 \\ 
    \bottomrule
    \end{tabular}
    }
    \vspace{-0.25em}
\end{table}


\sublabel{Oxford and Paris Retrieval.} To evaluate image retrieval performance beyond simple class retrieval, we use the Oxford and Paris Revisited datasets~\cite{oxford_paris, revisit_oxford}, which provide three splits of increasing difficulty (Easy, Medium, and Hard) for query/database pairs. We report Mean Average Precision (mAP) for the Medium (M) and Hard (H) splits, using mAP@100 for AVS based on short-list ranking, while full mAP and mAP@100 are provided for VS as a baseline comparison. Results, shown in \cref{tab:oxpar_ret}, indicate that HgVT  achieves competitive performance with similarly sized feature extractors, though performance on $\mathcal{R}$Oxford-M lags. This shortfall may stem from subtle landmark differences better captured multi-scale Conv-Nets and self-supervised learning, compared to the more salient focus driven by HgVT's classifier training. However, AVS outperforms VS on $\mathcal{R}$Oxford-M, demonstrating its ability to uncover finer feature similarities within the hypergraph structure.

\section{Conclusion and Future Directions}
In this work, we introduced the Hypergraph Vision Transformer (HgVT), a framework that integrates hypergraph structures into vision transformers to improve semantic understanding in visual tasks. HgVT achieves strong results on image classification and retrieval, outperforming prior tiny-scale isotropic models by 1.9\% on ImageNet-1k classification. Our methods, including population and diversity regularization and expert edge pooling, enhance semantic representation and efficiency by enabling dynamic hypergraph construction.
Future work will focus on exploring the scalability of hypergraph structures and integrating self-supervised learning to further improve adaptability and better decouple saliency from semantic vision graph generation.

{
    \small
    \bibliographystyle{ieeenat_fullname}
    \bibliography{main}
}


\onecolumn
\appendix

\let\addcontentsline\oldaddcontentsline  

\setcounter{tocdepth}{2}  

\begingroup
\renewcommand{\baselinestretch}{0.8}  
\setlength{\parskip}{0pt}  
\tableofcontents  
\endgroup

\clearpage  
\section{HgVT Model Architecture Details}
\label{app:app_arch_details}

The Hypergraph Vision Transformer (HgVT) adapts the architecture of standard Vision Transformers by incorporating hypergraph features to enhance image analysis capabilities. 
Similar to Vision Transformers, HgVT utilizes a patch embedding layer as its entry point, followed by an isotropic stack of $L$ HgVT blocks, each based on the ubiquitous Llama blocks\footnote{HgVT uses fixed sinusoidal position embeddings rather than rotary position embeddings.}
\cite{touvron2023llama}, culminating in feature pooling and a classifier head. 
Configured to process both vertex and edge information, the blocks include six main components: adjacency mask computation, vertex self-attention, edge aggregate attention, edge distribution attention, and separate feed-forward networks for vertices and edges. 
This configuration facilitates dynamic bipartite graph construction within each block, allowing the model to adaptively refine the input image's representative hypergraph.

\begin{figure}[ht]
  \centering
  \includegraphics[width=0.9\textwidth]{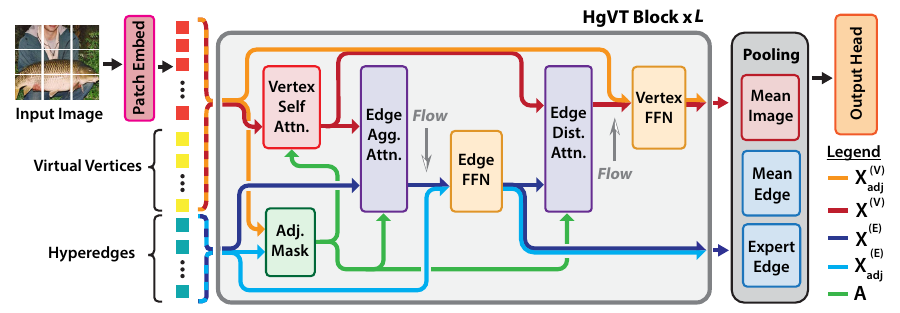}
  \caption{HgVT Architecture, composed of stacked HgVT blocks with adjacency matrix $\mathbf{A}$, vertex features $\mathbf{X}^\V$, and hyperedge features $\mathbf{X}^\E$. Pooling only applied to $\mathbf{X}^{(:iV)}$ and $\mathbf{X}^{(:vE)}$; edge attention flow shown with gray arrows; norms and residual omitted for clarity.}
  \label{fig:ap_hgvt_arch}
\end{figure}

Four key feature matrices -- $\mathbf{X}^\V$, $\mathbf{X}^\V_\adj$, $\mathbf{X}^\E$, and $\mathbf{X}^\E_\adj$ -- represent the bipartite hypergraph features and are progressively updated in each HgVT block in an interleaved manner. 
Each block also constructs a new adjacency matrix $\mathbf{A}$ from the input~$\mathbf{X}^\V_\adj$ and~$\mathbf{X}^\E_\adj$ matricies, which then contributes to the attention layers within that block. 
As illustrated in Figure \ref{fig:ap_hgvt_arch}, this approach allows for the dynamic integration and processing of these matrices within each HgVT block, facilitating effective feature interaction and updating.

For succinct discussion in subsequent sections, the update process for each layer $l$ is encapsulated using the following compact notation:
\begin{equation}
\textstyle 
\mathbf{X}^{(*, l+1)}_* = \mathbf{X}^{(*, l)}_* + \mathbf{X}^{(*, l)\prime}_*, \quad \mathbf{X}^{(*, l)\prime}_* = f\left(\mathrm{RN}\left(\mathbf{X}^{(*, l)}_*\right), \dots, \mathbf{A}^{(l)}\right)
\end{equation}
where $\mathrm{RN}(\cdot)$ denotes the RMS Norm \cite{zhang2019root}, and $\mathbf{X}^{(*, l)}_*$ includes both vertex features and hyperedge features, along with their respective adjacency features. The update function $f(\cdot)$ can utilize all four normalized feature matrices and the adjacency matrix $\mathbf{A}^{(l)}$, which is updated once per HgVT block.

\subsection{Dynamic Adjacency Formation}

Dynamically establishing the hypergraph structure is crucial for adaptability across varying semantic and spatial structures inherent in different image inputs.
Mirroring the query-key interactions found in attention mechanisms, HgVT utilizes cosine similarity to evaluate the alignment between vertex and hyperedge adjacency features.
This similarity assessment allows hyperedges to effectively ``query'' vertices for relevant features, establishing a scale-invariant comparison that focuses on the directionality of embedding vectors. 
To convert the cosine similarity to adjacency membership, we then form the soft adjacency matrix $\mathbf{A}$ with a sharpened sigmoid function, detailed as follows:
\begin{equation}
    \mathbf{A} = \sigma\left(\alpha \cdot \mathbf{\tilde{X}}^\V_\adj \left[\mathbf{\tilde{X}}^\E_\adj\right]^T\right), \;\;\; \mathbf{\tilde{X}}^{(*)}_\adj = \frac{\mathbf{X}^{(*)}_\adj}{||\mathbf{X}^{(*)}_\adj||_2 + \epsilon}
\end{equation}
where \(\mathbf{\tilde{X}}^{(*)}_\adj\) represents the L2-normalized adjacency feature matrix.
Here, $\sigma$ denotes the sigmoid function, and $\alpha=4$ acts as a sharpening factor, enhancing the sigmoid's effectiveness by pushing intermediate values toward the extremes.
The hard membership adjacency matrix \(\mathbf{\hat{A}} = [\mathbf{A} > 0.5]\) transforms these sigmoid outputs into binary memberships, crucial for defining significant hypergraph relationships and suitable for sparse attention masking.
In configurations where feature matrices and adjacency feature matrices are tied ($\mathbf{X}^{(*)}_\adj = \mathbf{X}^{(*)}$), $\mathbf{X}^{(*)}_\adj$ is computed as $\mathbf{X}^{(*)} \mathbf{W}_*$, using a learned projection matrix to adapt features for adjacency computation and maintain embedding adaptability.



\subsection{Vertex Message Passing with Sparse Self-Attention}

Shifting from traditional hypergraph models, which typically employ a gather $\rightarrow$ scatter mechanism for processing vertex-hyperedge interactions, HgVT reconceptualizes hyperedges as communication pools that facilitate dynamic and efficient information flow among vertices and their associated hyperedges.
Instead of relying on a single dense attention operation, HgVT organizes communication into two distinct streams: intra-hyperedge message passing and interactions between hyperedges  and their constituent vertices. 
By enabling direct message passing within hyperedges, the model significantly enhances inter-vertex communication, allowing for more nuanced integration of contextual information.
Furthermore, this configuration restricts interactions to vertices that share hyperedges, naturally inducing sparsity in the interaction matrix and substantially reducing computational overhead. 
The strategy for message passing between vertices within a hyperedge~\mbox{($\mathcal{V}_e\rightarrow \mathcal{V}_e$)} is implemented through the following update process:


\begin{equation}
\textstyle 
    \mathbf{X}^{\V\prime} = \mathrm{softmax}\left(\left(\mathbf{X}^{\V}\mathbf{W}_Q\right)\left(\mathbf{X}^{\V}\mathbf{W}_K\right)^T + \mathbf{B}\right)\left(\mathbf{X}^{\V}\mathbf{W}_V\right),\;\; \mathbf{B} = 1 -\left[\left(\mathbf{\hat{A}}\mathbf{\hat{A}}^T\right) > 0 \right]
\end{equation}
In this equation, the mask $\mathbf{B}\in\{0,1\}^{|V|\times|V|}$ is a dynamically computed based on the connectivity within the hyperedges, derived from the hard adjacency matrix $\hat{\mathbf{A}}\in\{0,1\}^{|V|\times|E|}$.
This masking ensures that attention computations are confined to vertices within the same hyperedge, enhancing communication efficiency. 
Additionally, for simplification, the typical attention scaling factor $1/\sqrt{d_k}$, which is generally used to stabilize the softmax calculations, is omitted from the above equation.




\subsection{Hyperedge Message Passing with Fuzzy Cross-Attention}

Completing the concept of hyperedges as dynamic communication pools outlined in the previous section, HgVT utilizes cross-attention mechanisms to facilitate interactions between hyperedges and their constituent vertices.
These mechanisms -- hyperedge aggregation attention~\mbox{($\mathcal{V}_e\rightarrow \mathcal{E}_e$)}, focusing on gathering information, and hyperedge distribution attention~\mbox{($\mathcal{E}_e\rightarrow \mathcal{V}_e$)}, dedicated to scattering information -- leverages HgVT's unique bipartite representation for effective management of information flows within these pools.
By modulating the attention logits with the soft adjacency matrix~$\mathbf{A}$ via a Hadamard product, the model introduces a layer of ``fuzziness'' to the typical cross-attention mechanism.
Such modulation dynamically aligns the model’s response to the varied connectivity patterns typical in hypergraph structures, thereby enhancing both precision and adaptability in processing information.
The equations that formalize these attention processes are presented below:
\begin{align}
    \textstyle
    \mathbf{X}^{\E\prime} &= \mathrm{softmax}\left(\left(\mathbf{X}^{\E}\mathbf{W}_Q\right)\left(\mathbf{X}^{\V}\mathbf{W}_K\right)^T\circ \mathbf{A}^T + \mathbf{M}^T\right)\left(\mathbf{X}^{\V}\mathbf{W}_V\right) \label{eqn:app_eagg}\\
     \mathbf{X}^{\V\prime} &= \mathrm{softmax}\left(\left(\mathbf{X}^{\V}\mathbf{W}_Q\right)\left(\mathbf{X}^{\E}\mathbf{W}_K\right)^T\circ \mathbf{A} + \mathbf{M}\right)\left(\mathbf{X}^{\E}\mathbf{W}_V\right) \label{eqn:app_edist}
\end{align}
In this framework, the soft adjacency matrix $\mathbf{A}\in\mathbb{R}^{|V|\times|E|}$ modulates the attention logits through a Hadamard product ($\circ$), dynamically reflecting the true connectivity of vertices to hyperedges and providing a gradient path to update the weights used to compute the adjacency feature matrices.
Concurrently, the static interaction mask $\mathbf{M}\in\{0,1\}^{|V|\times|E|}$ prevents virtual hyperedges~\mbox{($v\mathcal{E}$)} from interacting with image vertices ($i\mathcal{V}$), ensuring the maintenance of the hierarchical hypergraph structure described in Section \ref{sec:imposing_hierarchical_structure} within the architecture.
As before, the $1/\sqrt{d_k}$ factor is omitted for clarity.




\subsection{Sign Preserving Fuzzy Cross-Attention Modulation}

While simple to implement, the Hadamard modulation introduced in the previous section is sub-optimal due to properties of the softmax function, where weights of zero will bias the distribution (e.g. $e^0=1$). More specifically, since $A_{ij}\in[0,1)$, and we set $A_{ij} > 0.5$ to indicate membership, non-membership logits can still exhibit a positive attention contribution. Similarly, maximal dissimilarity ($A_{ij}=0$) will move negative logits closer to zero, potentially resulting in undesirable interactions. To address this issue, we adopt sign preserving modulation, which uses the shifted adjacency form ($\tilde{\mathbf{A}} =  2\mathbf{A}-1$), resulting in all non-membership logits becoming negative, while preserving the sign of the membership logits. 
\begin{equation}
    \tilde{\circ}(\mathbf{S},\tilde{\mathbf{A}}) = \underset{-1 \leq x \leq 1}{\mathrm{Clamp}}\left( \mathrm{Sign}(\mathbf{S}) + \mathrm{Sign}(\tilde{\mathbf{A}}) + 1\right)\circ \left(\mathbf{S} \circ \tilde{\mathbf{A}}\right)
\end{equation}
where $\mathbf{S}$ represents the pre-masked attention logits (e.g. $\mathbf{Q}\cdot\mathbf{K}^T$), and $\mathbf{A}$ is the soft adjacency matrix. The modified Hadamard product $\tilde{\circ}$ then replaces the normal Hadamard product in \cref{eqn:app_eagg,eqn:app_edist}. To better understand this functional form, we can consider the behavior table as shown in \cref{tab:app_modhad}. 

\begin{table}[h]
\centering
\begin{tabular}{c|lll}
$\tilde{\mathbf{A}}$ \textbackslash $\;\mathbf{S}$ & \hspace{4em}\textbf{+} & \hspace{4em}\textbf{0} & \hspace{4em}\textbf{-} \\ \midrule
\textbf{+} & 
$[3]\rightarrow 1: \tilde{\mathbf{A}}\circ\mathbf{S} > 0$ & 
$[2]\rightarrow 1: 1\cdot\mathbf{0} = 0$ & 
$[1]\rightarrow 1: \tilde{\mathbf{A}}\circ\mathbf{S} < 0$ \\
\textbf{0} & 
$[2]\rightarrow 1: 1\cdot\mathbf{0} = 0$ & 
$[1]\rightarrow 1: 1\cdot\mathbf{0} = 0$ & 
$[0]\rightarrow 0: 0\cdot\mathbf{0} = 0$ \\
\textbf{-} & 
$[1]\rightarrow 1: \tilde{\mathbf{A}}\circ\mathbf{S} < 0$ & 
$[0]\rightarrow 0: 0\cdot\mathbf{0}  = 0$ & 
$[-1]\rightarrow -1: -\tilde{\mathbf{A}}\circ\mathbf{S} < 0$ \\

\end{tabular}
\caption{Behavior table for the modified Hadarmard product $\tilde{\circ}$. Showing how the input signs for $\tilde{\mathbf{A}}=2\mathbf{A}-1$ and $\mathbf{S}$ affect the output, with the pre-clamp sum in square brackets, the clamp output after the $\rightarrow$, the resultant form, and the output sign.}
\label{tab:app_modhad}
\end{table}

To implement this modified Hadamard product, we pre-compute the element-wise correction term, defined as~\mbox{$\phi: (a, s) \in \mathbb{R}^2 \to \{-1, 0, 1\}$}, and compute the full function as a 3-element Hadamard product. Notably, this correction term has a zero derivative with respect to $a$ and $s$, except at the boundaries, where it is undefined. Therefore, we can avoid complications with differentiation by applying a gradient stop to the pre-sum correction term.

\subsection{Hypergraph Feature Processing}

Distinct point-wise feed-forward networks (FFNs) are utilized to process vertex and hyperedge features independently within the HgVT blocks, ensuring differentiated processing for each set within the bipartite representation.
These features are integrated with adjacency features through a dense, fully-connected GeGLU \cite{shazeer2020glu} layer, allowing the FFN to effectively combine both immediate and relational attributes.
By updating both feature types and their adjacency embeddings within the same FFN layer, the model centralizes computational tasks and simplifies the message passing process by focusing solely on feature updates, avoiding the direct involvement of adjacency features and thus improving computational efficiency. 
The update rules are governed by the following equation:
\begin{equation}
    \textstyle 
    \mathbf{X}^{(*)\prime}_\adj = \mathrm{FFN}\left(\mathbf{X}^{(*)}_\adj||\mathbf{X}^{(*)}\right), \;\;\;\;\mathbf{X}^{(*)\prime} = \mathrm{FFN}\left(\mathbf{X}^{(*)}_\adj||\mathbf{X}^{(*)}\right)
\end{equation}
Here, $(*)$ represents either the vertex set $\mathcal{V}$ or hyperedge set $\mathcal{E}$, and $||$ represents concatenation.

\subsection{Additional Variation Options for Efficiency}

To enhance efficiency, several potential paths exist to reduce parameters and FLOPS, aligned with the principles of graph neural networks. 
The feature matrices $\mathbf{X}^{(*)}$ and $\mathbf{X}^{(*)}_\adj$ can either be identical or have different dimensionalities, thereby simplifying the computational requirements of the FFN layers.
Additionally, tying both FFN layers to share weights further reduces the parameter count. 
Consistent with practices in Graph Attention Networks \cite{GAT_paper}, the weights for vertex self-attention~\mbox{($W_Q,W_K,W_V$)} and edge cross-attention~\mbox{($W_K, W_V$)} can also be tied. 
Implementing these strategies offers a range of options to tailor HgVT variants for balancing memory usage and computational efficiency, optimizing the model for various deployment environments based on performance needs.

\FloatBarrier
\section{Computational Overhead}
\label{app:app_comp_eff}

In this section, we explore the computational overhead of our proposed HgVT models relative to other isotropic models. All benchmarking experiments were conducted on an NVIDIA Quadro RTX 4000 GPU, using PyTorch 2.5.1 with CUDA 12.2. We evaluated all models in 32-bit precision with a batch size of 32. To ensure stable measurements, we aggregated statistics over 100 iterations, following an initial 10 warmup iterations to mitigate the impact of GPU initialization overhead. The comparative results are presented in \cref{tab:app_compeff}, which also includes a detailed cost breakdown, summed over all layers of the same type. For completeness, we also report computational performance for a theoretical HgVT-B model ($d_f = 448$, $d_a = 128$, $L = 16$, $h=14$) that was not trained but included to illustrate its expected cost. Notably, we were unable to benchmark ViHGNN \cite{vihgnn_paper} due to reproducibility issues with the publicly released code and have therefore excluded it from the table.

\newcommand{\PM}{\hspace{-0.1em}$\pm$\hspace{0.01em}}

\begin{table}[h!]
    \caption{Comparison of inference performance for HgVT and other isotropic networks. All results measured using 32-bit precision and a batch size of~32 on an NVIDIA Quadro RTX 4000 GPU. Further showing time per component and overall inference percentage, with Spatial denoting either Self-Attention or Graph Conv layers. \Transformer Transformer, 
    \GNN GNN, 
    and \HGVT HgVT. $^\dagger$Hypothetical HgVT-B model (not trained).
    }\label{tab:app_compeff}
    \vspace{-0.5em}
    \centering
    \resizebox{1.0\textwidth}{!}{
    \begin{tabular}{l|p{0.8cm} c| p{0.5cm} c|p{0.8cm} p{0.3cm}| p{1.4cm} p{1.4cm} | p{1.45cm} p{1.6cm} p{1.6cm} p{1.5cm} p{1.45cm} p{1.45cm}}
        \toprule
        & & & \multicolumn{2}{c|}{\textbf{ImageNet}} &
        \multicolumn{2}{c|}{\textbf{VRAM (MB)}} &
        \multirow{2}{*}{\makecell{\textbf{Batch}\\\textbf{Time (ms)}}}&
        \multirow{2}{*}{\makecell{\textbf{Speed}\\\textbf{(imgs/s)}}}&
        \multicolumn{6}{c}{\textbf{Time per Component (ms)}}
        \\
        
        \textbf{Model} & \textbf{Params} & \hspace{0.05cm}\textbf{FLOPs}\hspace{-0.1cm} & \textbf{\mbox{Top-1}} & \hspace{0.05cm}\textbf{Top-5}\hspace{-0.1cm} & 
        \textbf{Static} & \textbf{Peak} & & &
        \textbf{Patch} & \textbf{Spatial} & \textbf{FFN} &
        \textbf{Cluster} & \textbf{Aggregate} &\textbf{Distribute}
        \vspace{-1.1em}
        \\
        \midrule
        \Transformer DeiT-Ti  \cite{deit_paper}
        & 5.7M & 1.3B & 
        72.2 & 91.1 & 
        48.5 & 104 & 
        23.2 \PM 0.1 & 1370 \PM 8 & 
        0.6 (2.6\%) & 9.7 (41\%) & 9.5 (40\%) & 
         -- & -- & -- 
                                \\
        \GNN ViG-Ti \cite{vig_paper}
        & 7.1M & 1.3B & 
        73.9 & 92.0 & 
        54.3 & 331 & 
        79.5 \PM 0.3 & 402 \PM 1.6 & 
        3.0 (3.8\%) & 49.4 (62\%) & 13.5 (17\%) & 
         12 (15\%) & -- & -- 
                                \\
        \HGVT HgVT-Ti (ours) & 7.7M & 1.8B & 
        76.2 & 93.2 & 
        56.6 & 210 & 
        47.1 \PM 0.2 & 679 \PM 3.0  & 
        2.5 (5.4\%) & 9.0 (19\%) & 16.0 (34\%) & 
        1.5 (3.1\%) & 4.5 (9.6\%) & 5.0 (10\%) 
        \\

        \midrule
        \Transformer DINOv1-S  \cite{dinov1_paper}
        & 21.7M & 4.6B & 
        77.0 & -- &  
        119 & 249 & 
        68.4 \PM 0.4 & 468 \PM 2.5  & 
        1.2 (1.7\%) & 29.6 (43\%) & 32.8 (48\%) & 
         -- & -- & -- 
                                \\
        \Transformer DeiT-S  \cite{deit_paper}
        & 22.1M & 4.6B & 
        79.8 & 95.0 &  
        111 & 223 & 
        64.3 \PM 0.4 & 498 \PM 2.7  & 
        1.1 (1.8\%) & 25.7 (40\%) & 32.3 (50\%) & 
         -- & -- & -- 
                                \\
        \GNN ViG-S \cite{vig_paper}
        & 22.7M & 4.5B & 
        80.4 & 95.2 & 
        114 & 573 & 
        191 \PM 1.1 & 168 \PM 0.9  & 
        6.5 (3.4\%) & 120 (63\%) & 41.7 (22\%) & 
         20 (11\%) & -- & -- 
                                \\
        \HGVT HgVT-S (ours) & 22.9M & 5.5B & 
        81.2 & 95.5 & 
        116 & 365 & 
        113 \PM 0.5 & 282 \PM 1.3  & 
        6.0 (5.3\%) & 22.4 (20\%) & 45.9 (41\%) & 
         1.9 (1.7\%) & 11 (10\%) & 11 (9.9\%) 
        \\
        

        \midrule
        \Transformer ViT-B/16 \cite{vit16x16words} 
        & 86.4M & 55.5B & 
        77.9 & -- & 
        372 & 633 & 
        221 \PM 1.3 & 145 \PM 0.9  & 
        2.3 (1.0\%) & 87.6 (39\%) & 126 (56\%) & 
         -- & -- & -- 
                                \\
        \Transformer DeiT-B  \cite{deit_paper}
        & 86.4M & 17.6B & 
        81.8 & 95.7 & 
        357 & 579 & 
        213 \PM 1.3 & 150 \PM 0.9  & 
        2.2 (1.0\%) & 80.2 (37\%) & 124 (58\%) & 
         -- & -- & -- 
                                \\
        \GNN ViG-B \cite{vig_paper}
        & 86.8M & 17.7B & 
        82.3 & 95.9 & 
        359 & 1271 & 
        449 \PM 4.7 & 71.2 \PM 0.7  & 
        20 (4.4\%) & 281 (61\%) & 127 (28\%) & 
         27 (5.8\%) & -- & -- 
                                \\

        \HGVT HgVT-B (ours)$^\dagger$ & 87.9M & 20.4B & 
        -- & -- & 
        367 & 813 & 
        323 \PM 3.0 & 99.0 \PM 0.9  & 
        18 (5.6\%) & 56.0 (17\%) & 157 (49\%) & 
         2.5 (0.8\%) & 31 (9.5\%) & 32 (9.6\%) 
        \\
        

        \bottomrule
    \end{tabular}
    }
    \vspace{-1em}
\end{table}

The results are summarized in \cref{tab:app_compeff}, where models are grouped by scale and ordered by Top-1 ImageNet accuracy. The main points of comparison are other vision transformers \cite{dinov1_paper, vit16x16words, deit_paper} and ViG \cite{vig_paper}, a graph convolution-based model. We find that both ViG and HgVT exhibit higher latency and lower throughput than comparable vision transformers. For ViG, this increased cost is attributed to the computationally expensive graph convolution operations. In the case of HgVT, the increased cost stems from the second FFN layer (used for edges) and the additional attention operations for aggregation and distribution steps (as part of the bipartite hypergraph communication pool framework). However, despite this added complexity, HgVT remains within $2\times$ the performance of vision transformers. Notably, HgVT demonstrates lower self-attention cost despite operating on a larger sequence length (246 vs 196 for a $224^2$ resolution), resulting from the reduced hidden dimension.

We also find that the expert edge pooling strategy has a negligible effect on inference performance (accounting for less than~0.3\% of total inference cost), and HgVT’s regularization strategy exhibits no inference cost, as it is only used to learn how to construct well-structured hypergraphs that can generalize at inference time. When comparing with ViG, HgVT consistently outperforms in both throughput ($1.4\times-1.6\times$) and peak memory usage ($0.6\times$). Finally, HgVT’s implicit clustering approach is an order of magnitude faster than ViG’s KNN-based clustering, highlighting the benefits of learned self-sparsification with dynamic regularization over explicit clustering.

\subsection{Improving Computational Efficiency}

Although hypergraph-based models are often perceived as computationally expensive, HgVT demonstrates competitive performance and memory efficiency, outperforming ViG in both throughput and peak memory usage. However, further improvements in computational efficiency are possible through targeted optimizations. One promising direction is reordering the hypergraph block structure to enable more efficient batched matrix multiplications. This could potentially reduce the cost of the second FFN layer by up to 50\%. A significant source of overhead stems from the split representations~(\scaledmath{\mathbf{X}^\V, \mathbf{X}^\E, \mathbf{X}^\V_\adj, \mathbf{X}^\E_\adj}), which require multiple normalization and matrix multiplication steps that could be combined or parallelized. Additionally, the sparsity properties of the attention mechanism -- including diagonal symmetry in self-attention and sparsity in edge attention -- could be further leveraged through custom kernels. Moreover, the benefits from sparsity are expected to scale more effectively at higher resolutions, where the cost of attention operations grows quadratically with sequence length.

Alternatively, larger models may reduce the performance gap, as illustrated by the hypothetical HgVT-B model shown in \cref{tab:app_compeff}. The smaller gap in inference performance for HgVT-B suggests that increasing the computational workload per operation helps mitigate the relative impact of the call-graph overhead from the split representations. This indicates that scaling the model size may naturally improve computational efficiency by better amortizing fixed costs.

\FloatBarrier

\section{Hypergraph Quality}
\label{app:graph_quality}

To understand the structural quality of the generated hypergraph in HgVT, we consider how effectively it organizes features into coherent, distinct clusters. Unlike fully connected transformer architectures, HgVT uses hypergraphs to structure relationships in a way that preserves sparsity while capturing feature groupings. However, a na\"\i ve approach may achieve high-quality metrics on trivial tasks, strongly aligning with low-level features (such as textures) rather than assessing the model's ability to capture more nuanced structural qualities. We therefore propose using four key metrics -- Hyperedge Entropy, Intra-Cluster Similarity, Inter-Cluster Distance, and Silhouette Score -- to achieve a balanced assessment, while ensuring that these metrics are computationally feasible and well-defined for practical evaluation.

In the context of HgVT, a ``cluster''  corresponds to a primary hyperedge~($p\mathcal{E}$) within the hypergraph, where virtual hyperedges~($v\mathcal{E}$) are excluded due to the hierarchical graph structure. Each primary hyperedge represents a grouping of vertices~$\mathcal{V} = i\mathcal{V} \cup v\mathcal{V}$, where we primarily focus on image vertices~($i\mathcal{V}$). This approach excludes virtual vertices~($v\mathcal{V}$), which serve as summarization tokens and are expected to be largely distinct from the image vertices due to the diversity regularization. By defining clusters through primary hyperedges, we focus our evaluation on image-based feature groupings, assessing the quality of these groupings with respect to the specific properties captured by the following metrics.
\vspace{0.5em}

\begin{enumerate}[leftmargin=3\parindent]
    \item \textbf{Hyperedge Entropy (HE):} Assesses the internal diversity within clusters.
    \item \textbf{Intra-Cluster Similarity (ICS):} Measures cohesion among vertices within clusters.
    \item \textbf{Inter-Cluster Distance (ICD):} Evaluates separation between clusters.
    \item \textbf{Silhouette Score (SIL):} Provides an overall measure of clustering quality, balancing cohesion and separation.
\end{enumerate}
\vspace{0.5em}

The groupings within each primary hyperedge ($p\mathcal{E}$)  are defined with fuzzy weights derived from the soft adjacency matrix~$\mathbf{A}$, which encodes the membership strength between vertices and hyperedges. All clustering quality metrics are therefore exclusively computed on the vertex features $\mathbf{X}^\V$, where using the image subset $\mathbf{X}^{(:iV)}$ allows for direct correspondence with strong vision embeddings, such as from DINOv2~\cite{oquab2024dinov2} and CLIP~\cite{clip_paper}. Furthermore, many of the metrics can be simplified to utilize cluster centroids, resulting in more computationally efficient computations.  For a given cluster $j$, the centroid $E_{c,j}$ is calculated  as:
\begin{equation}
    E_{c,j} = \frac{\sum_{k\in\mathcal{V}}\mathbf{A}_{kj} X_k}{\sum_{k\in\mathcal{V}} \mathbf{A}_{kj}}
\end{equation}
where $X_k\in\mathbb{R}^{d}$ represents the feature vector for the $k$-th vertex in $\mathcal{V}$. This centroid formulation leverages the soft adjacency matrix $\mathbf{A}\in\mathbb{R}^{|V|\times|E|}$ to weigh each vertex’s contribution proportionally to its membership strength to the $j$-th hyperedge.

To standardize notation, we define two common functions for cosine similarity ($\mathrm{csim}$) and distance ($\mathrm{cdist}$), which are used throughout the metrics. Cosine similarity between two feature vectors $X_i$ and $X_j$ is defined as:
\begin{equation}
    \mathrm{csim}(X_i, X_j) = \frac{X_i \cdot X_j}{||X_i||\;||X_j||}
\end{equation}
where $\cdot$ represents the dot product, and $||X||$ denotes the L2 norm of $X$. Likewise, cosine distance -- used as a measure of dissimilarity -- is defined:
\begin{equation}
    \mathrm{cdist}(X_i, X_j) = 1 - \frac{X_i \cdot X_j}{||X_i||\;||X_j||}
\end{equation}

\subsection{Hyperedge Entropy}

Hyperedge Entropy (HE) measures the concentration of vertex features within each cluster (hyperedge), quantifying how~\mbox{``focused''} or homogeneous the feature distribution is within each cluster. Using entropy provides a measure of intra-cluster coherence, capturing the spread of vertex feature similarities with respect to the centroid feature for each hyperedge.

To compute HE for a given hyperedge $j$, we first calculate the cosine similarity between each vertex feature $X_i$ and the centroid feature $E_{c,j}$ of the cluster. This similarity score quantifies the alignment between individual vertex features and the core representation of the cluster. We then define $p_{ij}$ as a normalized similarity score, computed using a softmax function over these cosine similarities, limited to vertices belonging to the cluster $j$ as defined by the hard adjacency matrix $\hat{\mathbf{A}}$: 
\begin{equation}
    p_{ij} = \frac{\mathrm{exp}(\mathrm{csim}(X_i, E_{c,j}))}{\sum_{v\in \mathcal{E}_j} \mathrm{exp}(\mathrm{csim}(X_v, E_{c,j}))}
\end{equation}
where $\mathcal{E}_j$ represents the set of vertices (indexed by $v$) in the $j$-th hyperedge as defined by $\hat{\mathbf{A}}$. The entropy for each hyperedge $j$ is then calculate as:
\begin{equation}
    \mathrm{HE}_j = -\sum_{i\in\mathcal{E}_j} p_{ij} \;\mathrm{log}(p_{ij})
\end{equation}
This formulation yields an entropy distribution over the $|\mathcal{E}|$ hyperedges for a given graph, and a larger distribution when aggregated over an evaluation dataset. Here, lower entropy values indicate more concentrated, homogeneous feature distributions within the cluster, and higher entropy suggests more diverse or spread-out feature distributions.

From an interpretive standpoint, low HE values may signal that the cluster is dominated by homogeneous features, often associated with low-level structures, such as texture. For instance, in an image of a cat, a hyperedge with a low HE could indicate that fur-related features are overly concentrated, which may reflect a focus on surface-level details rather than high-level semantic structure. Conversely, a high HE can indicate poor intra-cluster coherence or semantic clustering, potentially caused by noise or irrelevant feature vectors within the cluster. Thus, balancing HE across clusters is desirable to ensure that hyperedges reflect meaningful, well-structured groupings of image features.

\subsection{Intra-Cluster Similarity}

Intra-Cluster Similarity (ICS) measures the cohesion of vertex features within each cluster (hyperedge), providing a sense of how similar the features are within each group. For each hyperedge, ICS is calculated as the average cosine similarity between each vertex feature $X_i$ and the centroid feature $E_{c,j}$ of the $j$-th hyperedge. This metric captures the internal consistency of each cluster, with higher values indicating more cohesive feature groupings.
\begin{equation}
    \mathrm{ICS}_j = \frac{1}{|\mathcal{E}_j|}\sum_{i\in\mathcal{E}_j} \mathrm{csim}(X_i, E_{c,j})
\end{equation}
where $\mathcal{E}_j$ represents the set of vertices (index by $i$) in hyperedge $j$ as defined by the hard adjacency matrix $\hat{\mathbf{A}}$. To ensure meaningful results, clusters with fewer than two vertices are omitted from this calculation, as they lack sufficient members to define intra-cluster similarity. 

\subsection{Inter-Cluster Distance}

Similar to ICS, Inter-Cluster Distance (ICD) measures how distinct different clusters (hyperedges) are from one another. Specifically, ICD quantifies the separation between clusters by measuring the cosine distance between the centroids of hyperedge pairs. This metric reflects how far apart different clusters are in feature space, with higher values indicating greater separation and, thus, more distinct feature groupings. For each pair of hyperedges $(j,k)$, ICD is computed as:
\begin{equation}
    \mathrm{ICD}_{j,k} = \mathrm{cdist}(E_{c,j}, E_{c,k})
\end{equation}
The overall ICD for the graph can then be aggregated by taking the average distance across all hyperedge pairs:
\begin{equation}
    \mathrm{ICD} = \frac{1}{|\mathcal{E}|(|\mathcal{E}| - 1)} \sum_{j\neq k} \mathrm{ICD}_{j,k}
\end{equation}

\subsection{Silhouette Score}

The Silhouette Score \cite{sil_score} combines both intra-cluster similarity (cohesion) and inter-cluster distance (separation) to provide an overall measure of the clustering quality. This score evaluates how well each vertex is clustered with respect to its assigned hyperedge and nearby clusters. For each vertex $i$ within a hyperedge $j$, two values are defined:
\vspace{0.5em}

\begin{itemize}[leftmargin=2\parindent]
    \item $a_{ij}$: the average distance between vertex $i$ and all other vertices within its assigned hyperedge $j$, computed using the soft adjacency matrix $\mathbf{A}$.
    \item $b_{ij}$: the lowest average cosine distance between vertex $i$ and all vertices in other hyperedges, effectively measuring how close $i$ is to its nearest neighboring cluster.
\end{itemize}
\vspace{0.5em}

\noindent These two values are calcualted as follows:

\begin{align}
    a_{ij} &= \frac{\sum_{v\in\mathcal{E}_j,v\neq i} \; \mathbf{A}_{vj} \; \mathrm{cdist}(X_i, X_v)}{\sum_{v\in\mathcal{E}_j,v\neq i} \; \mathbf{A}_{vj}}\\
    b_{ij} &= \min_{k,k\neq j}\frac{\sum_{v\in\mathcal{E}_k,v\neq i} \; \mathbf{A}_{vj} \; \mathrm{cdist}(X_i, X_v)}{\sum_{v\in\mathcal{E}_k,v\neq i} \; \mathbf{A}_{vj}}
\end{align}

The Silhouette Score $s_{ij}$ for the $i$-th vertex in the $j$-th hyperedge is computed as:
\begin{equation}
    s_{ij} = \frac{b_{ij} - a_{ij}}{\mathrm{max}(a_{ij}, b_{ij})}
\end{equation}
where the individual score $s_{ij}$ is bounded by $[-1, 1]$, where more positive indicates strong cluster cohesion, more negative indicates poor clustering, and zero indicates that the vertex lies on the boundary between clusters. Finally, the global Silhouette score for the graph can be computed by averaging $s_{ij}$ across all edges and vertices:
\begin{equation}
 \mathrm{SIL} = \frac{1}{|\mathcal{E}|\;|\mathcal{V}|}\sum_{j\in\mathcal{E}}\sum_{i\in\mathcal{V}} s_{ij}   
\end{equation}

The global Silhouette score omits clusters with fewer than two elements (as is standard), due to $s_{ij}$ being undefined for such pairs. From an interpretive standpoint, a higher SIL (closer to one) is ideal; however, it too can suffer from the same flaw as HE and ICS, where focus on trivial (texture) clustering result in better values, incorrectly suggesting strong clustering. Similarly, highly sparse graphs with small vertex counts per hyperedge can result in higher than expected SIL scores. For example, it is easier to form a tight cluster of two vertices than 20. We therefore suggest considering all four metrics en-aggregate, where a high SIL score is only meaningful with a high ICS, ICD, and a moderate to high HE (indicating diversity within each cluster). 

\subsection{Behavior with DINO Features}

Following the definitions of graph quality metrics, we explore how these metrics behave when HgVT-Lt’s feature representations are substituted with DINOv2 \cite{oquab2024dinov2} features of progressively richer semantic strength. This analysis serves two purposes. First, it allows us to validate our chosen graph quality metrics by observing whether they effectively capture structural differences as feature richness increases, supporting the interpretive value of these metrics within the HgVT framework. Second, it provides insight into the level of semantic detail the HgVT model's hypergraphs are focusing on, shedding light on the model’s capacity to capture and represent varying levels of semantic information.

We consider three pooling methods -- image pooling, expert pooling, and combined pooling -- within the~\mbox{HgVT-Lt} model trained on ImageNet-100. Image pooling considers only image vertices ($i\mathcal{V}$), ignoring the hypergraph structure; expert pooling incorporates hierarchical information flow through virtual hyperedges ($v\mathcal{E}$); and combined pooling integrates both approaches. 
For each configuration, we extract the hypergraphs of all ImageNet-100 validation images (totaling 5k). Specifically, we utilize the soft adjacency matrix $\mathbf{A}$ from the final layer of \mbox{HgVT-Lt} and then substitute the image vertex features $\mathbf{X}^{(:iV)}$ with the final~DINOv2 features (spanning model scales~\mbox{S, B, L, G}). Notably, the the pooling methods indirectly influence the hypergraph structure, as they primarily affect the classification head during training but subsequently affect the generated hypergraph structure through learned representations.

\begin{figure}[ht]
    \centering
    \begin{subfigure}[b]{.4\textwidth}
    \centering
    \includegraphics[width=\textwidth]{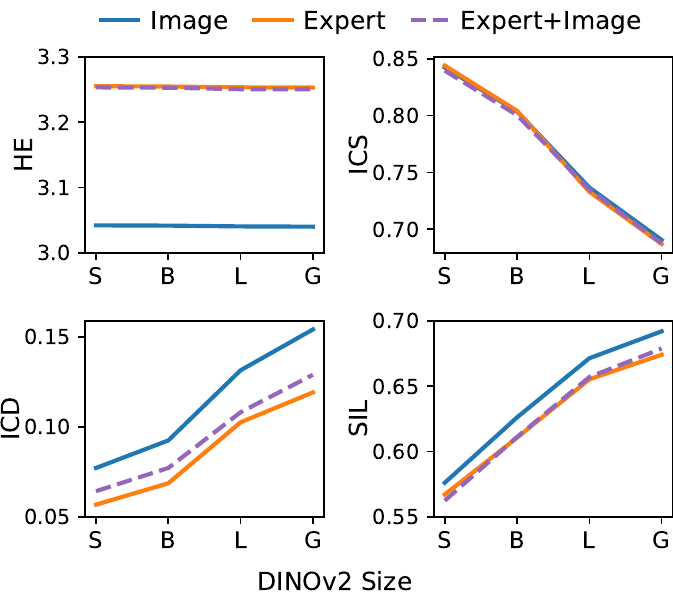}
    \caption{Metric Scaling.}
    \label{fig:dino_feat_raw}
    \end{subfigure}%
    \hspace{2em}
    \begin{subfigure}[b]{.4\textwidth}
    \centering
    \includegraphics[width=\textwidth]{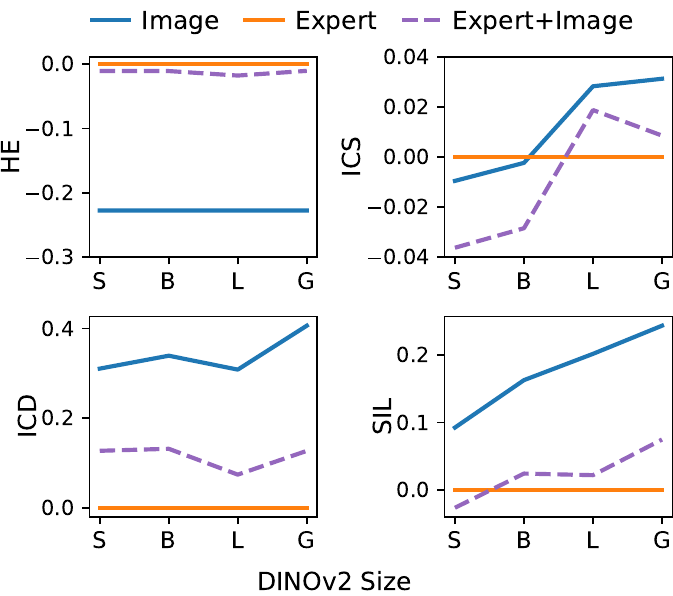}
    \caption{Cliff's Delta Metric Scaling.}
    \label{fig:dino_feat_cliff}
    \end{subfigure}%

\caption{Comparing graph quality metrics under DINOv2 feature scaling with HgVT-Lt trained on ImageNet-100. Further comparing expert, image, and combined pooling methods. Showing (a) the raw metric medians, and (b) the Cliff's D measure for the metric distributions against expert pooling as a basline. }
\end{figure}   

The spatial correspondence of both DINOv2 features and HgVT’s image vertices with the original input image allows us to substitute the original HgVT image vertex features with DINOv2 features. This alignment is well-established in applications such as object segmentation and depth estimation for DINOv2 features and verified through graph visualizations in~\cref{app:graph_visualization} for HgVT.
To preserve spatial coherence between the two models, we resize~DINO input images to~\mbox{280x280} from the original~\mbox{160x160} resolution. With a patch size of 14, this resizing yields~\mbox{20x20} image tokens, which are then aggregated using 2x2 patches to match HgVT-Lt’s 10x10 image vertex structure. For each pooling configuration, we compute the graph quality metrics across DINOv2 model scales (as shown in \cref{fig:dino_feat_raw}) and measure the effect sizes of these distributions, using expert pooling as a baseline with Cliff’s Delta for comparison (see \cref{fig:dino_feat_cliff}). Cliff’s Delta \cite{cliffs_delta} provides a non-parametric measure of effect size that quantifies the degree of separation between two distributions, with values close to $0$ indicating minimal difference and values approaching $\pm 1$ indicating strong differences in distribution. Notably, all measured distributions exhibit a statistically significant separation, as measured by a K-S test.

As DINO model size increases, all pooling methods exhibit consistent trends in clustering metrics. Hyperedge Entropy~(HE) remains stable, indicating that the overall spread of feature diversity within clusters is unaffected by feature scaling. However, Intra-Cluster Similarity~(ICS) decreases, revealing finer distinctions within existing clusters as DINO features scale. Meanwhile, Inter-Cluster Distance~(ICD) and Silhouette Score~(SIL) increase, reflecting improved separation among the fixed clusters. These trends suggest that as DINO models grow, they approach a more balanced clustering structure, similar to~HgVT’s~(with~ICS around 0.45 and ICD around 0.32). This convergence implies that HgVT may capture a level of semantic structure comparable to what would be achieved by a much larger DINO model, highlighting HgVT’s inherent efficiency in representing semantically rich information.

When comparing pooling methods, differences in clustering metrics for expert and image pooling remain mostly consistent~(when considering effect size). Image pooling yields marginally higher ICS with increasing DINO model size and noticeably higher ICD and SIL, resulting in clusters that are more cohesive and well-separated. This suggests that image pooling may focus on distinct, cohesive textures, with reduced graph inter-connectivity and cluster overlap. 
Expert pooling, by contrast, exhibits higher HE and lower ICD, indicating that clusters are more internally diverse and less distinctly separated. In this case, omitting the image vertices during classification allows for increased graph connectivity, which is reflected by a degradation of clustering metrics. Finally, the combined pooling method aligns closely with expert pooling, while recovering a slight improvement to ICD and SIL due to the direct inclusion of image vertices during classification.

\FloatBarrier
\section{Hypergraph Representations}
\label{app:representations}

In this section, we explore the spatial organization of feature representations using Uniform Manifold Approximation and Projection (UMAP) visualizations~\cite{umap_paper}, generated from the HgVT-Lt model on the ImageNet-100 validation set. UMAP enables a comparative analysis of how different components within the hypergraph structure distribute features in their learned latent space.  By reducing dimensionality to two components, UMAP highlights the spatial clustering of graph feature vectors, extracted from the model's final layer. To address varying group sizes ($i\mathcal{V}$, $v\mathcal{V}$, $p\mathcal{E}$, $v\mathcal{E}$), we standardize each plot’s sample size to the minimum group size, randomly sampling from other groups as needed to ensure consistency.

\subsection{Full Graph Feature Representations}

\begin{figure}[ht]
    \centering
    \begin{subfigure}[b]{.32\textwidth}
    \centering
    \includegraphics[width=\textwidth]{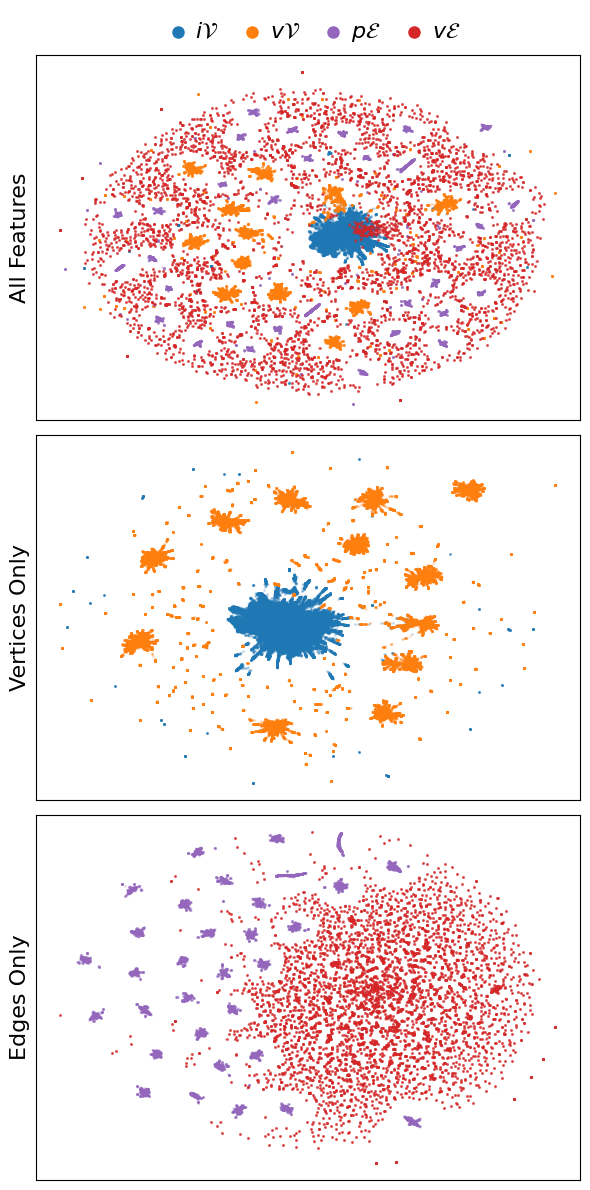}
    \caption{Expert Pooling.}
    \end{subfigure}%
    \hfill
    \begin{subfigure}[b]{.32\textwidth}
    \centering
    \includegraphics[width=\textwidth]{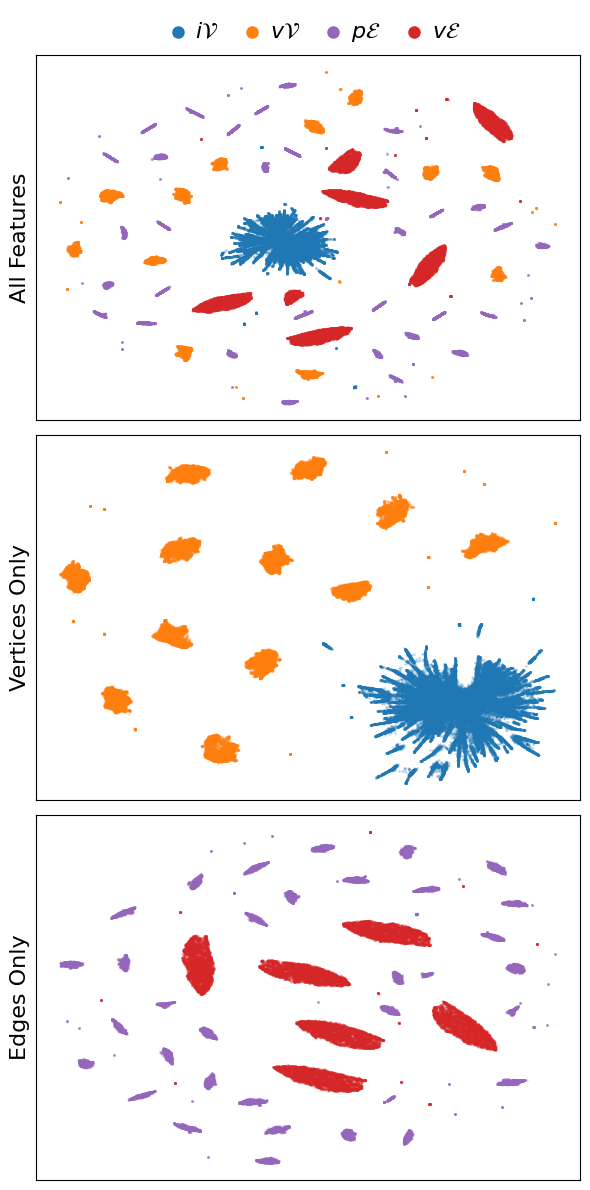}
    \caption{Image Pooling.}
    \end{subfigure}
    \hfill
    \begin{subfigure}[b]{.32\textwidth}
    \centering
    \includegraphics[width=\textwidth]{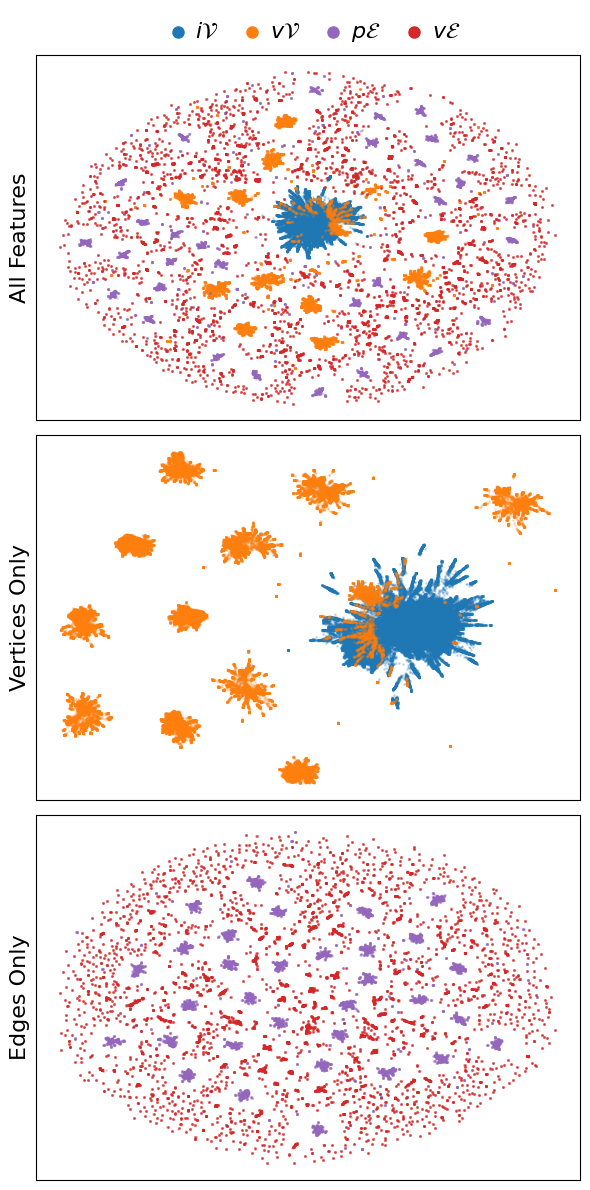}
    \caption{Expert + Image Pooling.}
    \end{subfigure}%
    
\caption{UMAP plots of the HgVT-Lt model under different pooling methodings: (a) Expert pooling, (b) Image pooling, and (c) both Expert and Image pooling. Showing image vertices ($i\mathcal{V}$), 12 virtual vertices ($v\mathcal{V}$), 32 primary hypereges ($p\mathcal{E}$), and 6 virtual hyperedges ($v\mathcal{E}$).}
\label{fig:app_umap_all}
\end{figure}

We explore the full graph feature representations by considering all features ($\mathcal{V}\cup\mathcal{E}$), only vertices ($\mathcal{V}$), and only edges ($\mathcal{E}$) across three pooling methods: expert pooling, image pooling, and a combined approach. The UMAP representations shown in~\cref{fig:app_umap_all} utilize a nearest neighbors setting of 10 and a minimum distance of 0.1, with consistent seeds for reproducibility.

From the UMAP results, we observe a distinct separation between expert and image pooling, with the combined method exhibiting characteristics of both. In all cases, image vertices~($i\mathcal{E}$) form relatively tight clusters, typically surrounded by other feature categories. Distinct clusters are evident for virtual vertices~($v\mathcal{V}$) and primary hyperedges~($p\mathcal{E}$), with 12~($|v\mathcal{V}|$) and~32~($|p\mathcal{E}|$) clusters, respectively.  Under image pooling, virtual hyperedges~($v\mathcal{E}$) form six~($|v\mathcal{E}|$) distinct groups, likely due to the absence of model incentives to leverage these features for classification. In contrast, in the expert and combined pooling cases, virtual hyperedges appear as diffuse clouds, suggesting strong interconnectivity with virtual vertices and primary hyperedges. 

For expert pooling, virtual hyperedges show overlap with image vertices, a phenomenon absent in the combined pooling case. This overlap likely represents low-level image features that must be transmitted through virtual hyperedges in the expert pooling scenario, whereas in the combined case, they can be transmitted directly through pooled image features. Additionally, we observe diffuse overlap of virtual vertices with image vertices in expert pooling, replaced by a single overlapping virtual vertex in the combined case. This distinction suggests two possible strategies for supporting lower-level image features: either a shared overlap across virtual vertices or a single dedicated virtual vertex providing feature support. Overall, the UMAP results align with the findings from the previous section.

\FloatBarrier
\subsection{Expert Pooling Feature Representations}

\begin{figure}[ht]
    \centering
    \begin{subfigure}[b]{.4\textwidth}
    \centering
    \includegraphics[width=\textwidth]{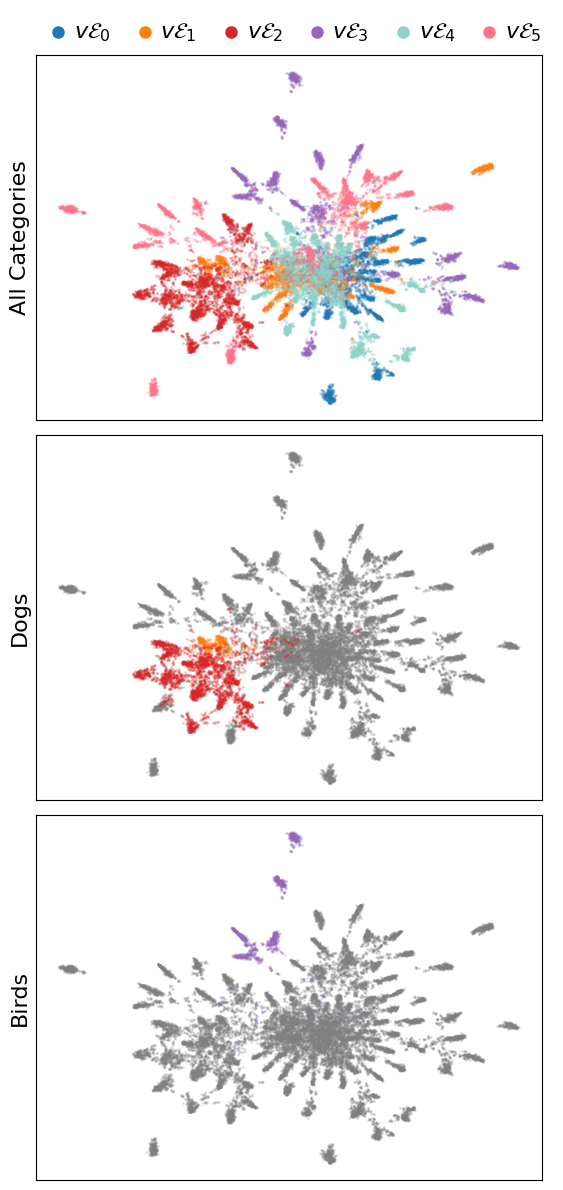}
    \caption{Expert Pooling.}
    \end{subfigure}%
    \hspace{2em}
    \begin{subfigure}[b]{.4\textwidth}
    \centering
    \includegraphics[width=\textwidth]{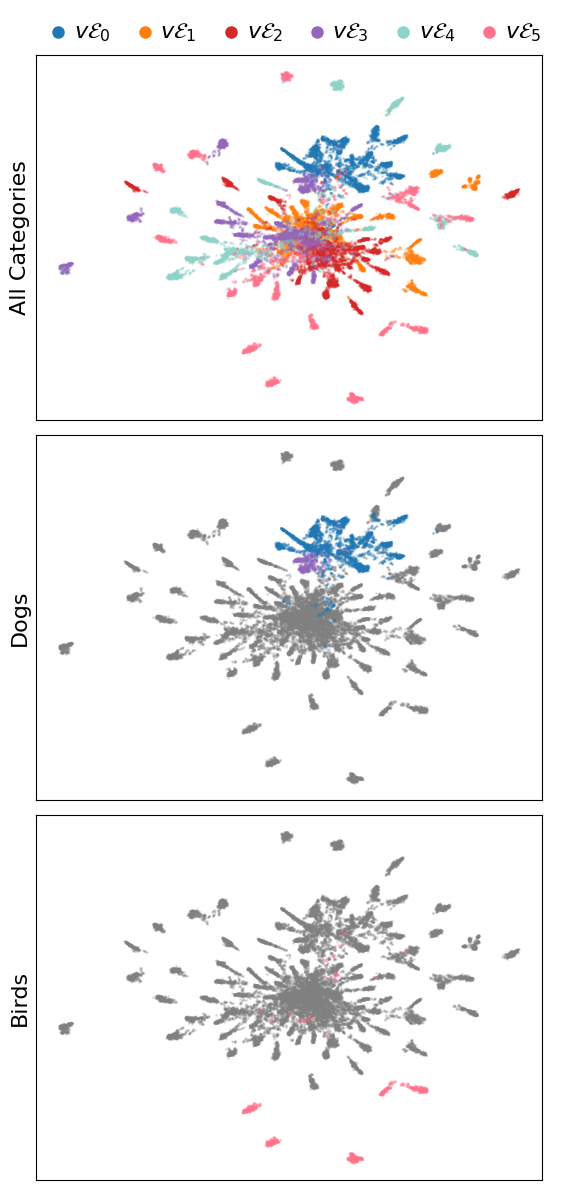}
    \caption{Expert + Image Pooling.}
    \end{subfigure}%

\caption{UMAP plots of virtual hyperedge classification allocation for the HgVT-Lt model under different pooling methodings: (a) Expert pooling, (b) both Expert and Image pooling. Showing overall expert allocation $v\mathcal{E}_i$, and select ImageNet-100 macro-classes: Dogs and Birds.}
\label{fig:app_umap_experts}
\end{figure}

Given the clustering behavior for virtual edges ($v\mathcal{E}$) observed in the previous section, we further examine their structure when plotted independently to determine if unique patterns emerge.
Specifically, we assess whether this structure correlates with specific experts (edge IDs) or macro-classess, such as Dogs and Birds in ImageNet-100, considering both the expert pooling and combined cases. Due to the diffuse nature of this feature type, we increase the nearest neighbors setting to 120 and set the minimum distance to 0.5 for clearer clustering in \cref{fig:app_umap_experts}.

The clustering of edge IDs suggests that specific edges capture both overlapping and distinct aspects of the feature space, with each cluster representing shared or distinct features specialized for certain macro-classes. This behavior is validated when considering the clusters corresponding to the dog macro-class, emerging in both the expert and combined pooling cases. In contrast, when considering birds, they consistently form a less compact cluster, occupying a unique sub-region with minimal interference from other categories. Notably, bird features are more tightly clustered in the expert pooling case, while in the combined pooling case, bird features are more dispersed, with some overlapping with the center. This increased spread in the combined case likely reflects the distributed influence of expert edges, which only partially contribute to the final clusters, whereas the expert-only case preserves more focused class-specific features. Additionally, we observe that birds consistently align with a single expert ID, while dogs are associated with no more than two expert IDs. This allocation pattern is further analyzed in \cref{app:macro_prediction}.

\FloatBarrier
\section{Graph Visualization}
\label{app:graph_visualization}

Visualizing the hypergraph structure in HgVT provides crucial insights into how various components -- such as virtual vertices, primary hyperedges, and image vertices -- interact to inform predictions. However, given the complexity of hypergraphs and the dense interconnections across vertices (nodes) and edges, a straightforward visualization would be overwhelming and challenging to interpret.  
To address this, we apply a pruned projection method that represents the hypergraph in  "slices,"  focusing on key relationships while filtering out less influential components. This approach balances interpretability with structural fidelity, offering a clearer view of the hypergraph’s hierarchical organization.

\begin{figure}[ht]
  \centering
  \includegraphics[width=0.5\textwidth]{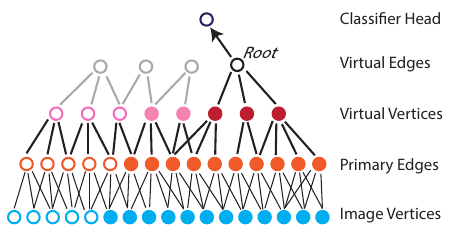}
  \caption{Example hypergraph structure used for visualization. Showing the four distinct feature types and the subset selection (top-1; root node) expert pooling used for classification - unused virtual edges are shown in light gray. Showing direct (0-hop; red) and indirect (1-hop; pink) virtual vertices, along with their membership primary hyperedges (orange), and the associated image vertices (blue). Features omitted in the graph visualizations are shown with open circles. Notably, a primary edge may be duplicated if it belongs to multiple virtual vertices.}
  \label{fig:app_graph_struc}
\end{figure}

In this method,  we begin by selecting the top-1 (most confident) virtual edge as the root node. From this root, we identify and rank the connected virtual vertices (vNodes) using the soft adjacency matrix $\mathbf{A}$, selecting those with contributions above a threshold of 0.1. For each vNode, we identify the top-H primary hyperedges (pEdges) and treat each as an individual slice in the visualization. Some pEdges appear in the top-H of multiple vNodes, enabling the visualization to capture overlapping and shared feature pathways effectively. Each pEdge is visualized as a 2D image, with patch dimming based on contribution intensity (no dimming for the highest contributions, maximal dimming for zero contributions). Finally, we add secondary virtual vertices linked to the primary hyperedges, further enriching each slice’s representation by showing indirect (1-hop) influences. \cref{fig:app_graph_struc} provides a graphical depiction of this hierarchical structure, illustrating the direct and indirect virtual vertices and the connecting elements, indicating which components are plotted or excluded due to the pruned slice mechanism.

The following figures present graph visualizations that highlight the autosegmentation properties and hierarchical feature localization within the hypergraph structure, with distinct regions corresponding to features like eyes and feet. Notably, these visualizations are derived solely from the adjacency matrix rather than attention layers, though they exhibit structural properties similar to what one might expect from attention visualizations. This demonstrates that the adjacency relationships within the hypergraph capture meaningful spatial and semantic organization independently of the attention mechanisms.

\begin{figure}[ht]
    \centering
    \begin{subfigure}[b]{.48\textwidth}
    \centering
    \includegraphics[width=\textwidth]{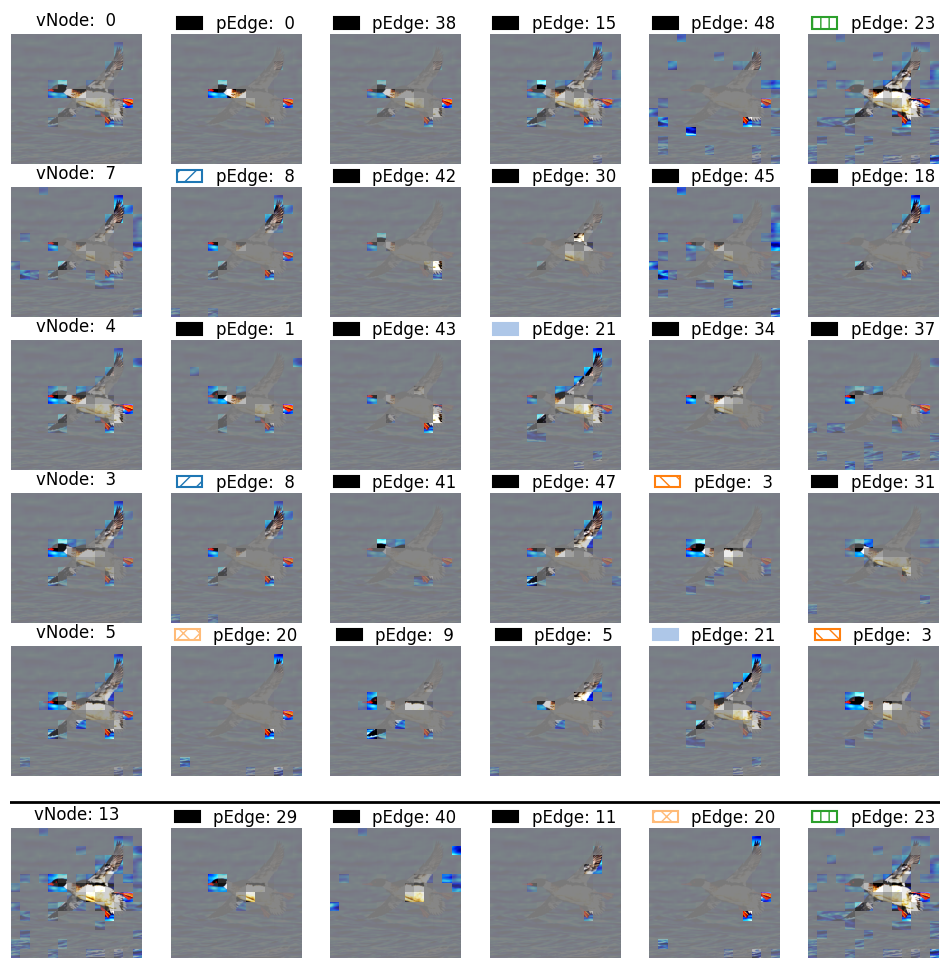}
    \caption{Class label=``Mergus serrator'' (98).}
    \end{subfigure}%
    \hfill
    \begin{subfigure}[b]{.48\textwidth}
    \centering
    \includegraphics[width=\textwidth]{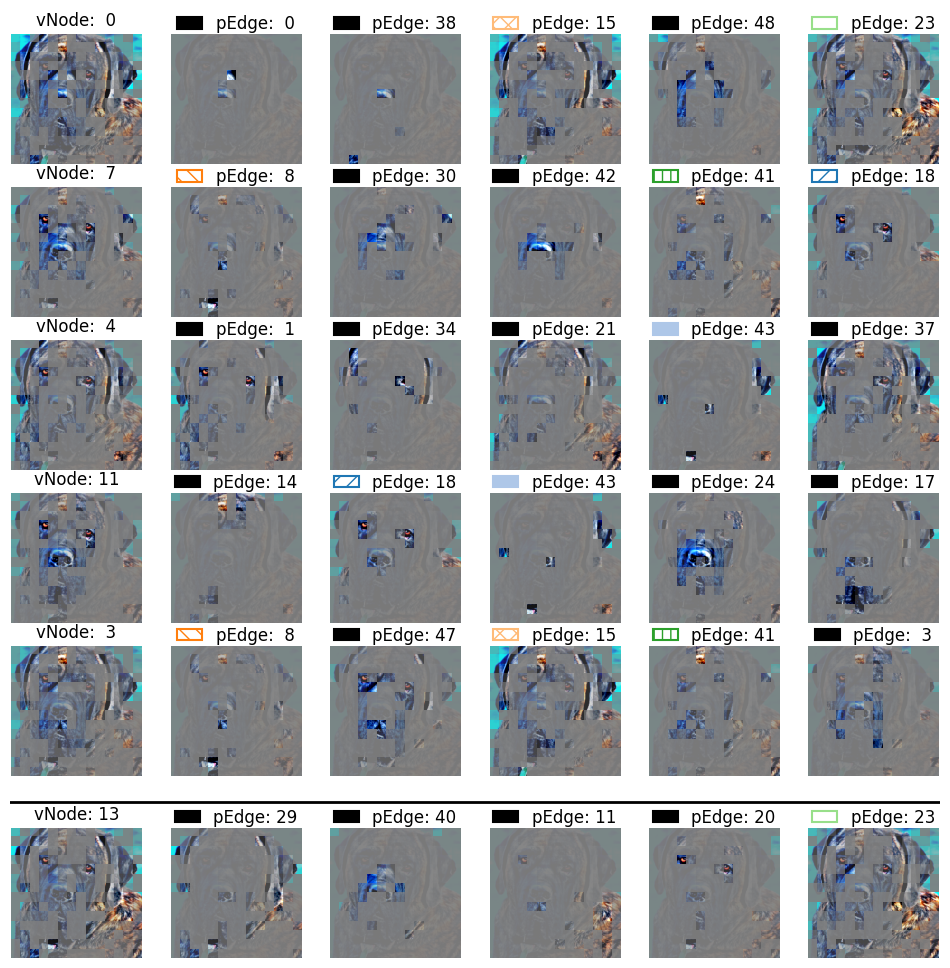}
    \caption{Class label=``Bull mastiff'' (243).}
    \end{subfigure}%

\caption{Graph visualizations from the HgVT-Ti model trained on ImageNet-1k, using samples from the ImageNet validation set. Showing top-5 direct virtual vertices and their top-5 highest contributing primary hyperedges above the horizontal line; top-1 indirect virtual vertex and its primary hyperedges below. Leftmost column shows aggregated summary of all primary hyperedges; remaining columns show individual primary hyperedges. Shared primary hyperedges are marked with unique identifier boxes; a black rectangle indicates no duplicates.}
\end{figure}

\begin{figure}[ht]
    \centering
    \begin{subfigure}[b]{.48\textwidth}
    \centering
    \includegraphics[width=\textwidth]{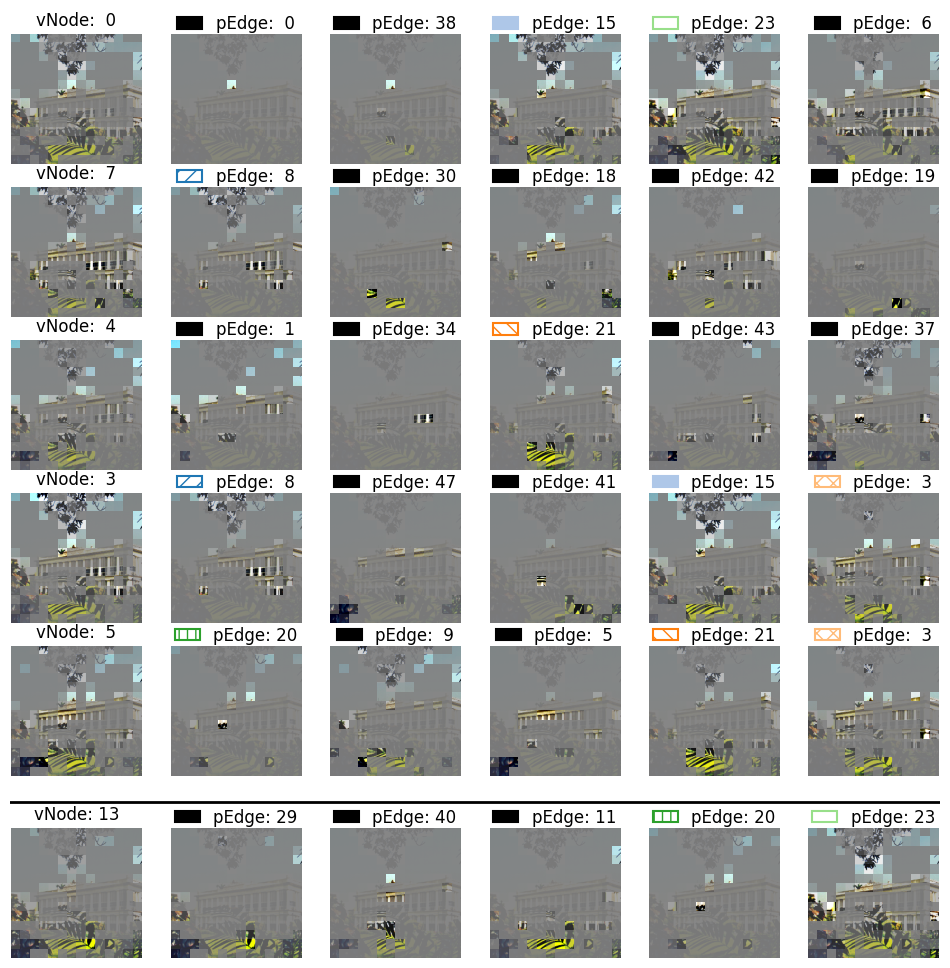}
    \caption{Class label=``Palace'' (698).}
    \end{subfigure}%
    \hfill
    \begin{subfigure}[b]{.48\textwidth}
    \centering
    \includegraphics[width=\textwidth]{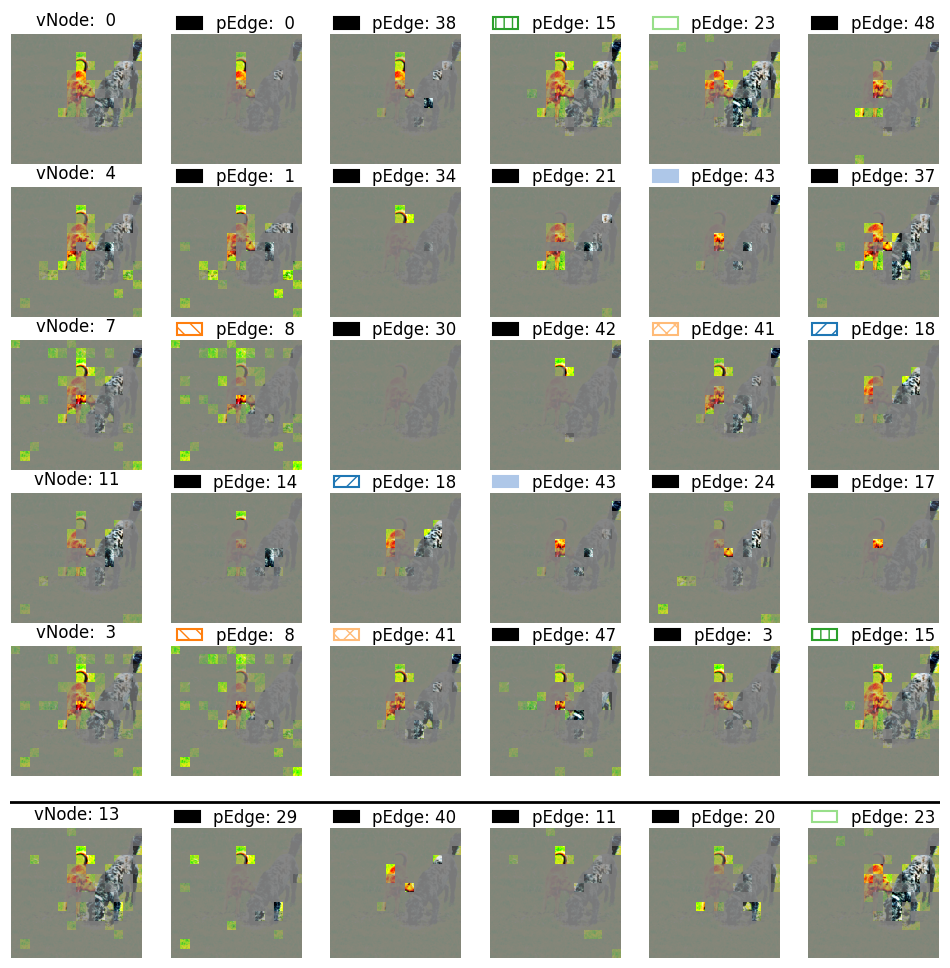}
    \caption{Class label=``Irish terrier'' (184).}
    \end{subfigure}%
    \vfill
    \begin{subfigure}[b]{.48\textwidth}
    \centering
    \includegraphics[width=\textwidth]{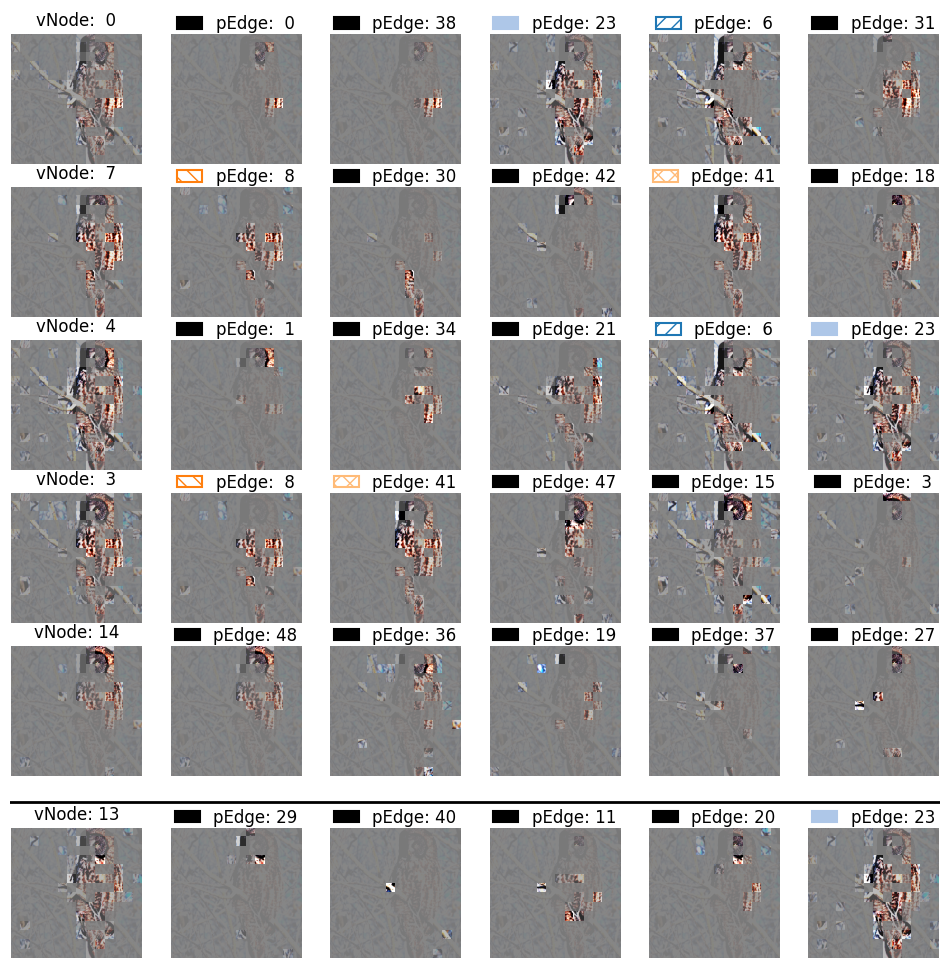}
    \caption{Class label=``Great grey owl'' (24).}
    \end{subfigure}%
    \hfill
    \begin{subfigure}[b]{.48\textwidth}
    \centering
    \includegraphics[width=\textwidth]{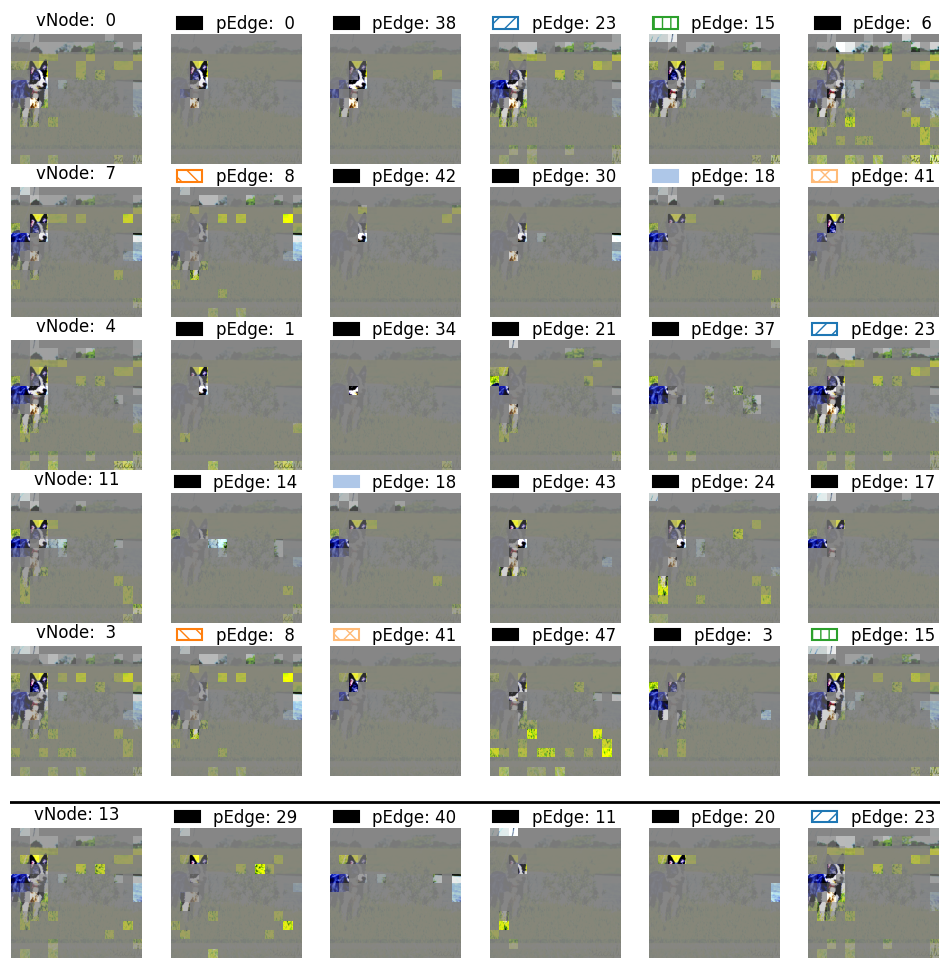}
    \caption{Class label=``Border collie'' (232).}
    \end{subfigure}%

\caption{Graph visualizations from the HgVT-Ti model trained on ImageNet-1k, using samples from the ImageNet validation set. Showing top-5 direct virtual vertices and their top-5 highest contributing primary hyperedges above the horizontal line; top-1 indirect virtual vertex and its primary hyperedges below. Leftmost column shows aggregated summary of all primary hyperedges; remaining columns show individual primary hyperedges. Shared primary hyperedges are marked with unique identifier boxes; a black rectangle indicates no duplicates.}
\end{figure}    

\FloatBarrier
\clearpage
\section{Semantic Segmentation}

In this section, we evaluate the performance of HgVT on the dense prediction task of semantic segmentation. Given the transformer backbone, we adopt the training protocol proposed in DINOv2 \cite{oquab2024dinov2}, which involves an initial finetuning phase at higher input resolutions on ImageNet-1k with positional embedding interpolation, followed by freezing the backbone and training segmentation heads. Final segmentation is then performed by merging overlapping ``stencil'' predictions at the segmentation training resolution (i.e. 512x512). Notably, freezing the backbone deviates from standard semantic segmentation training protocols. This is due to the fact that semantic segmentation relies heavily on spatial features (image vertices), and there is no straightforward gradient pathway for the hyperedge features, thereby preventing effective full-backbone finetuning.

\subsection{Resolution Finetuning}

To bridge the gap between pretraining and dense prediction tasks, we perform resolution finetuning, a process where the model is further trained on ImageNet-1k at a higher input resolution. While DINOv2 employs a resolution finetuning strategy at $416^2$ for 10k steps using a cosine annealing learning rate schedule, we adopt a more lightweight approach inspired by TransNeXt~\cite{Shi2023TransNeXtRF}. Specifically, we finetune the model at a resolution of $384^2$ for 5 epochs using a constant learning rate of 1e-5.

Additionally, to maintain consistent sparsity in the hypergraph representations at the higher resolution, we adjust the maximum population regularization value ($\beta$) to $|\mathcal{V}|/4$, where $|\mathcal{V}|$ is the number of vertices. This adjustment ensures that the model’s structural regularization scales appropriately with the increased resolution. All other training hyperparameters (including data augmentation) remain identical to those used during the initial pretraining phase.

\begin{figure}[ht]
    \centering
    \begin{minipage}{0.56\linewidth}
        \centering
        \captionof{table}{Ablations on resolution finetuning HgVT-S.}
        \label{tab:app_resolution_ablation}
        \vspace{2em}
        \resizebox{1.0\textwidth}{!}{
        \begin{tabular}{c | c c c | c c | c c}
         & 
        \multirow{2}{*}[1.0em]{\rhead{FT. Res.}} & 
        \multirow{2}{*}[2.0em]{\rhead{Interp. PEs}} & 
        \multirow{2}{*}[2.7em]{\rhead{Pop Max ($\beta$)}} &
        \multicolumn{2}{c|}{Top-1} &
        \multicolumn{2}{c}{Sparsity} \\ 
        \cmidrule(lr){5-8} Method  & & & & 
        $224^2$ & $512^2$ & $224^2$ & $512^2$ \\
        \toprule
        Baseline & -- & -- & $1/6\cdot|\mathcal{V}_{224}|$ & 
        81.20 & 78.59 & 0.637 & 0.750\\
        A0 & $224^2$ & -- & $1/4\cdot|\mathcal{V}_{224}|$ & 
        80.95 & 77.84 & 0.603 & 0.721\\
        \midrule
        B0 & $384^2$ & \No & $1/6\cdot|\mathcal{V}_{224}|$ & 
        78.24 & 81.13 & 0.875 & 0.951\\
        B1 & $384^2$ & \Yes & $1/6\cdot|\mathcal{V}_{224}|$ & 
        78.84 & 81.54 & 0.875 & 0.951\\
        \midrule
        C0 & $384^2$ & \Yes & $1/6\cdot|\mathcal{V}_{384}|$ & 
        80.03 & 82.26 & 0.611 & 0.764\\
        C1 & $384^2$ & \Yes & $1/4\cdot|\mathcal{V}_{384}|$ & 
        80.11 & 82.32 & 0.592 & 0.743\\
        C2 & $384^2$ & \Yes & $1/3\cdot|\mathcal{V}_{384}|$ & 
        80.01 & 82.28 & 0.619 & 0.771\\
        \bottomrule
        \end{tabular}
        }
    \end{minipage}
    \hfill
    \begin{minipage}{0.42\linewidth}
        \centering
        \includegraphics[width=\linewidth]{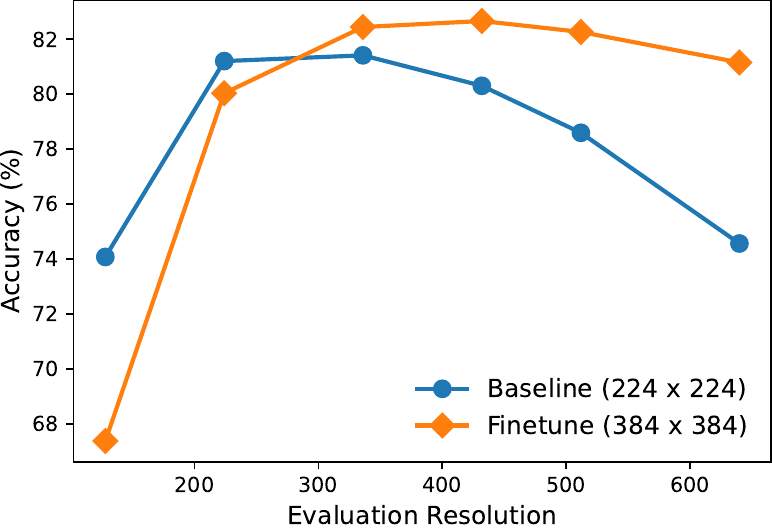}
        \vspace{-2em}
        \caption{Top-1 ImageNet-1k accuracy for HgVT-S before and after resolution finetuning.}
        \label{fig:app_resolution_accuracy}
    \end{minipage}
\end{figure}

In \cref{tab:app_resolution_ablation}, we present an ablation study evaluating the impact of interpolating versus reinitializing positional embeddings, as well as the effect of varying the maximum population regularization value ($\beta$). We find that interpolating positional embeddings leads to better performance, while increasing $\beta$ helps prevent over-sparsification, with $\beta=\frac{1}{4}|\mathcal{V}|$ yielding the best results at higher resolutions. Interestingly, this setting slightly degrades performance when maintaining the original training resolution, suggesting that the benefits of a larger population regularization are resolution-dependent. \cref{fig:app_resolution_accuracy} shows the Top-1 ImageNet accuracy across resolutions for the baseline HgVT-S and method C1, revealing trends that are remarkably consistent with the resolution finetuning behavior observed in DINOv2 \cite{oquab2024dinov2}.

\FloatBarrier
\subsection{Segmentation Results}

\begin{figure}[ht]
    \centering
    \begin{minipage}{0.3\linewidth}
        \centering
        \captionof{table}{Segmentation Hyperparameters.}
        \label{tab:app_seg_hyperparams}
        \vspace{-0.5em}
        \resizebox{1.0\textwidth}{!}{
      \begin{tabular}{l c}
        \toprule
        \textbf{Parameter} & \textbf{Value} \\
        \midrule
        Train Resolution & $512\times512$ \\
        Global Batch Size & 16 \\
        \midrule
        Schedule & Poly \\
        Power & 1.0 \\
        Total Steps &  40k\\
        Warmup Steps & 1.5k \\
        \midrule
        Optimizer & AdamW \\
        Peak LR & 1e-3 \\
        Weight Decay & 1e-4 \\
        $(\beta_1,\beta_2)$ & (0.9, 0.999) \\
        \midrule 
        Resize Ratio & $0.5-2.0$ \\
        Augmentations & \makecell{Random Crop, Flip, \\ Photometric} \\
        
        \bottomrule
      \end{tabular}
        }
    \end{minipage}
    \hfill
    \begin{minipage}{0.69\linewidth}
        \centering
        \includegraphics[width=\linewidth]{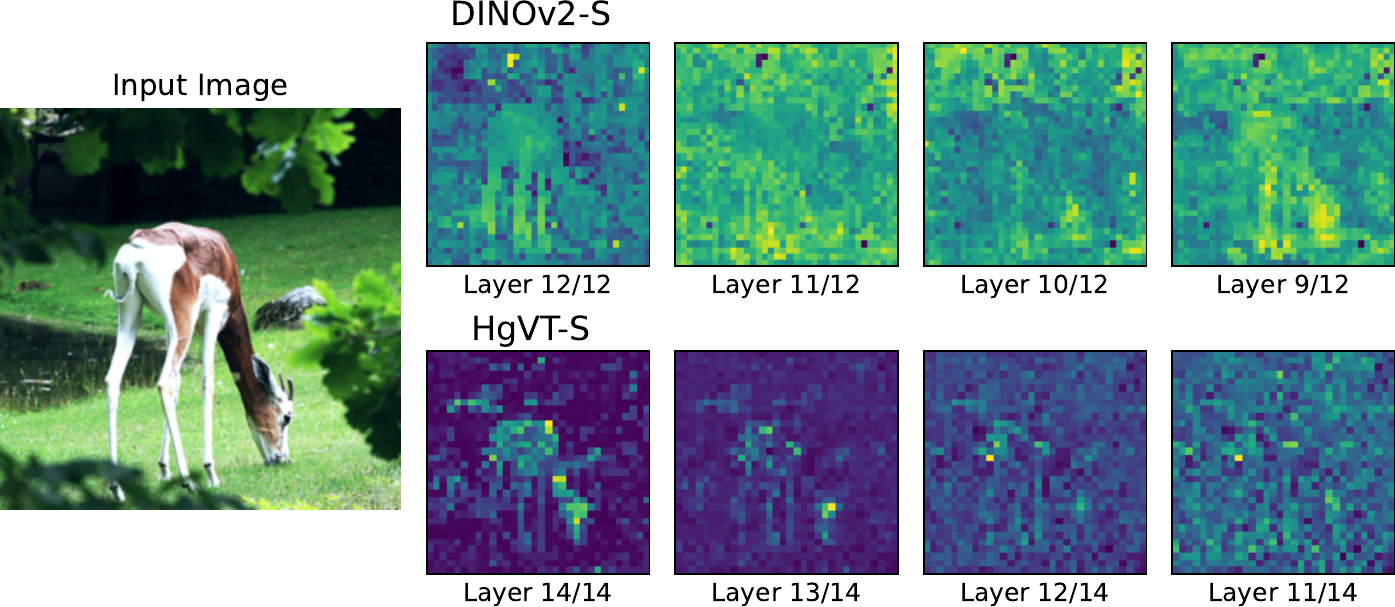}
        \vspace{-1.5em}
        \caption{Comparison of DINOv2-S (top) and HgVT-S (bottom) spatial features for the last 4 layers in each network. Plotting the per-token L2 norm to visualize the HgVT feature sparsity.}
        \label{fig:app_feature_sparsity}
    \end{minipage}
\end{figure}

Following the finetuning phase, we train segmentation heads on top of the frozen backbone, following the protocol used by DINOv2, with training hyperparameters summarized in \cref{tab:app_seg_hyperparams}. We evaluate performance on the ADE20k \cite{ade20k}, CityScapes \cite{citscapes_dataset}, and PASCAL VOC \cite{voc_dataset} datasets. To better understand the feature representations learned by HgVT, we compare the L2 feature norms of the last four layers of HgVT-S and DINOv2-S for an example image, as shown in \cref{fig:app_feature_sparsity}. Notably, HgVT exhibits significantly sparser feature activations compared to DINOv2. This suggests that relying solely on the final feature layer may limit segmentation performance, leaving gaps in otherwise contiguous regions.


Given the uncertain behavior of the sparse feature activations, we explore several segmentation head architectures to assess the effectiveness of each in decoding the sparse image vertex features. 

\begin{itemize}[leftmargin=2\parindent]
    \item \textbf{Linear Head:} A simple linear projection following a batch normalization layer as used in DINOv2.
    \item \textbf{MLP Head:} A two-layer perceptron following Linear-BN-SiLU-Linear.
    \item \textbf{Conv-MLP Head:} Similar to the MLP head but with a 3x3 convolution as the input layer.
    \item \textbf{Pyramid Pooling Module (PPM):} Module proposed by PSPNet \cite{PSPNet_paper}, which utilizes multi-level pooling for isotropic input features. 
    \item \textbf{Upsampled PPM Head (PPMU):} An enhanced PPM implementation which uses a 2x up-sampling step with pixel shuffle before the final MLP. 
\end{itemize}


\begin{table}[h!]
    \caption{
    Semantic Segmentation results on ADE20k using the frozen HgVT-S backbone. Head method includes input configuration: \texttt{-1}~last backbone layer only, \texttt{-4}~last four backbone layers concatenated. Showing mIoU (\%) and Pixel Accuracy (\%) where available.
    $^*$Our evaluation.
    $^\dagger$Results from \texttt{github.com/CSAILVision/semantic-segmentation-pytorch}. 
    $^\ddagger$Results from DINOv2~\cite{oquab2024dinov2}.} \label{tab:app_seg_results}
    \centering
    \vspace{-0.5em}
    \resizebox{1.0\textwidth}{!}{
    \begin{tabular}{c c c | c c c | c c | c c  | c c }
    \toprule
    \multicolumn{3}{c|}{Backbone} &
    \multicolumn{3}{c|}{Head} &
    \multicolumn{2}{c|}{ADE20k} &
    \multicolumn{2}{c|}{CityScapes} &
    \multicolumn{2}{c}{PASCAL VOC} \\
    \midrule
    Method &
    Size &  
    Frozen &
    Method &
    Size & 
    Multiscale &
    mIoU &
    Acc. &
    mIoU &
    Acc. &
    mIoU &
    Acc. \\
    \midrule


    Swin-Ti \cite{swintransformer} & 28.3M & \No & 
    UperNet \cite{upernet} & 60M & \Yes &
    46.1 & -- &
    -- & -- &
    -- & -- \\

    TransNeXt-Ti \cite{Shi2023TransNeXtRF} & 28.2M & \No &
    UperNet \cite{upernet} & 59M & \No &
    51.1 & -- &
    -- & -- &
    -- & -- \\

    TransNeXt-Ti \cite{Shi2023TransNeXtRF} & 28.2M & \No &
    UperNet \cite{upernet} & 59M & \Yes &
    51.7 & -- &
     -- & -- &
    -- & -- \\

    TransNeXt-Ti \cite{Shi2023TransNeXtRF} & 28.2M & \No &
    Mask2Former \cite{mask2former} & 47.5M & \No &
    53.4 & -- &
    -- & -- &
    -- & -- \\

    \midrule

    ResNet-18 \cite{resnet_paper} & 11.5M & \No &
    PPM-1 & 12.9M & \No &
    \hspace{0.5em}33.8$^\dagger$ & \hspace{0.5em}76.1$^\dagger$ &
    -- & -- &
    -- & -- \\

    ResNet-50 \cite{resnet_paper} & 25.6M & \No &
    PPM-1 & 23.2M & \No &
    \hspace{0.5em}41.3$^\dagger$ & \hspace{0.5em}79.7$^\dagger$ &
    -- & -- &
    -- & -- \\

    ResNet-101 \cite{resnet_paper} & 44.5M & \No &
    PPM-1 & 23.2M & \No &
    \hspace{0.5em}42.2$^\dagger$ & \hspace{0.5em}80.6$^\dagger$ &
    78.4 & -- & 
    82.6 & -- \\

    DINOv2-S/14 \cite{oquab2024dinov2} & 22.1M & \Yes &
    Linear-1 & 59.3k & \No &
    44.3 & \hspace{0.4em}79.5$^*$ &
    66.6 & -- &
    81.1 & \hspace{0.5em}95.9$^*$ \\

    DINOv2-S/14 \cite{oquab2024dinov2} & 22.1M & \Yes &
    Linear-4 & 237k & \No &
    \hspace{0.3em}46.0$^*$ & \hspace{0.5em}80.1$^*$ &
    -- & -- &
    \hspace{0.5em}81.8$^*$ & \hspace{0.5em}96.0$^*$ \\

    DINOv2-S/14 \cite{oquab2024dinov2} & 22.1M & \Yes &
    Linear-4 & 237k & \Yes &
    47.2 & -- &
    77.1 & -- &
    82.6 & -- \\

    DINOv2-G/14 \cite{oquab2024dinov2} & 1.10B & \Yes &
    Linear-1 & 237k & \No &
    49.0 & -- &
    71.3 & -- &
    83.0 & -- \\


    OpenCLIP-G/14 \cite{open_clip_paper} & 1.01B & \Yes &
    Linear-1 & 214k & \No &
    \hspace{0.5em}39.3$^\ddagger$ & -- &
    \hspace{0.5em}60.3$^\ddagger$ & -- &
    \hspace{0.5em}71.4$^\ddagger$ & -- \\

    \midrule
    HgVT-S/16  & 22.9M & \Yes &
    Linear-1 & 34.6k & \No &
    12.0 & 43.3 &
    30.2 & 72.9 &
    34.0 & 81.4 \\

    HgVT-S/16  & 22.9M & \Yes &
    Linear-4 & 138k & \No &
    26.7 & 68.5 &
    52.4 & 89.3 &
    66.7 & 91.7 \\

    HgVT-S/16  & 22.9M & \Yes &
    MLP-4 & 235k & \No &
    28.5 & 71.8 &
    58.0 & 91.7 &
    72.9 & 93.6 \\

    HgVT-S/16  & 22.9M & \Yes &
    ConvMLP-4 & 1.84M & \No &
    33.5 & 74.3 &
    64.5 & 93.1 &
    76.1 & 94.4 \\


    HgVT-S/16 & 22.9M & \Yes &
    PPM-4 & 15.5M & \No &
    36.0 & 75.7 &
    68.0 & 93.8 &
    77.9 & 94.9 \\

    HgVT-S/16  & 22.9M & \Yes &
    PPMU-4 & 17.4M & \No &
    37.6 & 76.4 &
    69.8 & 94.3 &
    79.0 & 95.1 \\

    
    \bottomrule
    \end{tabular}
    }
\end{table}

Segmentation results are shown in \cref{tab:app_seg_results}. Consistent with the feature norm analysis, the Linear Head underperforms, particularly when applied solely to the final feature layer. To investigate this further, we also evaluate a linear head that combines features from the last four backbone layers (consistent with the multiscale method in DINOv2). While this approach improves performance compared to using only the final layer, it still falls short of more complex architectures. This suggests that deeper features mitigate some of the sparsity effects observed in \cref{fig:app_feature_sparsity}, while linear projections alone are insufficient for fully decoding the hypergraph representations. While the more complex PPMU method achieves an mIoU of 37.6\% on ADE20K, it falls short of both DINOv2-S and state-of-the-art methods. 

In contrast, results on CityScapes and PASCAL VOC are stronger, with the PPMU heads closing the gap on PASCAL VOC and surpassing DINOv2 Linear-1 classifier on CityScapes. Notably, all convolution-based methods outperform OpenCLIP-G on these two datasets, suggesting that (1) the poor ADE20K results are partially attributable to class confusion and (2) the sparse features result in discontinuous regions, which degrade segmentation performance. The convolution-based methods help smooth out these discontinuities, improving overall performance. Additionally, the reduced class count (20 vs. 150) likely mitigates class confusion, contributing to stronger performance on CityScapes and PASCAL VOC.

\begin{figure}[t]
    \centering
    \includegraphics[width=0.9\linewidth]{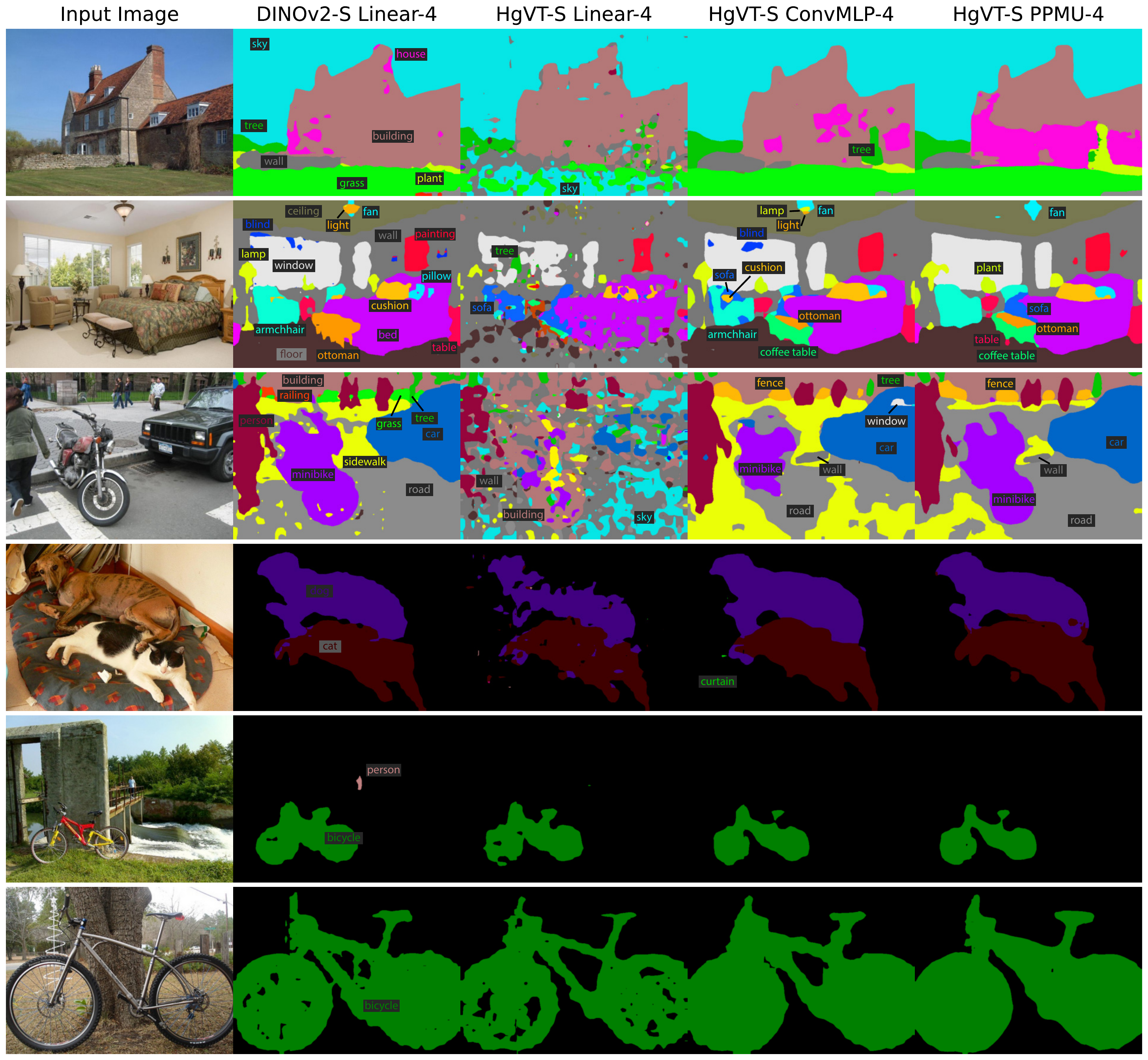}
    \vspace{-0.5em}
    \caption{Semantic Segmentation Visualization. Showing examples from ADE20k (top) and PASCAL VOC (bottom).}
    \label{fig:app_hyperseg}
    \vspace{-1.5em}
\end{figure}

We attribute this low performance to several factors. First, the lack of backbone fine-tuning leads to object class confusion, where similar classes (e.g., cushion and pillow) that were not targeted during ImageNet-1k training are incorrectly assigned. Second, the high degree of feature sparsity encouraged by population regularization may result in localization errors, where objects are not encoded at the correct pixel location. As supporting evidence, we measure a 4.3\% lower mIoU and 2.7\% lower pixel accuracy on ADE20k when using a Linear-4 head with configuration B1 in \cref{tab:app_resolution_ablation}. Third, the patch size of 16×16 pixels further reduces segmentation localization compared to the more commonly used 8×8 down-sampling. Notably, DINOv2 uses a 14×14 patch size, self-supervised learning, and ImageNet-22k pretraining, resulting in denser features (see \cref{fig:app_feature_sparsity}), which likely accounts for part of the performance gap. Finally, a large amount of information --including the hyperedge features and virtual nodes -- is not directly used in semantic prediction due to the lack of direct spatial alignment. Leveraging these additional features may improve boundary detection and class distinction, highlighting areas for future exploration.

The segmentation visualizations in \cref{fig:app_hyperseg} align with these findings. The linear head on HgVT produces discontinuous segmentation regions, whereas the convolution-based methods help fill these gaps. The PPMU head appears to over-smooth the results, leading to missed fine details (e.g., the bedroom and bicycle in the second and last rows). In certain PASCAL VOC examples, HgVT with a linear head outperforms DINOv2, where increased sparsity results in more well-defined segmentation regions (e.g., the large bicycle image in the last row). Finally, class confusion can be seen in the bedroom scene (second row), where the top of an ottoman is correctly identified while the bottom is misclassified as a coffee table. This supports our hypothesis that object class confusion is occurring and may partially explain the poor ADE20K performance.

\subsection{Using Semantic Segmentation for Interpretability}


Aside from benchmark evaluation, the linear segmentation results also provide insight into how the model encodes information. Large contiguous regions are sparsely represented by the correct class, with gaps filled by high-frequency or default classes (e.g., wall and sky). This suggests that the model assigns the correct class to a small subset of vertices, efficiently summarizing the local structure rather than encoding them uniformly. The hypergraph visualization results in \cref{app:graph_visualization} support this interpretation, showing that regions like water, grass, and sky are not contiguously covered but instead exhibit sparse coverage. This pattern may be analogous to a dithering effect used to represent continuous shading with binary values. A convolution operation would efficiently reconstruct the full structure by locally propagating this summarized information, which is supported by the ConvMLP results.

\FloatBarrier

\section{Image Retrieval}
\label{app:retrieval}

This section expands upon the image retrieval description in the main paper to provide additional implementation details and supporting evidence. Our image retrieval framework is structured around two primary first-pass search methods: pooled similarity and volumetric similarity. Both methods leverage the pooled embedding, which serves as the input to the classifier head and integrates both pooled image features and expert edge features.
\vspace{0.5em}

\begin{itemize}[leftmargin=2\parindent]
    \item \textbf{Pooled Similarity (PS):} This method computes similarity scores by comparing the pooled embeddings through a cosine similarity metric. The pooled embedding serves as a generalized representation of each image, aligning with standard vector-based similarity searches, making it both effective and efficient as a first-pass retrieval approach.
    \item \textbf{Volumetric Similarity (VS):} Unlike pooled similarity, volumetric similarity incorporates the hypergraph structure by treating the pooled embedding as a centroid. Similarity is determined using an approximate Mahalanobis distance, which accounts for the distributional spread around the centroid based on a subset of primary hyperedges. This approach captures overlap with less prominent, yet relevant, features, enabling a spatially-aware similarity measure that aligns more closely with nuanced structural characteristics.
\end{itemize}
\vspace{0.5em}

Individually, both methods perform effectively as first-pass search strategies; however, to further harness the structure of the hypergraph, we introduce an adaptive reranking phase. This phase refines retrieval results by re-evaluating similarity across a short list of top $R$ candidates, using a more detailed hypergraph similarity measure. The adaptive reranking can be applied to each of the first-pass methods, resulting in Adaptive Volumetric Similarity (AVS) and Adaptive Pooled Similarity (APS). By capitalizing on the hierarchical and relational information embedded within the hypergraph, these adaptive methods enhance retrieval precision beyond the initial search.

\subsection{Graph Pruning}

For methods that leverage the hypergraph structure, we employ a pruned graph representation based on primary hyperedge features ($p\mathcal{E}$). The pruning process begins by selecting the top-1 expert edge as the root, which serves as the initial focus for identifying key structural components. From this root, we identify the top $M$ virtual vertices that contribute most significantly to the expert edge. For each of these $M$ virtual vertices, we further select the top $N$ primary hyperedges connected to it. This yields a total of $M\times N$ hyperedge features, where we choose $M=3$, $N=4$, and $M\times N=12$ to prove a balanced between representation coverage and computational efficiency. Finally, the selected hyperedge features are deduplicated and ranked based on their overall contribution to the final prediction. Notably, this process is very similar to the slice visualization described in \cref{app:graph_visualization}, and illustrated in \cref{fig:app_graph_struc}.

\subsection{Volumetric Similarity}

Volumetric similarity leverages the hypergraph structure by treating the pooled embedding $x$ of each image as a centroid, with the pruned primary hyperedges defining a spread around this centroid. Each of the two distributions can then be represented by a centroid and covariance matrix, $(x_1, \Sigma_1)$ and $(x_2,\Sigma_2)$.  We then quantify the similarity between these distributions using the Mahalanobis distance with a combined covariance matrix, capturing both the central positions and spreads of the distributions to measure their overlap.
\begin{equation}
    d_M(x_1, x_2)^2 = \sqrt{(x_1 - x_2)^T \left(\frac{\mathbf{\Sigma}_1 + \mathbf{\Sigma}_2}{2}\right)^{-1}(x_1 - x_2)}
\end{equation}
\noindent where $\mathbf{\Sigma} = \frac{\mathbf{\Sigma}_1 + \mathbf{\Sigma}_2}{2}$ is the average covariance matrix between the two distributions.  To reduce computational complexity, we approximate $\mathbf{\Sigma}$ as a diagonal matrix, assuming minimal covariance between features. Each feature's combined variance simplifies to $\sigma^2 = (\sigma_1^2 + \sigma_2^2)/2$, yielding:
\begin{equation}
    d_M(x_1, x_2)^2 \approx \sum_{i}\frac{(x_{1,i} - x_{2,i})^2}{\sigma_i^2}
\end{equation}
where $\sigma_i^2$ represents the average variance of the $i$-th feature across the two distributions. 

While the diagonal approximation reduces complexity, calculating $1/\sigma_i^2$ for each feature remains computationally demanding. To further optimize, we approximate each variance term $\sigma_i^2 \approx \bar{\sigma}^2 + \delta_i^2$, where $\bar{\sigma}^2$ is the mean variance across features, and $\delta_i^2$ represents the deviation from this mean. Finally, we can then express $1/\sigma_i^2$ using a Taylor series expansion:
\begin{equation}
    \frac{1}{\sigma_i^2} \approx \frac{1}{\bar{\sigma}^2}( 1 - \eta_i + \eta_i^2 + \dots),\quad \eta_i = \frac{\delta^2_i}{\bar{\sigma}^2}
\end{equation}

By truncating this expansion after the first few terms, we achieve an efficient approximation for the Mahalanobis, requiring at most a single division per comparison:
\begin{equation}
    d_M(x_1, x_2)^2 \approx \rho \sum_{i}(x_{1,i} - x_{2,i})^2(1 - (\rho\cdot\delta_i^2) + (\rho\cdot\delta_i^2)^2),\quad \rho=\frac{1}{\bar{\sigma}^2}
\end{equation}
This approach allows for efficient computation that remains relatively close to the simpler cosine similarity measure, while also capturing greater variance introduced by the hypergraph structure. 

In practical terms, while these approximations do not hold universally, the deviation is small enough that the simplified form remains effective for our retrieval framework. When truncating the Taylor series to the first-order approximation the term~\mbox{$(1 - \rho\cdot \delta_i^2)$} must be clamped to a positive value, as large deviations in certain elements can cause this term to become negative, violating the mathematical definition of variance. Notably, this clamping is unnecessary for the second-order approximation, where additional terms sufficiently stabilize the variance without requiring this constraint.

\subsection{Adaptive Reranking}

The adaptive reranking process refines the initial retrieval results by re-evaluating a short list of top $R$ entries selected through one of the first-pass similarity methods. For each of these $R$ entries, we perform a graph-based similarity search, focusing on the pruned primary hyperedge features of each graph. The similarity is computed as the average distance between corresponding primary hyperedges in the query and candidate graphs.

While effective, this approach can be computationally expensive, requiring $\mathcal{O}(R \cdot(M\times N)^2)$ operations, where $M$ and $N$ represent the number of virtual vertices and primary hyperedges, respectively. However, the diversity regularization applied during training ensures minimal overlap between comparisons, resulting in a sparse correlation matrix with mostly zero similarities. This sparsity leads to redundant computations, making the process well-suited for optimization through hash-based acceleration.

To take advantage of this sparsity, we employ a centroid-based hashing mechanism, which reduces the number of necessary comparisons. Specifically, we learn a set of $H$ centroids that define $H$ distinct bins, with each primary hyperedge feature in the pruned graphs (both query and candidate) hashed into these bins. By limiting comparisons to features within the same bin, and only considering the top $C$ most relevant comparisons (defined by the query graph), we can reduce the overall complexity to $\mathcal{O}(R\cdot C)$. This approach enables adaptive reranking to achieve higher precision with significantly reduced computational costs, leveraging the sparse structure introduced by hypergraph regularization.

\subsection{Centroid Hashing}

To implement the hashing mechanism described in adaptive reranking, we learn a set of $H$ centroids that define bins for efficient similarity comparisons. Empirically, we find that setting $H=10$ provides effective separation when $M\times N = 12$, balancing coverage with computational efficiency.

These centroids are trained using the Adam optimizer over a dataset created from all pruned primary hyperedge features across the test dataset. The optimization objective involves minimizing the distance to the closest centroid while maximizing the distance to all other centroids, thereby ensuring distinct and well-separated bins. Additionally, we incorporate the same density regularization term applied to the expert edges, promoting a broader feature spread within each centroid bin. The combined loss function is thus:
\begin{equation}
\mathcal{L}_\mathrm{centroid} =  || y - c_{n1} ||^2 -  \lambda_{ICD} \cdot || y - c_{n2} ||^2 + \lambda_{DEN} \cdot \mathrm{den}(c_{n1})
\end{equation}
where $y$ is the input feature vector, $c_{n1}$ and $c_{n2}$ are the nearest and second nearest centroids, $\lambda$ is a loss weight factor, and $\mathrm{den}(\cdot)$ is the density regularization term computed over the batch. This objective minimizes the distance of each feature to its nearest centroid, while enforcing a margin with the second-closest centroid. The regularization term further ensures that centroids remain well-utilized across the feature space.

Emperically, we find that a learning rate of $4\times10^{-3}$ works well for a batch size of 512, setting $\lambda_{ICD}=0.1$ and $\lambda_{DEN}=0.5$. In practice, centroid training converges rapidly, requiring only two epochs on larger datasets such as ImageNet and CIFAR. For smaller datasets, such as Oxford and Paris, training requires approximately eight epochs. 

\FloatBarrier

\subsection{Retrieval Hyperparameter Ablations}

\begin{figure}[ht]
    \centering
    \begin{subfigure}[b]{.30\textwidth}
    \centering
    \includegraphics[width=\textwidth]
    {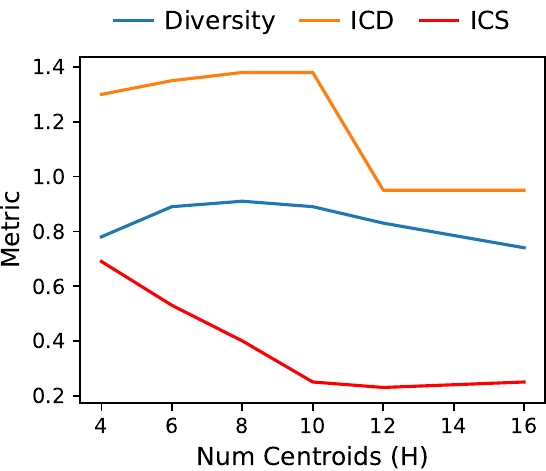}
    \caption{Hash Centroid Scaling.}
    \label{fig:retreival_hyper_centroids}
    \end{subfigure}%
    \hspace{6em}
    \begin{subfigure}[b]{.45\textwidth}
    \centering
    \includegraphics[width=\textwidth]
    {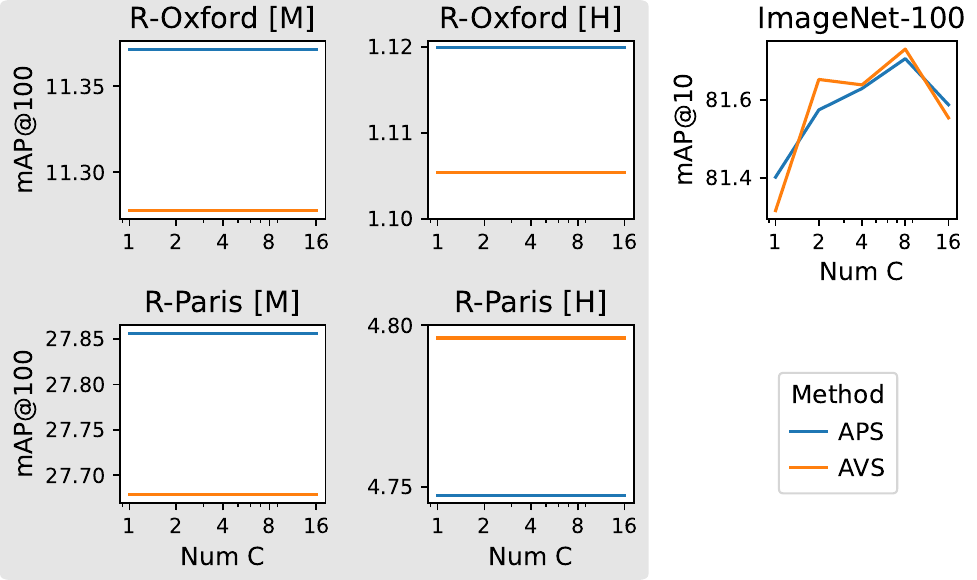}
    \caption{Graph similarity Comparison Scaling.}
    \label{fig:retreival_hyper_compares}
    \end{subfigure}%

\caption{Hyperparameter scaling behavior for adaptive re-rank method. (a) impact of hash bin clustering metrics as a function of centroid count on CIFAR-100; (b) retrieval performance as a function of graph similarity comparisons for: (left) Oxford and Paris (right) and KNN retrieval on ImageNet-100 with HgVT-Lt. Notably, Oxford and Paris are insensitive, likely due to reduced feature diversity from landmarks. }
\label{fig:retreival_hyper}
\end{figure}

\begin{figure}[ht]
    \centering
    \begin{subfigure}[b]{.43\textwidth}
    \centering
    \includegraphics[width=\textwidth]
    {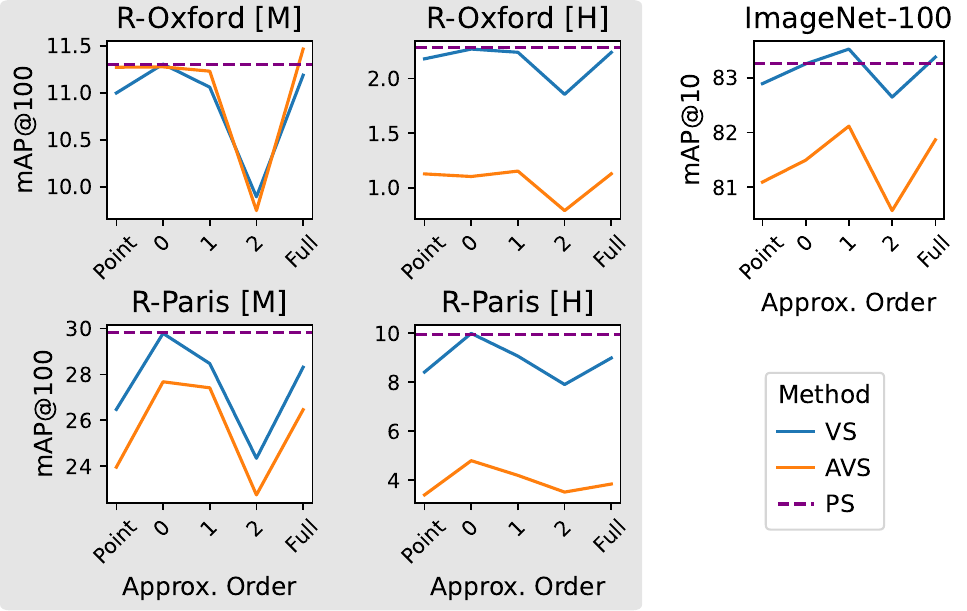}
    \caption{Mahalanobis Order.}
    \label{fig:retreival_hyper2_maha}
    \end{subfigure}%
    \hspace{4em}
    \begin{subfigure}[b]{.45\textwidth}
    \centering
    \includegraphics[width=\textwidth]
    {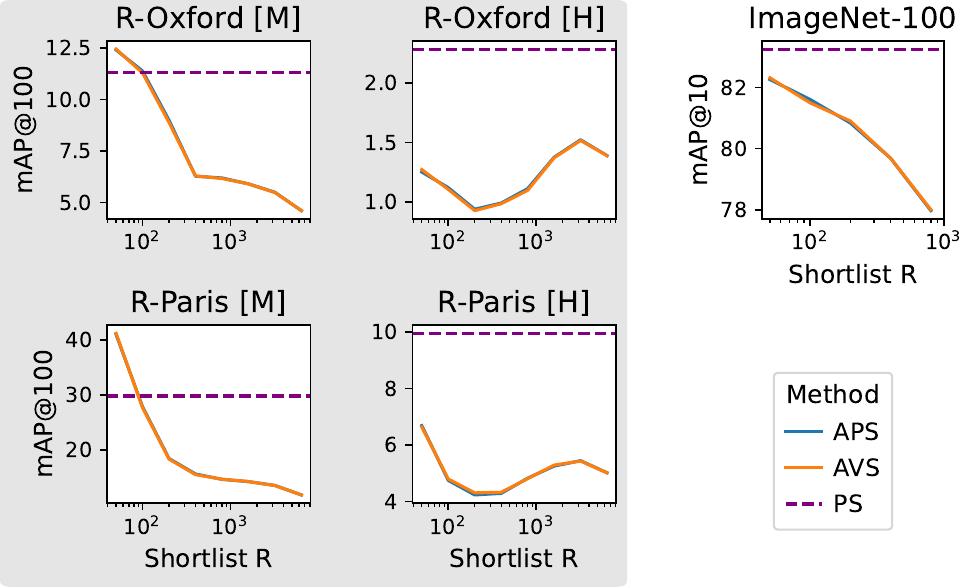}
    \caption{Re-rank short-list size.}
    \label{fig:retreival_hyper2_shortlist}
    \end{subfigure}%

\caption{Hyperparameter scaling behavior for adaptive re-rank methods with HgVT-Lt trained on ImageNet-100. (a) impact of Mahalanobis approximation order, showing point-wise, 0th, 1st, 2nd order, and full ($N=\infty$); (b) impact of short-list size $R$ on metrics. In both figures: (left) mAP retrieval on Oxford and Paris (right) and KNN retrieval for ImageNet-100. Also showing baseline using pooled similarity (PS) as horizontal purple dashed line. }
\label{fig:retreival_hyper2}
\end{figure}

We evaluate the influence of four critical hyperparameters on retrieval performance: the number of centroids $H$, the number of graph similarity comparisons $C$, the Mahalanobis approximation order, and the shortlist rank $R$ used in adaptive reranking. Results for $H$ and $C$ are presented in \cref{fig:retreival_hyper}, while \cref{fig:retreival_hyper2} highlights the effects of the Mahalanobis approximation order and shortlist rank. 

\textbf{Effect of Centroid Count:} \cref{fig:retreival_hyper_centroids} presents the relationship between $H$ and final centroid training metrics. Namely we use a diversity measure (1.0 represents uniform distribution among centroids), inter-cluster distance (ICD), and intra-cluster similarity (ICS) for the HgVT-Mu model trained on CIFAR-100. Increasing $H$ improves all metrics up to a point, followed by a degradation due over granularization. We find that $H=10$ achieves the best result for the chosen graph configuration ($N=3$ virtual vertices, $M=4$ primary hyperedges), with a notable drop in ICD at $H\ge12 = N\times M$. This choice allows the bins to remain distinct enough to provide adequate separation, while also providing sufficient overlap with an expectation value of $1.2$ hyperedges per bin.

\textbf{Effect of Comparison Count:} \cref{fig:retreival_hyper_compares} illustrates the performance of varying $C$ for the HgVT-Lt model, trained on ImageNet-100, across different retrieval benchmarks. While the Oxford and Paris datasets exhibit insensitivity to $C$, potentially due to their dependence on salient features emphasized by the diverse ImageNet-100 set, ImageNet-100 retrieval performance peaks at $C=8$. For computational efficiency, $C=4$ is selected as a trade off, maintaining comparable mAP@10 performance while requiring fewer similarity comparisons. 

\textbf{Effect of Mahalanobis Approximation Order:} \cref{fig:retreival_hyper2_maha} examines the impact of the Mahalanobis approximation order on volumetric similarity performance. Several configurations are evaluated, including point-wise approximation (where the query variance is set to 0 and the full candidate variance is precomputed as $1/\sigma_i^2$), as well as 0th, 1st, and 2nd order Taylor series approximations, and the full computation of $1/(\sigma^2_{1,i} + \sigma^2_{2,i})$. Results indicate that the 0th order approximation consistently achieves the best performance, balancing accuracy and efficiency by leveraging only $\bar{\sigma}^2$.  Conversely, the 2nd order approximation fails across all cases, likely due to instability from the $(\delta_i^2)^2$ term becoming larger than 1, causing the approximation to break down. These findings suggest that the simpler 0th order approach is both effective and computationally optimal for volumetric similarity.

\textbf{Effect of Shortlist Size:} \cref{fig:retreival_hyper2_shortlist}  explores the effect of shortlist size $R$ on adaptive metric performance. Across all methods, performance degrades as $R$ increases, driven by confusion in the graph similarity metric, which becomes more susceptible to distraction by sub-salient features. Despite this trend, a shortlist size of $R=100$ strikes a suitable balance, limiting significant distractions while maintaining enough candidates to sufficiently approximate the full mAP metric, which favors smaller $k$-rank evaluations (mAP@$k$).

\FloatBarrier

\subsection{Visualizing Adaptive Reranking}

This section provides a visual analysis of the adaptive reranking process using the Oxford dataset, demonstrating how structural similarities in hypergraphs influence retrieval precision.

\begin{figure}
    \centering

\tabskip=0pt
\valign{#\cr
  \hbox{%
    \setcounter{subfigure}{0}
    \begin{subfigure}{.46\textwidth}
    \centering
    \includegraphics[width=\textwidth]{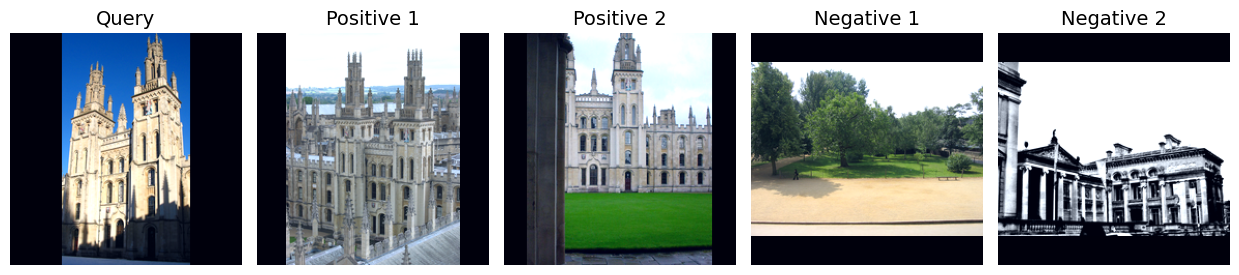}
    \caption{Query and Test Images.}
    \end{subfigure}%
  }\vfill
  \hbox{%
    \setcounter{subfigure}{1}
    \begin{subfigure}{.46\textwidth}
    \centering
    \includegraphics[width=\textwidth]{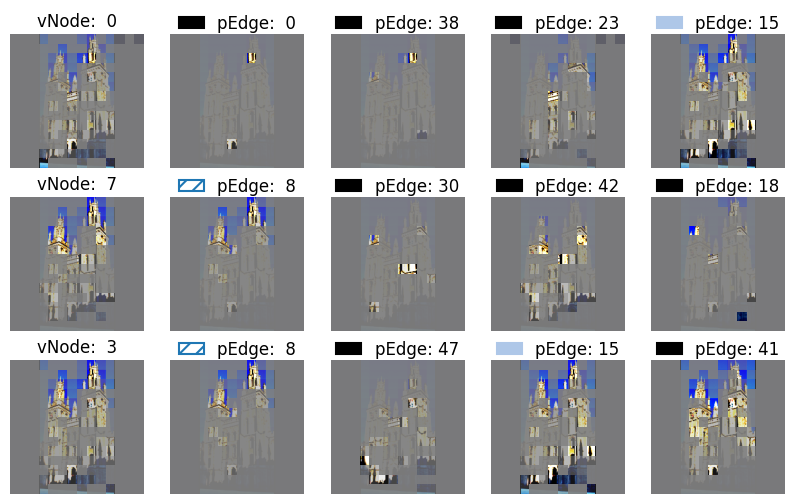}
    \caption{Pruned Query Hypergraph.}
    \end{subfigure}%
  }\vfill
  \hbox{%
    \setcounter{subfigure}{3}
    \begin{subfigure}{.46\textwidth}
    \centering
    \includegraphics[width=\textwidth]{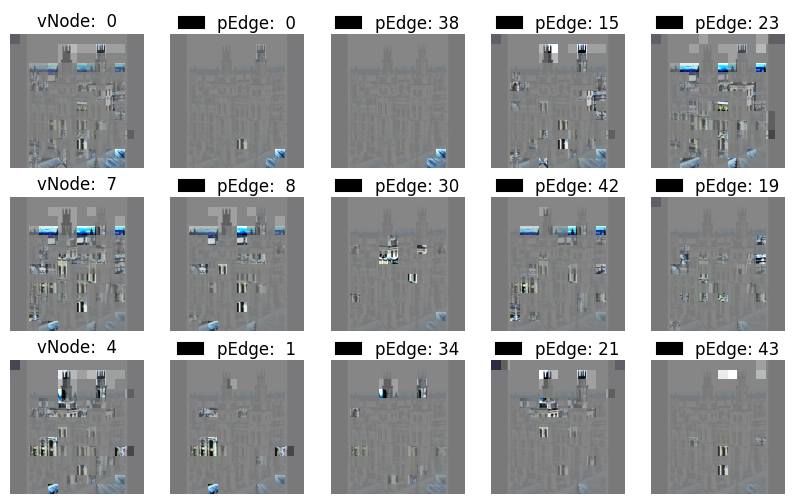}
    \caption{Pruned Positive 1 Hypergraph.}
    \end{subfigure}%
  }\vfill
  \hbox{%
    \setcounter{subfigure}{5}
    \begin{subfigure}{.46\textwidth}
    \centering
    \includegraphics[width=\textwidth]{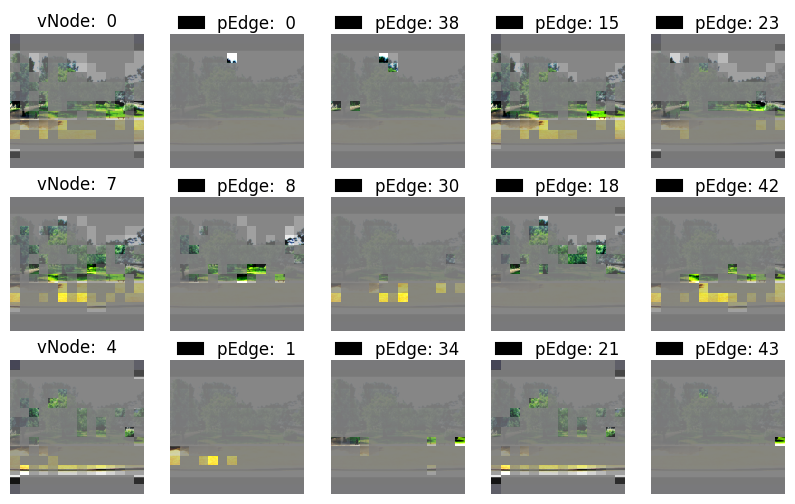}
    \caption{Pruned Negative 1 Hypergraph.}
    \end{subfigure}%
  }\cr
  \noalign{\hfill}
  \hbox{%
    \setcounter{subfigure}{2}
    \begin{subfigure}[b]{.48\textwidth}
    \centering
    \includegraphics[width=\textwidth]{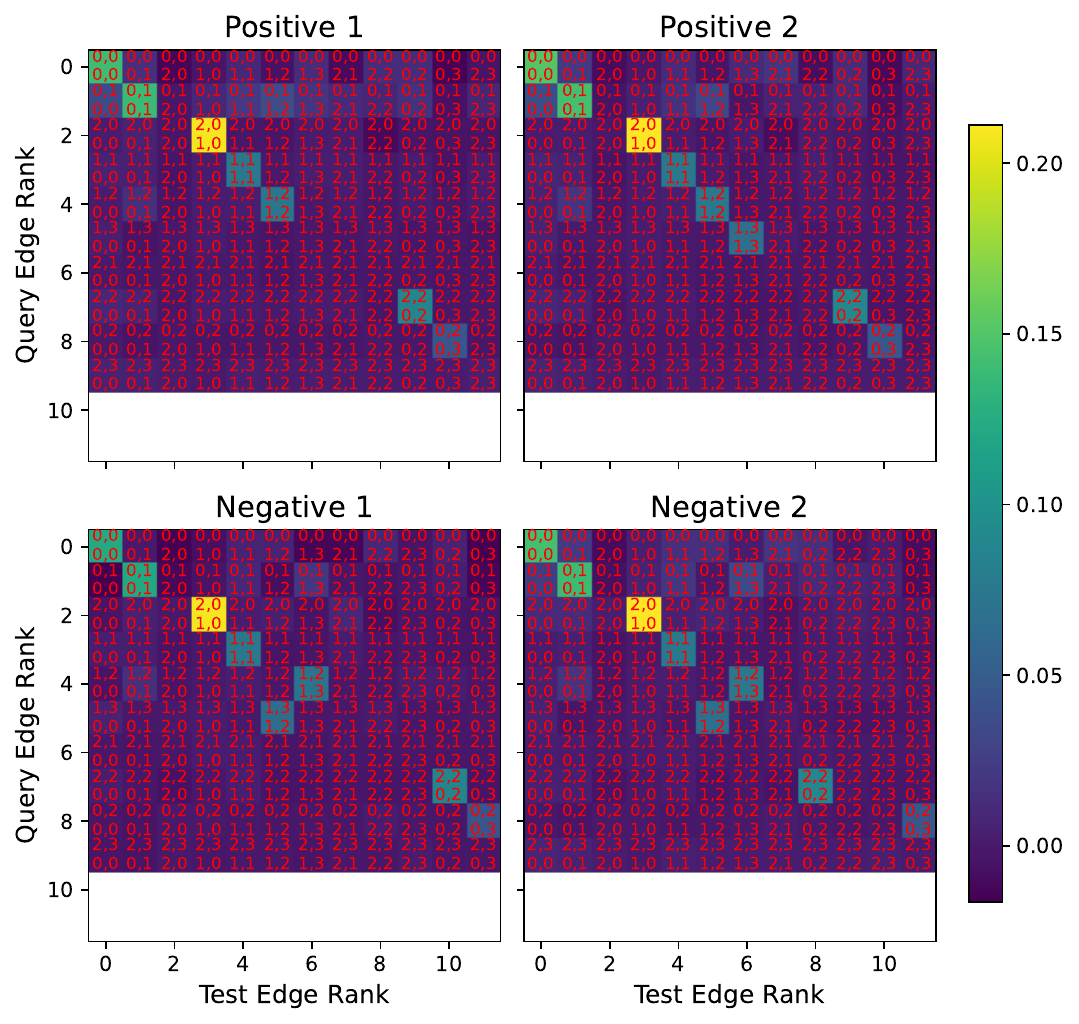}
    \caption{Aggregate Feature Similarities.}
    \end{subfigure}%
  }\vfill
  \hbox{%
    \setcounter{subfigure}{4}
    \begin{subfigure}[b]{.46\textwidth}
    \centering
    \includegraphics[width=\textwidth]{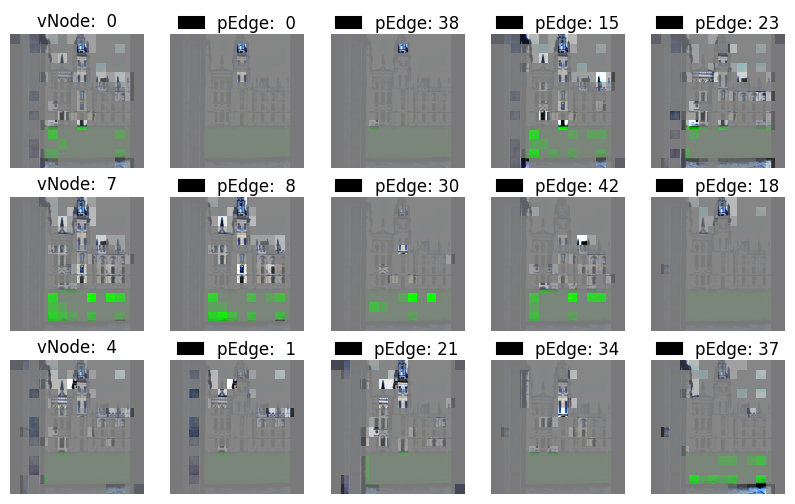}
    \caption{Pruned Positive 2 Hypergraph.}
    \end{subfigure}%
  }\vfill
  \hbox{%
    \setcounter{subfigure}{6}
    \begin{subfigure}[b]{.46\textwidth}
    \centering
    \includegraphics[width=\textwidth]{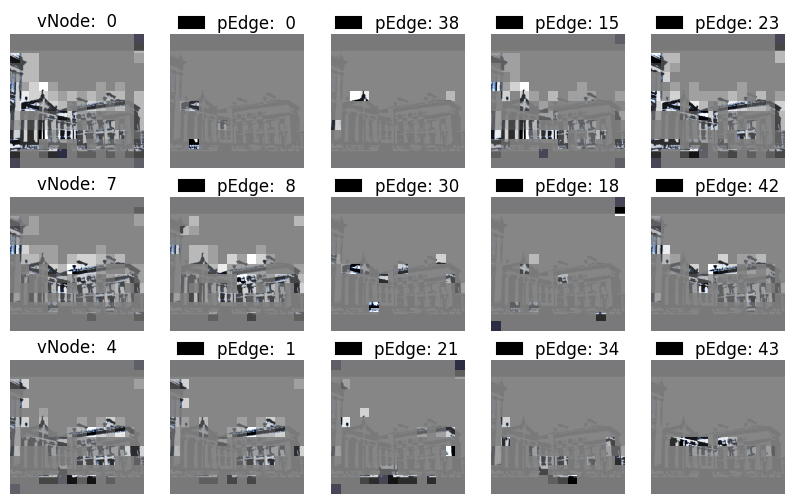}
    \caption{Pruned Positive 2 Hypergraph.}
    \end{subfigure}%
  }\cr
}

    \caption{Example Revisted Oxford retrieval for query (Ashmolean Museum), with two positive and two negative results for HgVT-Ti. (a) showing input images, (b) pruned hypergraph visualization for the query image, (c) aggregate hyperedge similarity scores, (d-e) pruned hypergraph visualizations of the positive image pairs, (f-g) pruned hypergraph visualizations of the negative image pairs. All hypergraph visualizations label the top-3 virtual vertices (vNode) and their corresponding top-4 primary hyperedges (pEdge). If a primary hyperedge connects to multiple virtual vertices, this link is indicated by a unique marker other than solid black. In (c), the corresponding query~(top) and test hyperedge~(bottom) coordinates are indicated by red numbers: as \texttt{vNode,pEdge}. For example: pEdge 47 in the query hypergraph would be \texttt{2,1}. In all cases, query pEdge 8 (\texttt{2,0}) has the highest similarity with pEdge 8 (\texttt{1,0}) in the test images.}
    \label{fig:rox_sim_ex}
\end{figure}

In \cref{fig:rox_sim_ex}, we present a test query image alongside two known positive images and two known negative images. For each of these five images, we show the pruned hypergraph visualizations, including similarity scores for each of the primary hyperedges. Notably, distinct structural patterns emerge in the similarity scores, with higher scores between the query and positive images compared to the negative images. Additionally, we observe that the query edge rank in this example stops at 9, while the test edge ranks extend to 12. This discrepancy arises because two of the primary hyperedges in the pruned query hypergraph are duplicates, removed during deduplication, resulting in a total of 10 unique hyperedges.

\begin{figure}
    \centering
    \begin{subfigure}[b]{.45\textwidth}
    \centering
    \includegraphics[width=\textwidth]{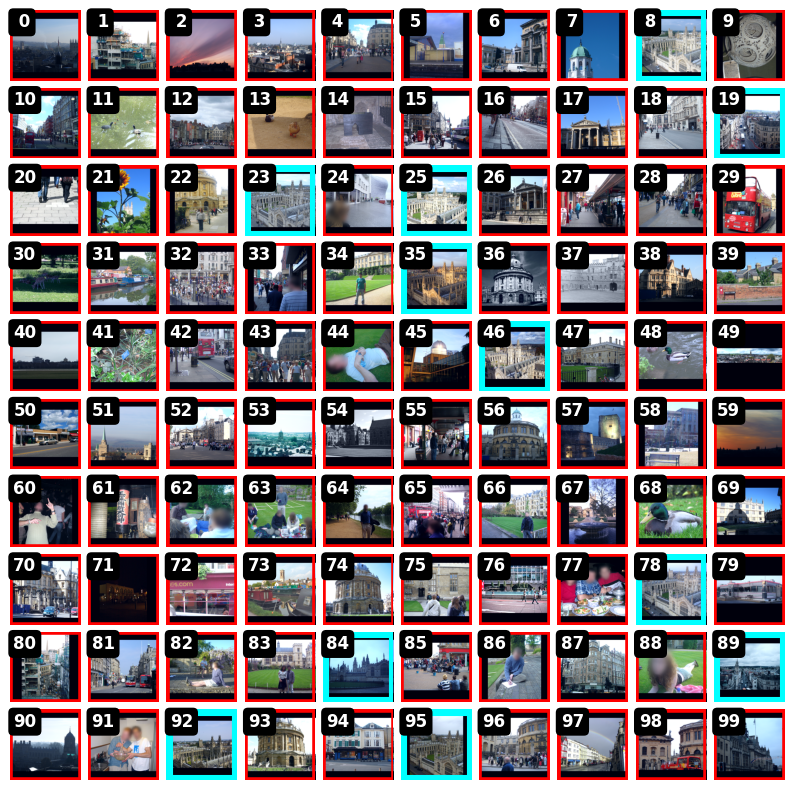}
    \caption{Pooled Similarity Ranking.}
    \end{subfigure}%
    \hspace{1em}
    \begin{subfigure}[b]{.45\textwidth}
    \centering
    \includegraphics[width=\textwidth]{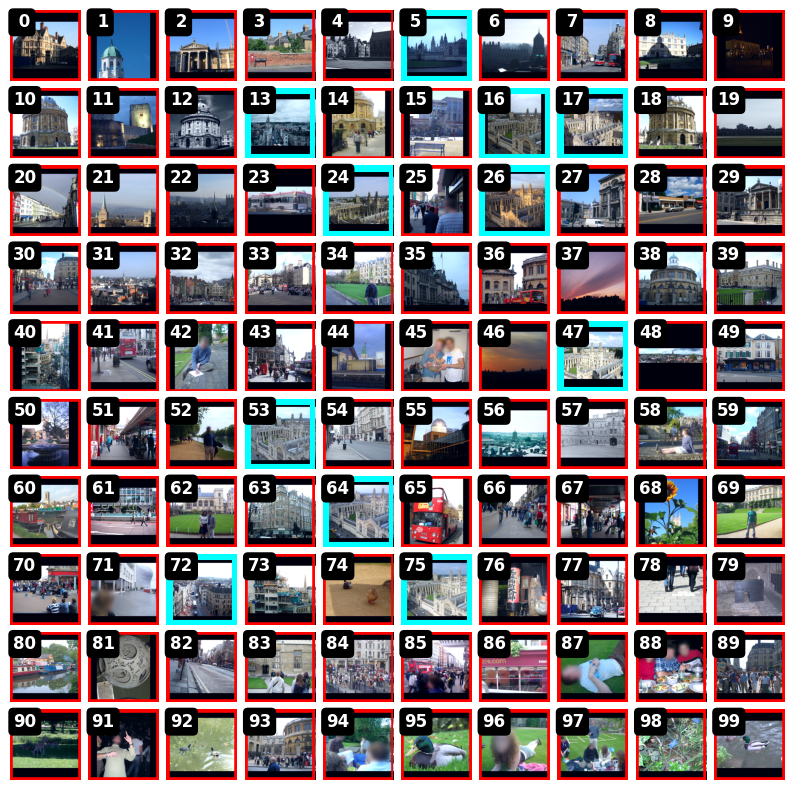}
    \caption{Adaptive Pooled Similarity Ranking.}
    \end{subfigure}%

\caption{Top-100 ranking for the medium split of the query in \cref{fig:rox_sim_ex} using HgVT-Ti. Showing (a) the results using pooled similarity ranking, (b) after re-ranking the top-100 shortlist using pruned hypergraph similarity. Positive matches are shown with thick cyan boarders, while negative matches use red boarders. The rank position is indicated by a number in the upper left corner of each image. }
\label{fig:rox_rerank}
\end{figure}    

\cref{fig:rox_rerank} further illustrates the impact of adaptive reranking on retrieval quality for the Oxford Medium dataset. Using the same query image from \cref{fig:rox_sim_ex}, we first display the top $R=100$ images retrieved based on pooled similarity ranking. In this initial retrieval, positive images are dispersed throughout the ranks, and several irrelevant images, including those without buildings, appear near the top. Applying adaptive reranking significantly improves the results: positive images are shifted to higher ranks, while irrelevant images are moved toward the end of the list. This visual evidence highlights the effectiveness of adaptive reranking in refining retrieval results by leveraging hypergraph structural information to enhance semantic alignment.

\FloatBarrier
\section{Additional Ablations}
\label{app:more_ablations}

This section evaluates the design choices and hyperparameters shaping the performance and efficiency of the HgVT models. The primary ablations are conducted on HgVT-Lt, trained on ImageNet-100, to analyze architectural trade-offs between accuracy, computational cost, and model size. To explore the impact of population regularization hyperparameters more comprehensively, we utilize a smaller model, HgVT-Mu, trained on CIFAR-100 (details in \cref{app:implement}). This allows for detailed hyperparameter sweeps to assess their effects on graph quality, sparsity, retrieval performance, and inter-metric correlations. Additionally, we investigate the influence of expert pooling regularization parameters using HgVT-Mu to better understand their role in balancing sparsity and performance. Insights from these evaluations guide the selection of optimal configurations and provide a deeper understanding of the underlying model behavior.

\subsection{Population Regularization Sweeps}

\begin{figure}[ht]
  \centering
  \begin{subfigure}[b]{1.0\textwidth}
        \includegraphics[width=\textwidth]{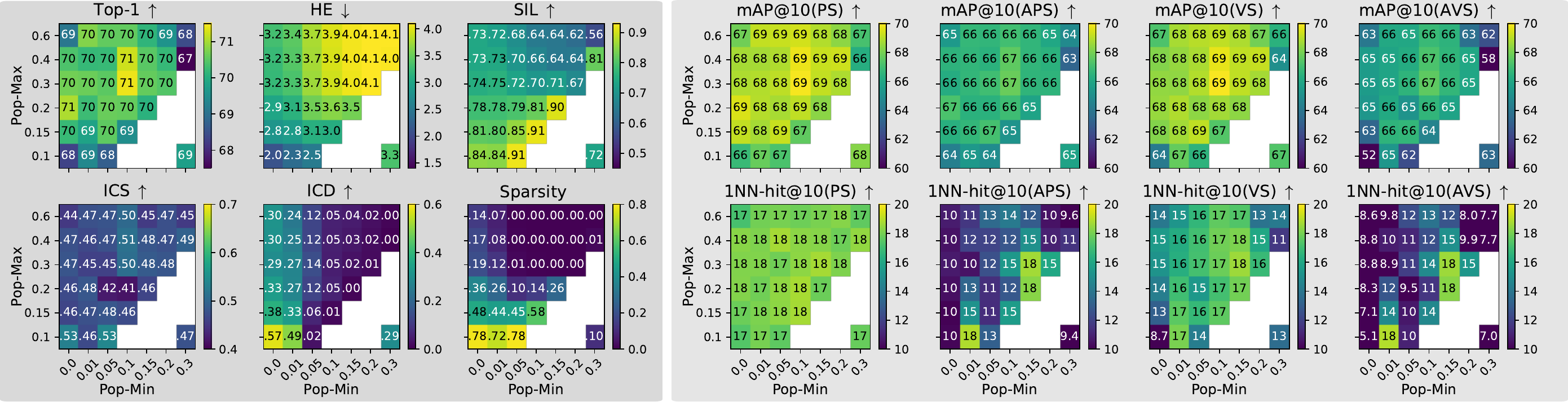}
        \caption{Standard Hadamard Soft Membership Modulation.}
        \label{fig:f1}
    \end{subfigure}
    \hfill 
  \begin{subfigure}[b]{1.0\textwidth}
        \includegraphics[width=\textwidth]{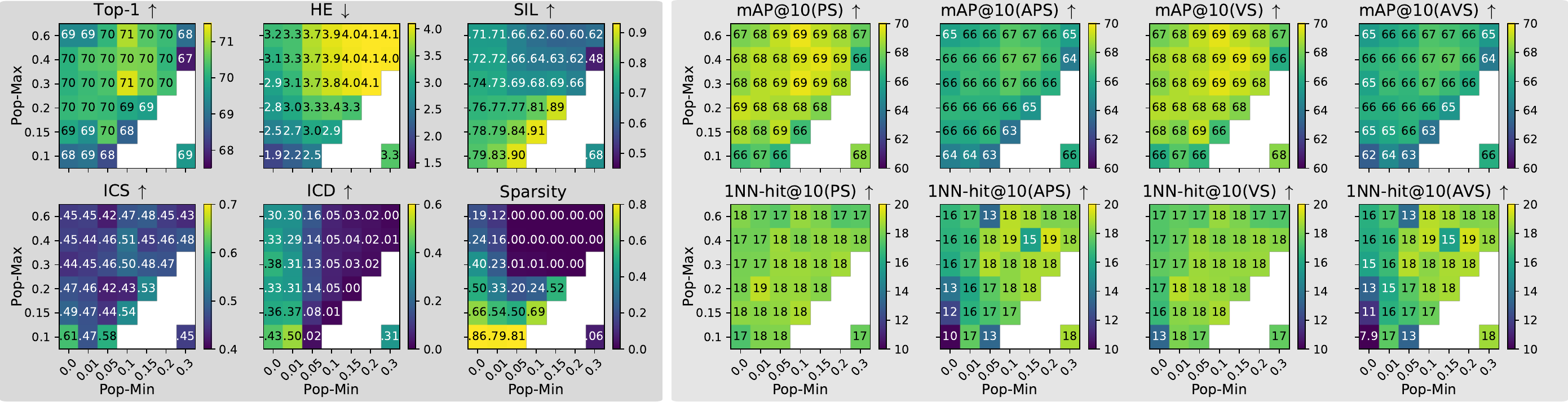}
        \caption{Modified Hadamard Soft Membership Modulation.}
        \label{fig:f1}
    \end{subfigure}
  \caption{Effect of population regularization minimum ($\gamma$) and maximum ($\beta$) density limits, for CIFAR-100. Regularization is normalized by $|\mathcal{V}|$, such that 1.0 corresponds to $\beta,\gamma=|\mathcal{V}|$. Lower-right cell in each subplot represents population regularization disabled. Left, showing top-1 accuracy and graph quality metrics hyperedge entropy~(HE), silhouette score~(SIL), intra-cluster similarity~(ICS), inter-cluster distance~(ICD), and sparsity~(spA). Right, showing mAP@10 for image retrieval and top-10 hit-rate with top-1 CLIP-B ranking for four retrieval methods: standard, adaptive~(A), volumetric overlap~(V), and adaptive volumetric~(VA). Further comparing (a) with standard Hadamard (bounded between 0 and 1), and (b) with modified Hadamard (bounded between -1 and 1) modulation in edge attention. }
  \label{fig:cifar_sweep}
\end{figure}

The population regularization mechanism facilitates learned self-sparsification and clustering within the generated hypergraphs. It is defined by the population regularization minimum density ($\gamma$) and maximum density ($\beta$), with the regularization terms encouraging soft adjacency membership contributions to remain within these bounds. A sweep of these parameters, normalized to the vertex count $|\mathcal{V}|$ is presented in \cref{fig:cifar_sweep} for the HgVT-Mu model, comparing the standard Hadamard edge attention modulation to the modified Hadamard edge attention modulation.

The results in \cref{fig:cifar_sweep} indicate that the modified Hadamard modulation consistently outperforms the standard approach. This improvement aligns with expectations, as the modified modulation removes the positive influence of non-membership vertices, thereby enhancing the accuracy of edge relationships. Both parameter grids form distinct performance landscapes, revealing regions where over-sparsification occurs and others where structural collapse leads to a maximally connected graph~\mbox{(sparsity = 0)}.
Interestingly, top-1 accuracy and retrieval metrics generally favor the maximally connected case initially, but performance begins to degrade beyond a certain point. This suggests that the metrics benefit from a weakly maximally connected graph -- characterized by softer membership weights -- over a strongly maximally connected graph with more rigid weights. However, while a maximally connected structure may boost certain metrics temporarily, it ultimately hinders precise structural extraction and efficient computation, both of which rely on maintaining an appropriate level of sparsity.

\begin{figure}[ht]
  \centering
  \begin{subfigure}[b]{0.48\textwidth}
        \includegraphics[width=\textwidth]{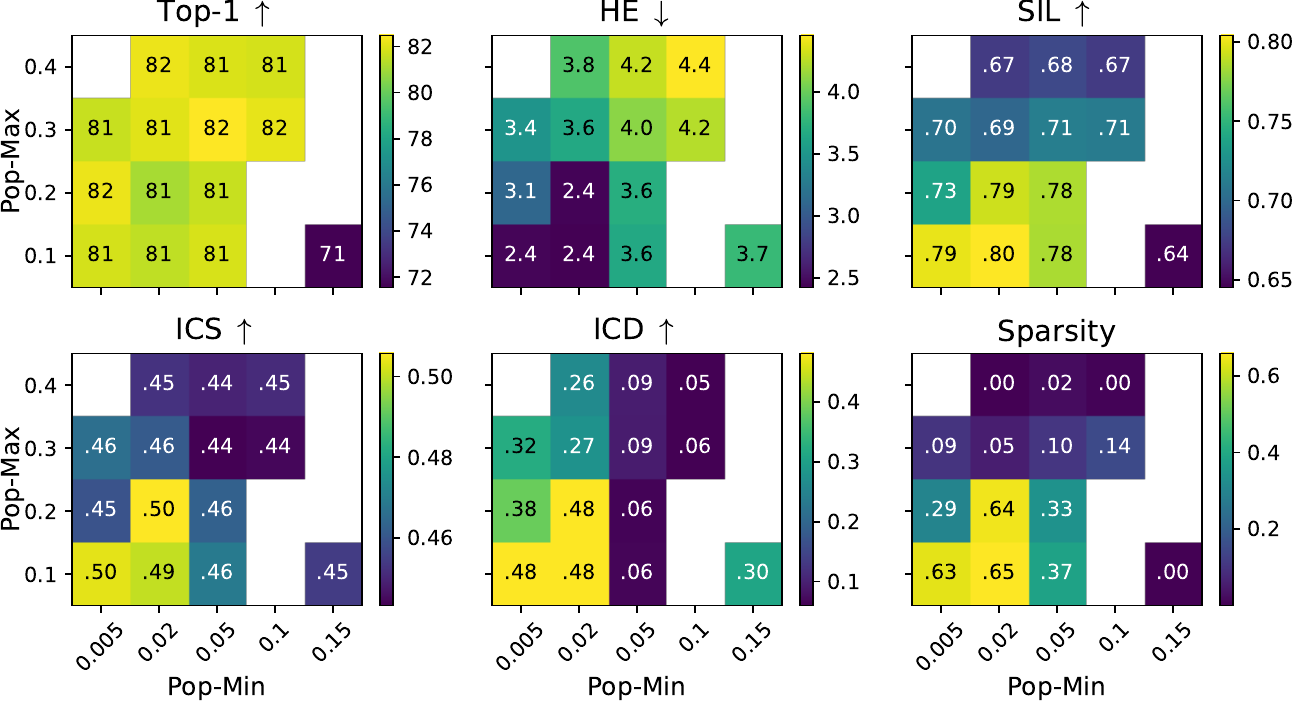}
        \vspace{-1.5em}
        \label{fig:f1}
    \end{subfigure}
  \caption{Effect of population regularization minimum ($\gamma$) and maximum ($\beta$) density limits, for ImageNet-100. Regularization is normalized by $|\mathcal{V}|$, such that 1.0 corresponds to $\beta,\gamma=|\mathcal{V}|$. Lower-right cell in each subplot represents population regularization disabled. Showing top-1 accuracy and graph quality metrics hyperedge entropy~(HE), silhouette score~(SIL), intra-cluster similarity~(ICS), inter-cluster distance~(ICD), and sparsity~(spA). }
  \label{fig:im100_sweep}
\end{figure}

To validate the findings from the HgVT-Mu model at scale, \cref{fig:im100_sweep} presents a similar population regularization analysis for the HgVT-Lt model, trained on ImageNet-100. This analysis uses a coarser parameter grid and focuses on Top-1 accuracy and graph quality metrics. The results demonstrate a similar performance pattern to that observed with HgVT-Mu, but with a noticeable shift toward lower values of the normalized population minimum density ($\gamma$). Notably, the best results are achieved when $\gamma$ is maintained at an absolute value of $0.5$, rather than scaling it with the vertex count $|\mathcal{V}|$. This suggests that a fixed minimum density is sufficient to ensure effective graph sparsity and clustering, even as the model scales, while also preventing over-sparsification (sparsity $\rightarrow 1.0$). In contrast, the population maximum density $\beta$ benefits from scaling, with $\beta=1/6 \cdot |\mathcal{V}|$ performing well across the Mu, Lt, and Ti scales. This configuration yields an average graph sparsity of approximately 30\% to 60\%, striking a balance between maintaining structural integrity and enabling efficient computation. These findings reinforce the generalizability of the population regularization framework across scales while providing practical guidance for selecting $\gamma$ and $\beta$ values.

\subsection{Correlation Analysis of Metrics}

\begin{figure}[ht]
  \centering
  \begin{subfigure}[b]{1.0\textwidth}
        \includegraphics[width=\textwidth]{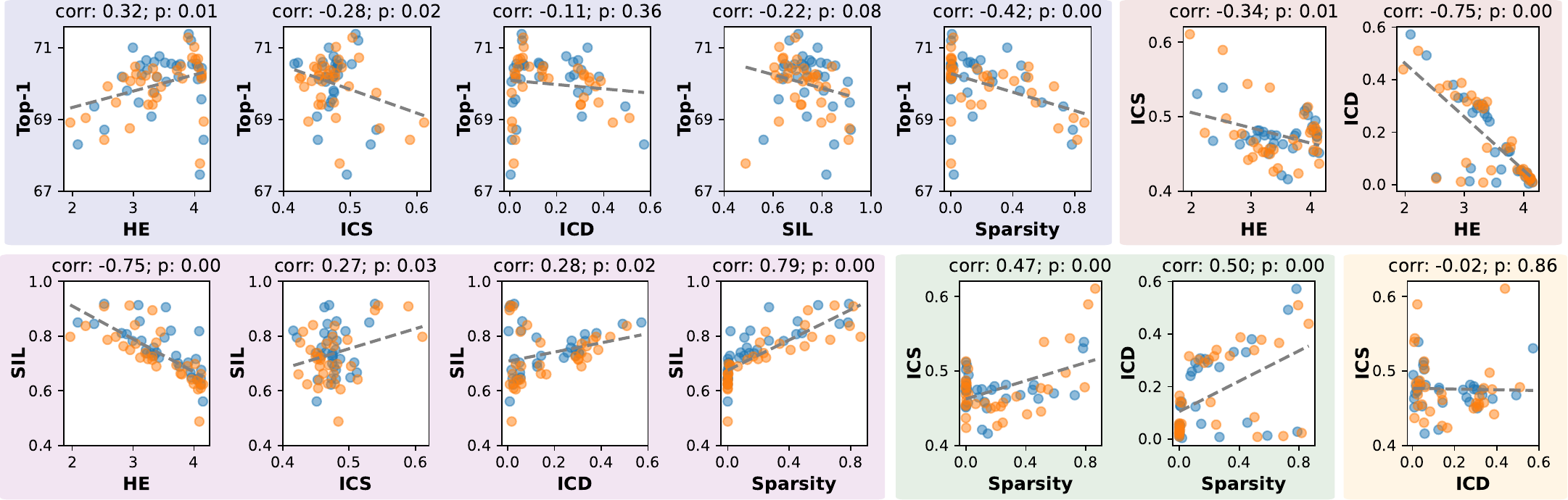}
        \caption{Graph Structure Correlations.}
        \label{fig:cifar_corr_graph}
    \end{subfigure}
    \hfill 
  \begin{subfigure}[b]{1.0\textwidth}
        \includegraphics[width=\textwidth]{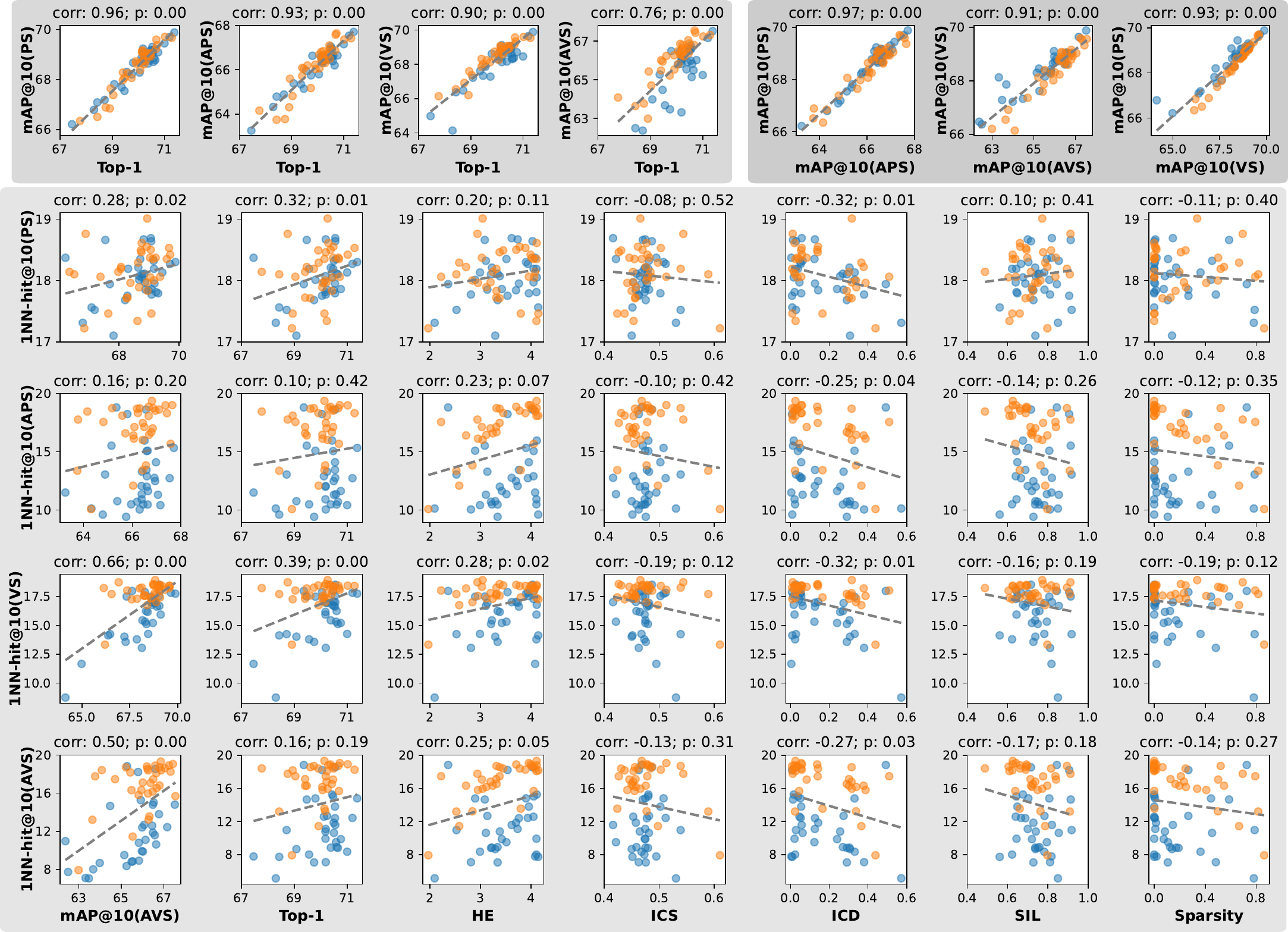}
        \caption{Retrieval Accuracy Correlations.}
        \label{fig:cifar_corr_ret}
    \end{subfigure}
  \caption{Comparing structural correlations obtained by the population regularization sweep on CIFAR-100 from \cref{fig:cifar_sweep}. (a) measuring top-1 accuracy and intra-structural correlations; (b) showing structural correlations with image retrieval accuracy. Plotting both standard Hadamard (blue) and modified Hadamard (orange) soft membership modulation. Correlation coefficients and significance p-values plotted above each subplot, with correlation trendlines shown as gray-dashed lines.}
  \label{fig:cifar_corr}
\end{figure}

To further investigate the interactions between different metrics, we compute correlations across the HgVT-Mu population regularization sweep for both the standard Hadamard and modified Hadamard modulation methods. These correlations are visualized in \cref{fig:cifar_corr}, with graph quality metrics (HE, ICS, ICD, SIL, sparsity) analyzed in \cref{fig:cifar_corr_graph} and retrieval performance metrics (mAP@10 and 1NN-hit@10 for PS, VS, APS, AVS) in \cref{fig:cifar_corr_ret}. Each plot includes a best-fit trendline alongside the correlation coefficient and p-value to assess statistical significance. 

\textbf{Graph Quality Metrics:} Top-1 accuracy shows weak correlations with all graph quality metrics, positively with hyperedge entropy (HE) and negatively with all others, including sparsity. This supports the observation that maximally connected graphs tend to yield better Top-1 performance. SIL is negatively correlated with HE and positively correlated with sparsity, suggesting a trade-off between hyperedge feature variance and graph separation. Similarly, ICD is negatively correlated with HE, while ICS and ICD exhibit no correlation with each other. Most other interactions between graph quality metrics are relatively weak.

\textbf{Retrieval Metrics:} All mAP@10 metrics are highly correlated with Top-1 accuracy, with AVS exhibiting the largest variance. Adaptive methods (APS, AVS) are strongly correlated with their non-adaptive counterparts (PS, VS), while PS and VS also display strong mutual correlation. These relationships highlight consistent dependencies between retrieval metrics and Top-1 accuracy.

\textbf{1NN-hit@10 Metrics:}  Acting as a proxy for semantic alignment with CLIP, the 1NN-hit@10 results reveal distinct groupings based on modulation type, with the modified Hadamard method outperforming the standard method. Interestingly, correlations in this category are generally weak, with the strongest observed between mAP@10 for the AVS method and 1NN-hit@10. This correlation is particularly notable when comparing VS and AVS within the 1NN-hit@10 metric. These findings suggest that while retrieval metrics align well with accuracy, their connection to semantic alignment is more nuanced and varies across methods.

\FloatBarrier
\subsection{Expert Pooling Regularization}

\begin{figure}[ht]
\vspace{1em}
    \centering
        \begin{minipage}{0.52\linewidth}
  \centering
  \begin{subfigure}[b]{\textwidth}
        \includegraphics[width=\textwidth]{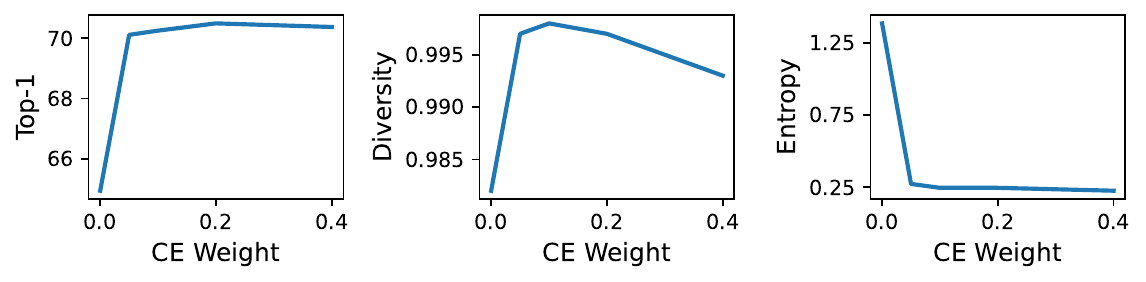}
        \vspace{-1.5em}
        \caption{Impact of CE Weight.}
        \label{fig:f1}
    \end{subfigure}
    \vfill 
  \begin{subfigure}[b]{\textwidth}
        \includegraphics[width=\textwidth]{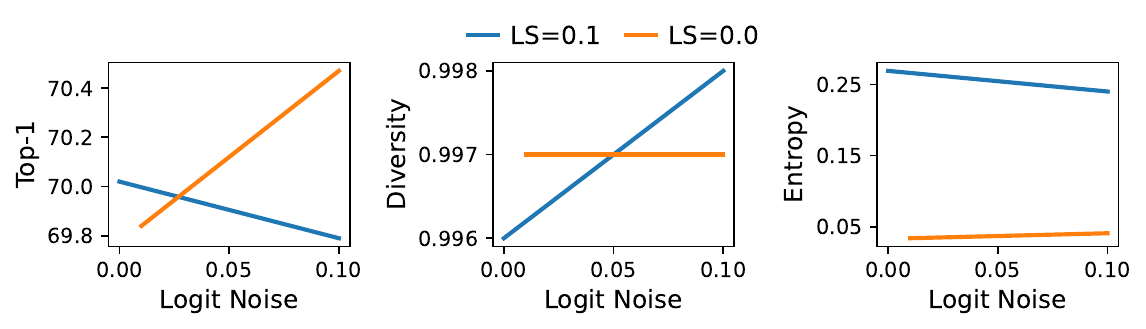}
        \vspace{-1.5em}
        \caption{Impact of Noise Injection.}
        \label{fig:f1}
    \end{subfigure}

  \label{fig:arch_pareto}
    \end{minipage}
    \hfill
\begin{minipage}{0.44\linewidth}
        \centering
        \captionof{table}{Ablating Expert Edge pooling regularization methods for the HgVT-Mu model trained on CIFAR-100. 
    } \label{tab:expert_reg}
    \centering
    \vspace{-0.5em}
    \resizebox{1.0\textwidth}{!}{
    \begin{tabular}{c c c | c c c }
    \toprule
    Density Loss & \parbox{1.5cm}{\hspace{1em}Label\\ Smoothing} & Dropout & Top-1 & Diversity & Entropy \\
    \midrule
    \Yes & \Yes & \Yes & 70.25 & 0.998 & 0.245 \\
    \Yes & \Yes & \No & 70.18 & 0.995 & 0.256 \\
    \Yes & \No & \Yes & 69.81 & 0.987 & 0.039 \\
    \Yes & \No & \No & 69.95 & 0.994 & 0.043 \\
    \midrule
    \No & \Yes & \Yes & 66.37 & 0.0 & 1.386 \\
    \No & \Yes & \No & 63.34 & 0.329 & 1.386 \\
    \No & \No & \Yes & 64.54 & 0.323 & 1.386 \\
    \bottomrule
    \end{tabular}
    }
    \end{minipage}
    \vspace{-0.5em}
      \caption{Parameter sweep of Expert Edge hyperperameters for HgVT-Mu trained on CIFAR-100. (a) Varying cross-entropy loss weight; (b) varying logit noise injection strength with (LS=0.1) and without (LS=0.0) label smoothing. For both figures, showing top-1 prediction accuracy, diversity (1.0 indicates equal distribution among experts), and selection entropy (lower indicates higher confidence).}
          \label{fig:expert_reg}
\end{figure}

Expert pooling regularization is evaluated on the HgVT-Mu model trained on CIFAR-100, focusing on the cross-entropy (CE) weight and logit noise injection strength. \cref{fig:expert_reg} examines these parameters, presenting Top-1 accuracy, expert diversity (where 1.0 indicates uniform expert utilization), and expert entropy (lower values indicate higher confidence). A CE weight of 0.1 achieves a good balance, yielding confident routing and high accuracy. Without the CE weight, expert entropy increases significantly, reflecting low-confidence routing that hinders performance. For logit noise injection, higher noise levels ($10^{-1}$) outperform label smoothing, improving both accuracy and diversity. This indicates that noise injection is a more effective regularization strategy, avoiding the higher entropy associated with label smoothing.


\cref{tab:expert_reg} explores the combinatorial effects of diversity loss, label smoothing, and dropout regularization. Diversity loss proves essential, preventing expert collapse and achieving the highest diversity metric. Label smoothing and dropout individually have minor effects but, when combined, produce the best Top-1 accuracy and diversity results. However, label smoothing increases entropy, potentially reducing confidence. This is mitigated by omitting label smoothing and using higher logit noise instead, which preserves confidence while improving diversity and accuracy.


\FloatBarrier




\subsection{Additional HgVT-Lt Model Ablations}

\begin{figure}[ht]
    \centering
        \begin{minipage}{0.49\linewidth}
  \centering
  \vspace{2em}
  \includegraphics[width=\textwidth]{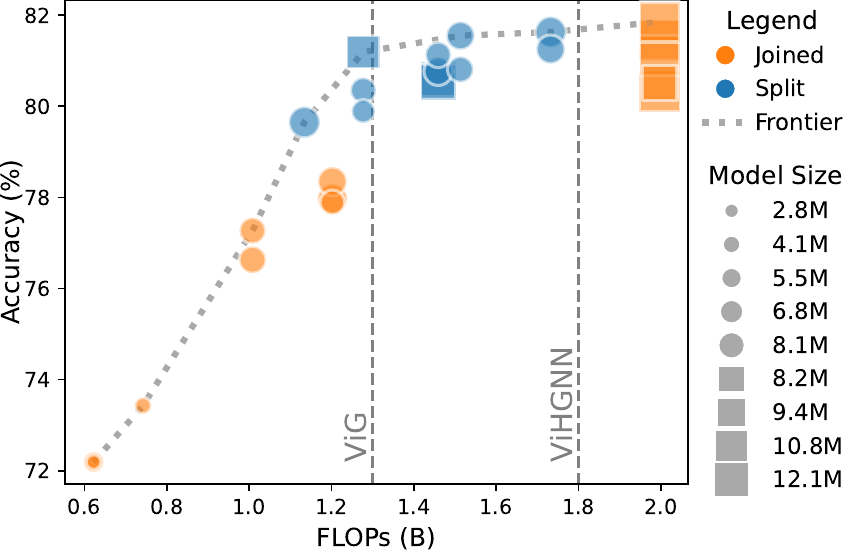}

    \end{minipage}
    \hfill
    \begin{minipage}{0.5\linewidth}
        \centering
        \captionof{table}{
        Architectural Ablations for HgVT-Lt trained on ImageNet-100. All experiments presented use average edge pooling.
        } \label{tab:app_more_abl}
        \centering
        \vspace{-0.5em}
        \resizebox{1.0\textwidth}{!}{
    \begin{tabular}{c c | c c c c c | c c c}
    \toprule
    $X_\adj = X$ & Joint FFN & $L$ & $d_f$ & $d_a$ & $h$ & $d_k$ & Top-1 & Params & FLOPs \\
    \midrule
    \No & \No & 10 & 96 & 96 & 3 & 32 & 81.19 & 8.3M & 1.3G \\
    \No & \No & 10 & 128 & 64 & 4 & 32 & 80.59 & 9.2M & 1.5G \\
    \No & \No & 10 & 64 & 128 & 2 & 32 & 80.59 & 7.7M & 1.1G \\
    \No & \No & 10 & 128 & 64 & 2 & 64 & 80.43 & 9.2M & 1.5G \\
    \No & \No & 10 & 64 & 128 & 1 & 64 & 80.25 & 7.7M & 1.1G \\
    \midrule
    \No & \Yes & 10 & 128 & 64 & 4 & 32 & 80.77 & 6.6M & 1.5G \\
    \No & \Yes & 10 & 96 & 96 & 3 & 32 & 80.35 & 5.7M & 1.3G \\
    \No & \Yes & 12 & 128 & 64 & 4 & 32 & 81.63 & 7.6M & 1.7G \\
    \No & \Yes & 12 & 96 & 96 & 3 & 32 & 81.55 & 6.5M & 1.5G \\
    \midrule
    \Yes & \No & 10 & 96 & 96 & 3 & 32 & 72.19 & 4.0M & 0.6G \\
    \Yes & \No & 10 & 128 & 128 & 4 & 32 & 77.27 & 5.9M & 1.0G \\
    \Yes & \No & 12 & 128 & 128 & 4 & 32 & 78.35 & 6.8M & 1.2G \\
    \Yes & \Yes & 12 & 128 & 128 & 4 & 32 & 77.89 & 5.4M & 1.2G \\
    \Yes & \No & 10 & 192 & 96 & 6 & 32 & 81.85 & 11.8M & 2.0G \\
    \Yes & \No & 10 & 192 & 96 & 3 & 64 & 81.11 & 11.8M & 2.0G \\
    \Yes & \Yes & 10 & 192 & 96 & 3 & 64 & 80.51 & 9.1M & 2.0G \\
    \bottomrule
    \end{tabular}
        }
    \end{minipage}

    \vspace{-0.5em}
      \caption{Showing ImageNet-100 classification accuracy~vs~forward compute (in FLOPs) for an architectural sweep of the HgVT-Lt model using expert pooling. Parameter count is shown by marker size, where models larger than ViHGNN-Ti~\cite{vihgnn_paper} are represented by squares rather than circles. All FLOPs and Parameters are measured using the equivalent HgVT-Ti models on ImageNet-1k with expert pooling. Further showing models with joined ($\mathbf{X}^{(*)}_\adj = \mathbf{X}^{(*)}$; orange), and split ($\mathbf{X}^{(*)}_\adj \neq \mathbf{X}^{(*)}$; blue) adjacency features, along with the Pareto frontier.}
      \label{fig:arch_pareto}
\vspace{-1.5em}
\end{figure}

Additional ablations on the HgVT-Lt model trained on ImageNet-100 explore various structural configurations. \cref{tab:app_more_abl} lists these configurations, reporting their Top-1 accuracy, parameter count, and FLOPs. \cref{fig:arch_pareto} visualizes the results, plotting accuracy against FLOPs, with marker size representing model size. The Pareto frontier is highlighted, alongside comparisons with ViG and ViHGNN, providing a reference point for FLOPs and parameter count.

The findings indicate that using split adjacency and feature matrices ($X_\adj \neq X$) improves performance. Allocating more dimensions to the feature matrix than the adjacency matrix ($d_f > d_a$) strikes a balance between accuracy and computational overhead. Using more attention heads with smaller key dimensions ($d_k=32$) outperforms fewer heads with a larger dimension. Furthermore, sharing the same feed-forward network (FFN) between edges and vertices reduces parameters with minimal accuracy loss. Several alternative configurations to the one chosen for HgVT-Lt are noted, offering trade-offs between computational overhead and accuracy for future scaling considerations.

\FloatBarrier
\section{Implementation Details}
\label{app:implement}

All models were trained using PyTorch with automatic mixed precision, leveraging the PyTorch-Lightning framework. Vertex self-attention was implemented efficiently using the xformers library~\cite{xFormers2022}, while edge attention utilized einsum operations reodered for memory efficiency with \verb|torch.compile|. The Timm library~\cite{rw2019timm} was employed for data augmentation, learning rate scheduling, and optimizer initialization, with the Fused AdamW optimizer from the Apex library~\cite{nvidia_apex}.

Retrieval methods were implemented by storing precomputed features in HDF5 tables and conducting similarity searches directly on the GPU via PyTorch. The pooled embeddings of the full database were compact enough to reside in VRAM, enabling batch comparisons and efficient similarity sorting. Reranking computations were performed using Numpy on the shortlist features, eliminating the need to store these features on the GPU and maintaining computational efficiency.





\subsection{Training Hyperparameters}

\begin{table}[h!]
  \caption{Details of data augmentation parameters, common to all runs. }
  \label{tab:dataset_parameters}
  \vspace{-0.5em}
  \centering
  \resizebox{0.48\textwidth}{!}{
  \begin{tabular}{l c}
    \toprule
    \textbf{Parameter} & \textbf{Value} \\
    \midrule
    Random Erase Mode & Pixel \\
    Random Erase Probability & 0.25 \\
    Random Erase Count & 1 \\
    \midrule
    Label Smoothing & 0.1 \\
    Mixup $\alpha$ & 0.8 \\
    CutMix $\alpha$ & 1.0 \\
    Mixup Probability & 0.8 \\
    Mixup Switch probability & 0.5 \\
    Mixup Mode & Batch \\
    \midrule
    Repeat Augmentation Count & 2 \\
    \midrule
    Color Jitter & 0.4 \\
    Interpolation Mode & Random \\
    Random Scale Range & [0.08, 1.0] \\
    Random Aspect Ratio Range & [0.75, 1.33] \\
    Random HFlip Probability & 0.5 \\
    Auto-Agumentation Config. & \texttt{rand-m9-mstd0.5-inc1} \\
    \bottomrule
  \end{tabular}
  }
  \vspace{-1em}
\end{table}

\begin{table}[h!]
  \caption{Details of training hyper-parameters.}
  \label{tab:hyper_params}
  \centering
  \begin{tabular}{l c c c c}
    \toprule
    \textbf{Parameter \textbackslash \hspace{0.2em} Scale $\rightarrow$} & \textbf{Mu} & \textbf{Lt} & \textbf{Ti}  & \textbf{S} \\
    \midrule
    Dataset    & CIFAR100 & ImageNet-100 & ImageNet-1k & ImageNet-1k \\
    Resolution & 32 x 32 & 160 x 160 & 224 x 224 & 224 x 224 \\
    Parameters & 2.90M & 6.82M & 7.76M & 22.94M\\
    Fwd. FLOPS & 0.15G & 0.92G & 1.80G & 5.48G\\
    \midrule
    Optimizer & AdamW & AdamW & AdamW & AdamW \\
    Peak Learning Rate & 1e-3 & 1e-3 & 1e-3 & 1e-3 \\
    Betas & [0.9, 0.999] & [0.9, 0.999] & [0.9, 0.999] & [0.9, 0.999]\\
    Eps & 1e-8 & 1e-8 & 1e-8 & 1e-8 \\
    Weight Decay & 5e-2 & 5e-2 & 5e-2 & 5e-2\\
    Gradient Clip & 1.0 & 1.0 & 1.0 & 1.0\\
    \midrule
    Training Epochs & 400 & 200 & 300 & 300\\
    Warmup Epochs & 10 & 16 & 10 & 10 \\
    Global Batch Size & 512 & 512 & 1024 & 1024\\
    Grad. Accum. Steps & 1 & 1 & 1 & 2\\
    Training Hardware & 1x A6000 & 1x A6000 & 2x A6000 & 2x A6000\\
    Precision & bfloat16 & bfloat16 & bfloat16 & bfloat16\\
    Attn. Precision & float32 & float32 & float32 & float32\\
    Training Time & 2 Hours & 8 Hours & 139 Hours & 255 Hours\\
    \midrule
    Depth ($L$) & 10 & 12 & 12 & 14 \\
    Feature Dim ($d_f$) & 64 & 128 & 128 & 224 \\
    Adj. Dim ($d_a$) & 64 & 64 & 64 & 96 \\
    Heads ($h$) & 2 & 4 & 4 & 7 \\
    Joint FFN & True & True & True & True \\
    $\mathbf{X}_\adj = \mathbf{X}$ & False & False & False & False \\
    \midrule
    Patch Size & 4 & 16 & 16 & 16 \\
    Image Verts. ($|i\mathcal{V}|$) & 64 & 100 & 196 & 196\\
    Virtual Verts. ($|v\mathcal{V}|$) & 5 & 12 & 16 & 16\\
    Primary Edges ($|p\mathcal{E}|$) & 8 & 32 & 50 & 50\\
    Virtual Edges ($|v\mathcal{E}|$) & 4 & 6 & 8 & 8\\
    Use Conv. Stem & True & True & True & True \\
    \midrule
    Stochastic Path Drop & 0.1 & 0.1 & 0.1 & 0.1 \\
    Class Dropout & 0.1 & 0.0 & 0.0 & 0.0 \\
    Drop Decay & False & True & True & True \\
    Pop Max ($\beta$) & 10.05 & 20.7 & 36.04 & 36.04 \\
    Pop Min ($\gamma$) & 0.5 & 0.5 & 0.5 & 0.5 \\
    $\lambda_\mathrm{POP}$ & 1.0 & 1.0 & 1.0 & 1.0 \\
    $\lambda_\mathrm{DIV}$ & 1.0 & 1.0 & 1.0 & 1.0 \\
    $\lambda_\mathrm{EXP}$ & 1.0 & 1.0 & 1.0 & 1.0 \\
    \midrule
    Pooling Method & Expert & Expert+Image & Expert+Image & Expert+Image \\
    Expert Top-k & 1 & 1 & 1 & 1 \\
    Expert $\lambda_\mathrm{CE}$ & 0.1 & 0.1 & 0.1 & 0.1 \\
    Expert Noise & 0.1 & 0.1 & 0.1 & 0.1 \\
    Expert Dropout & 0.1 & 0.1 & 0.1 & 0.1 \\
    Expert Label Smoothing & 0.0 & 0.0 & 0.0 & 0.0 \\
    
    \bottomrule
  \end{tabular}
\end{table}

\FloatBarrier
\newpage
\FloatBarrier
\section{Macro-Class Clustering with Expert Edge Pooling}
\label{app:macro_prediction}

This section provides taxonomy trees illustrating the macro-class clusters formed by our proposed expert pooling method. These clusters emerge as experts learn to select subsets of the hypergraph, revealing groupings aligned with high-level semantic categories.To illustrate, we present clusters from two models: HgVT-Lt, trained on ImageNet-100, and HgVT-S trained on ImageNet-1k. Given the reduced class count in ImageNet-100, the clusters for HgVT-Lt are more directly analyzable, whereas the larger taxonomy of ImageNet-1k consists of a broader set of categories.

Class-to-expert assignments are determined by histograms aggregated over the respective validation sets and follow a $2/3$ probability density rule: each class is assigned initially to its highest-probability expert, and subsequent experts are added if the most recently added expert contains less than $2/3$ of the remaining probability, until the total cumulative probability reaches 80\%. For example, probability ranking $[54\%, 28\%,  12\%, 6\%]$ would assign the first two experts, while $[46\%, 24\%, 22\%,  8\%]$ would assign the first three experts. This allocation method produces a pattern of mostly single-expert assignments, tapering off with smaller groups assigned to two or more experts, which we visualize in the taxonomy trees in the following subsections.

\subsection{HgVT-Lt on ImageNet-100}

\vspace{-1em}
\begin{figure}[ht]
    \centering
    \begin{subfigure}[b]{.95\textwidth}
    \centering
    \includegraphics[width=\textwidth]{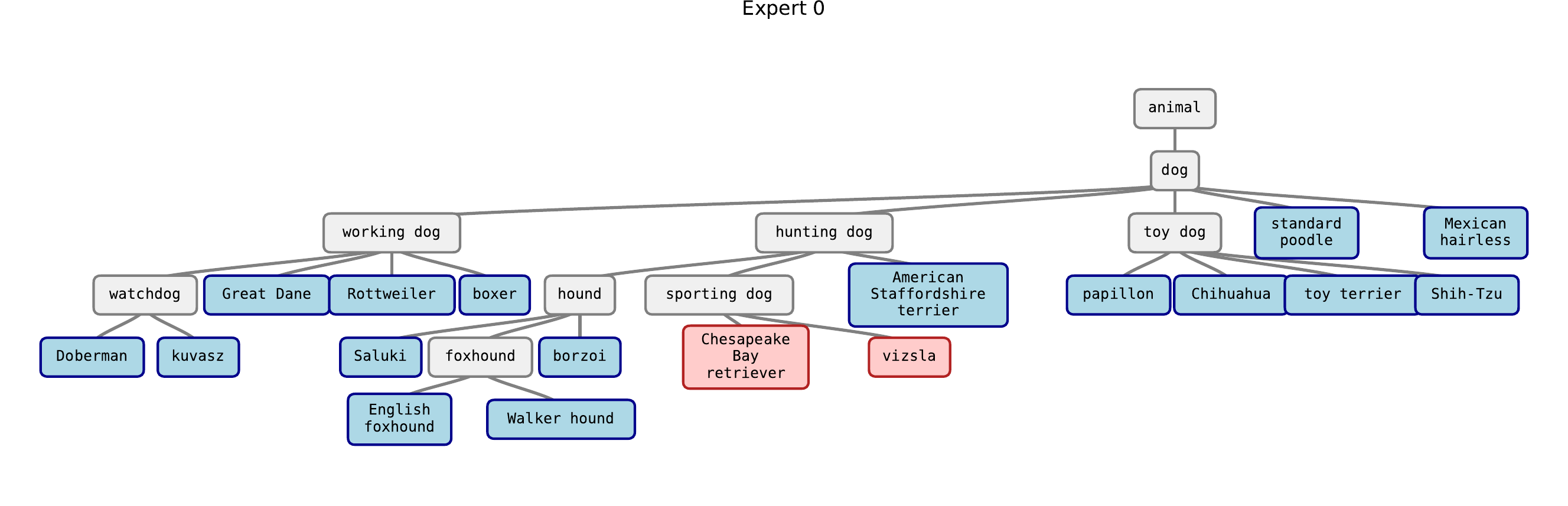}
    \end{subfigure}%
    \vfill
    \vspace{-1em}
    \begin{subfigure}[b]{.95\textwidth}
    \centering
    \includegraphics[width=\textwidth]{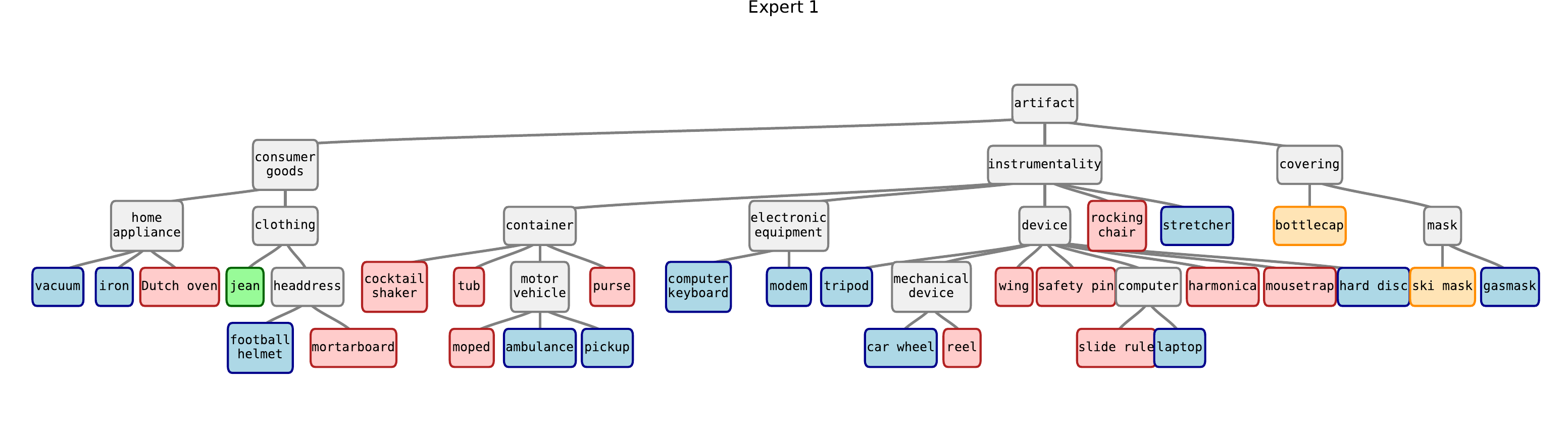}
    \end{subfigure}%
    \vfill
    \begin{subfigure}[b]{.95\textwidth}
    \centering
    \includegraphics[width=\textwidth]{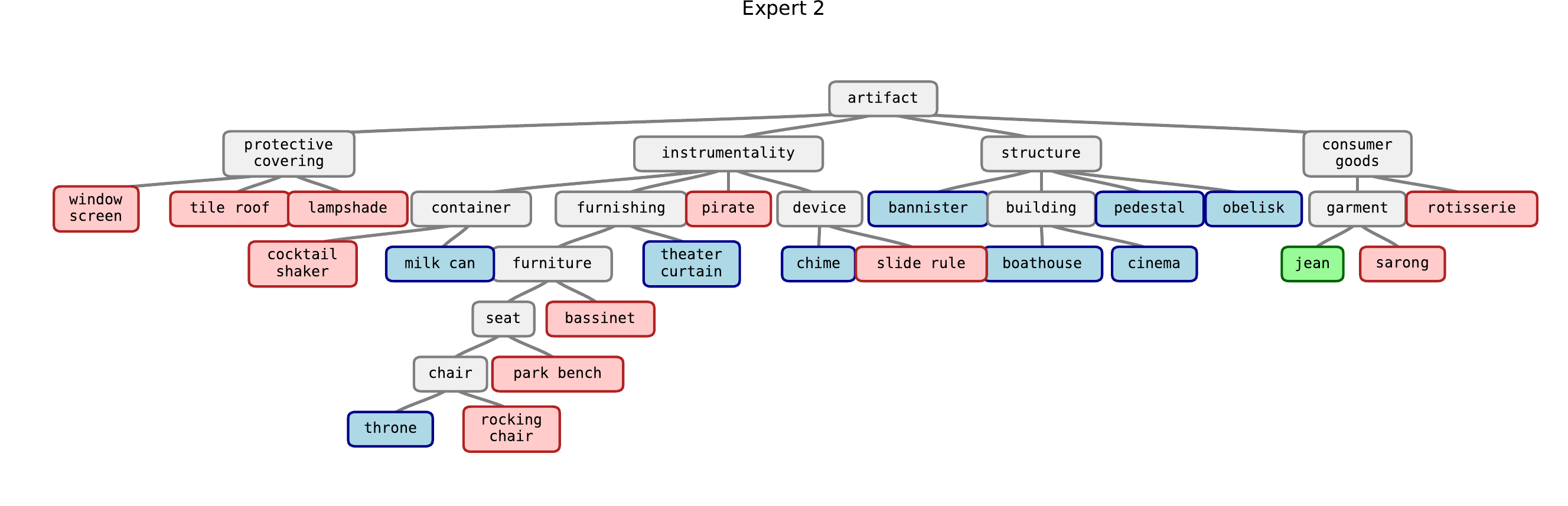}
    \end{subfigure}%
    \vspace{-1em}
\caption{Macro-class clustering for the first three expert edges of HgVT-Lt on the ImageNet-100 validation set. Nodes are shaded using gray for intermediate nodes, and colored for leaf nodes as follows: (blue) grouped to a single edge, (red) split over two edges, (orange) split over three edges, (green) split over four edges.}
\end{figure}

\begin{figure}[ht]
    \centering
    \begin{subfigure}[b]{.95\textwidth}
    \centering
    \includegraphics[width=\textwidth]{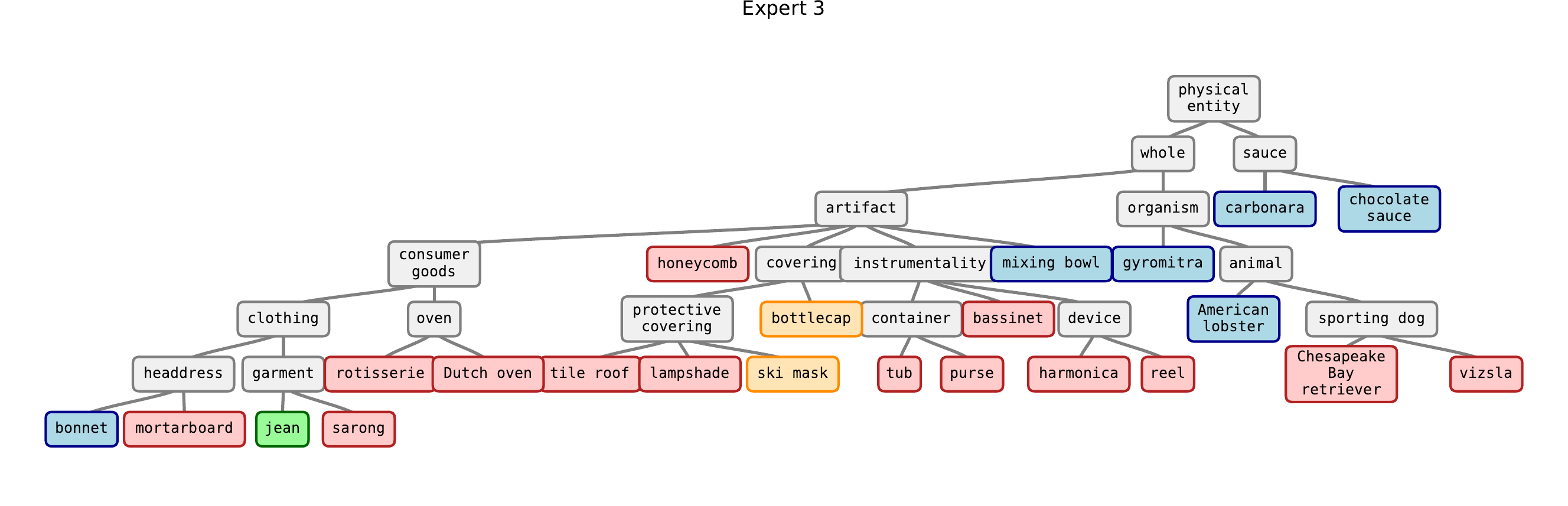}
    \end{subfigure}%
    \vfill
    \begin{subfigure}[b]{.95\textwidth}
    \centering
    \includegraphics[width=\textwidth]{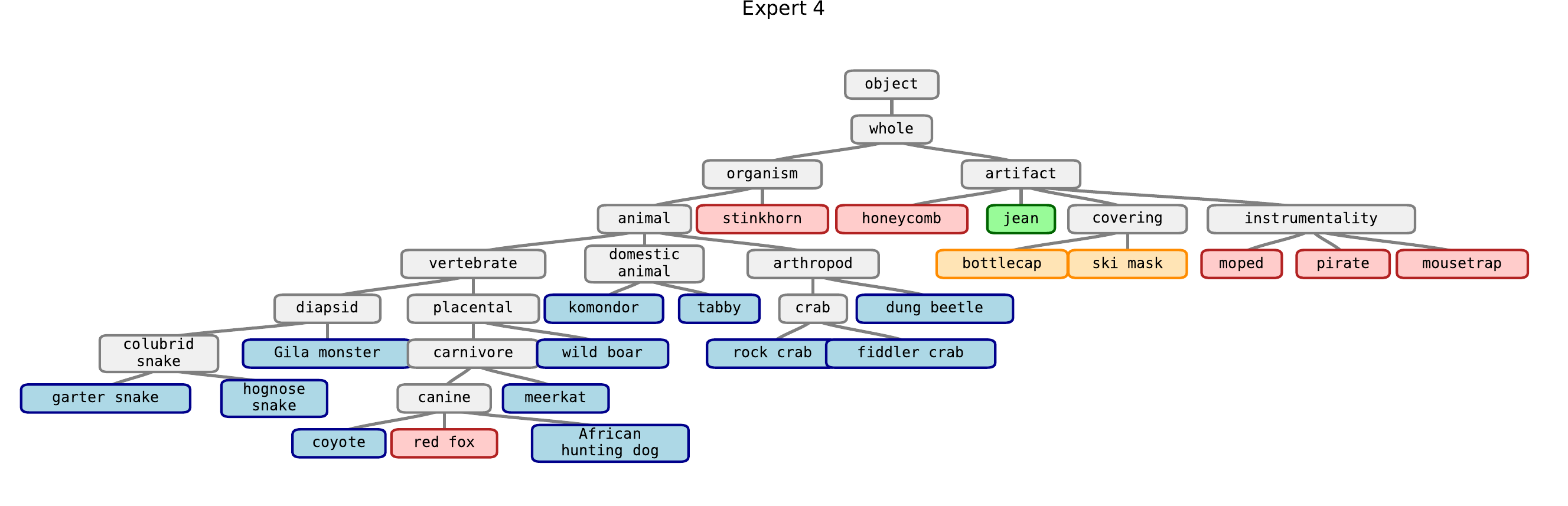}
    \end{subfigure}%
    \vfill
    \begin{subfigure}[b]{.95\textwidth}
    \centering
    \includegraphics[width=\textwidth]{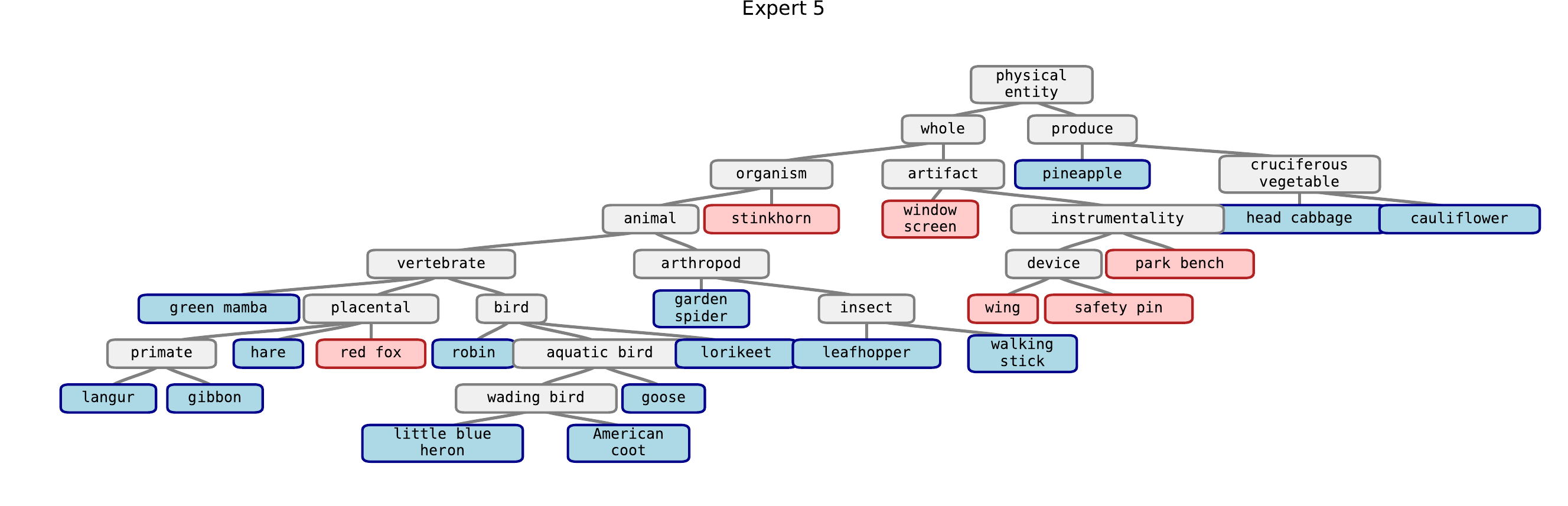}
    \end{subfigure}%

\caption{Macro-class clustering for the second three expert edges of HgVT-Lt on the ImageNet-100 validation set. Nodes are shaded using gray for intermediate nodes, and colored for leaf nodes as follows: (blue) grouped to a single edge, (red) split over two edges, (orange) split over three edges, (green) split over four edges.}
\end{figure}

\FloatBarrier

\newpage
\subsection{HgVT-S on ImageNet-1k}

\begin{figure}[ht]
    \centering
    \includegraphics[width=\textwidth]{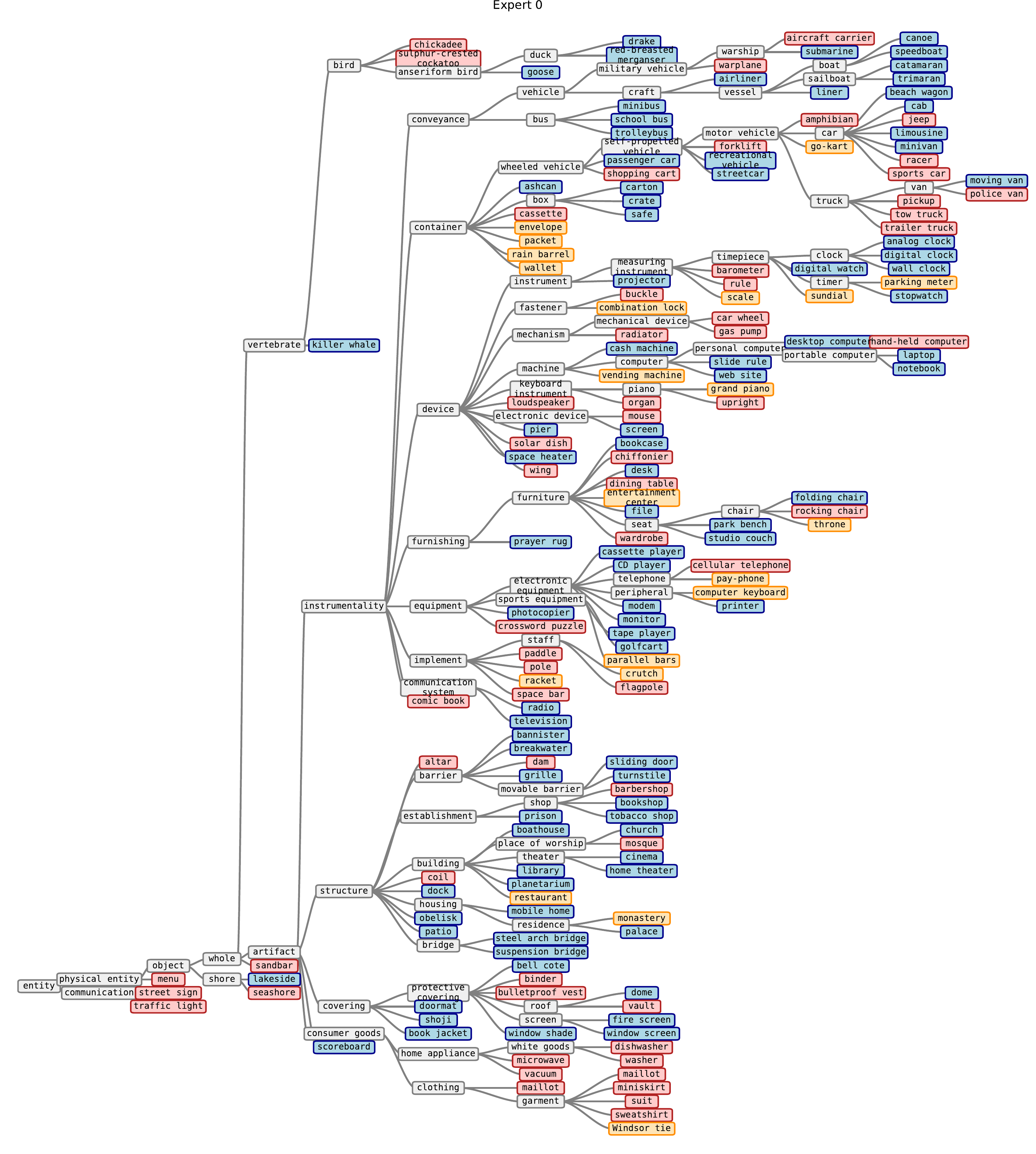}

\caption{Macro-class clustering for expert $0/8$ of HgVT-S on the ImageNet-1k validation set. Nodes are shaded using gray for intermediate nodes, and colored for leaf nodes as follows: (blue) grouped to a single edge, (red) split over two edges, (orange) split over three edges.}
\end{figure}   

\begin{figure}[ht]
    \centering
    \includegraphics[width=\textwidth]{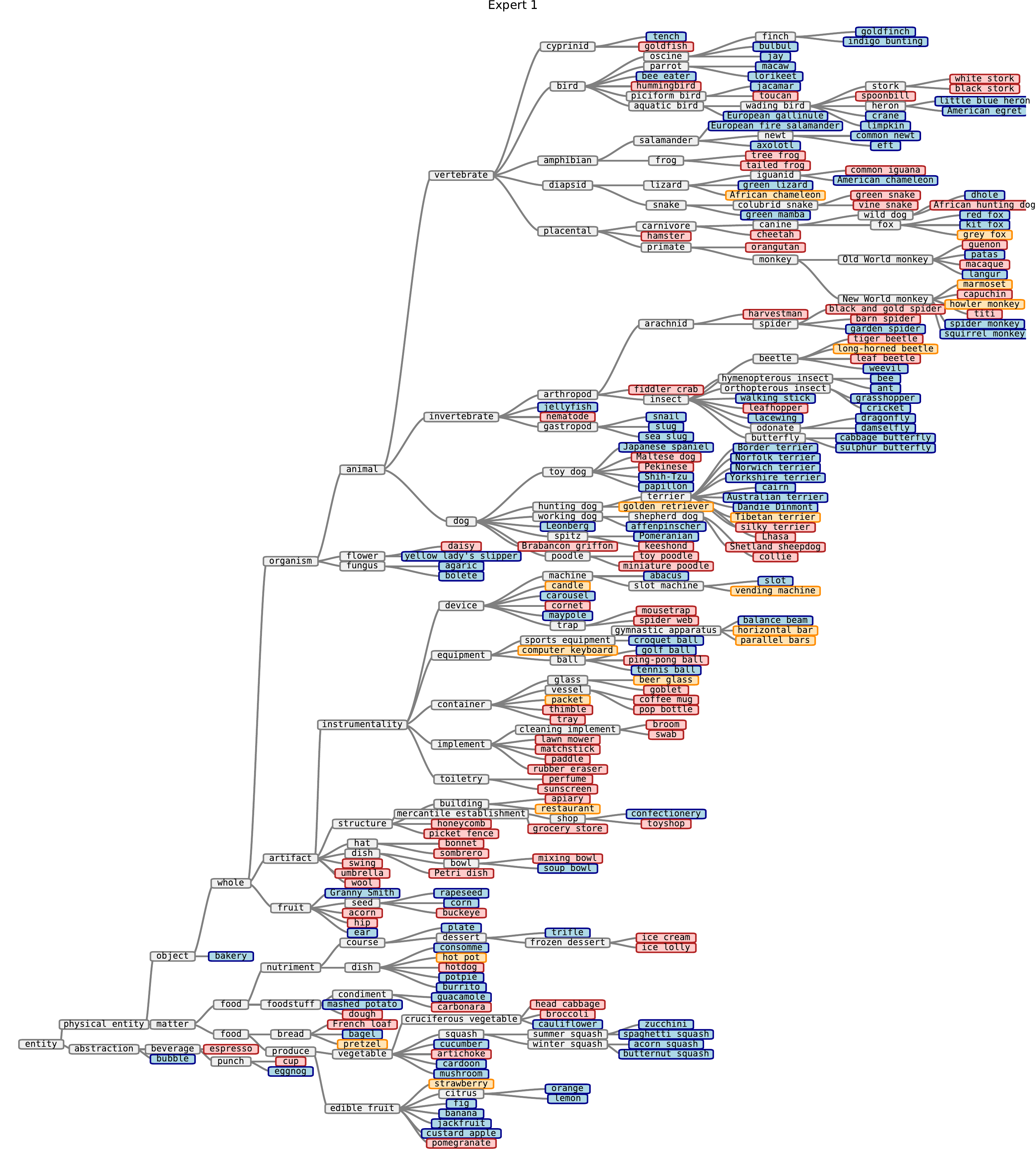}

\caption{Macro-class clustering for expert $1/8$ of HgVT-S on the ImageNet-1k validation set. Nodes are shaded using gray for intermediate nodes, and colored for leaf nodes as follows: (blue) grouped to a single edge, (red) split over two edges, (orange) split over three edges.}
\end{figure}   

\begin{figure}[ht]
    \centering
    \includegraphics[width=\textwidth]{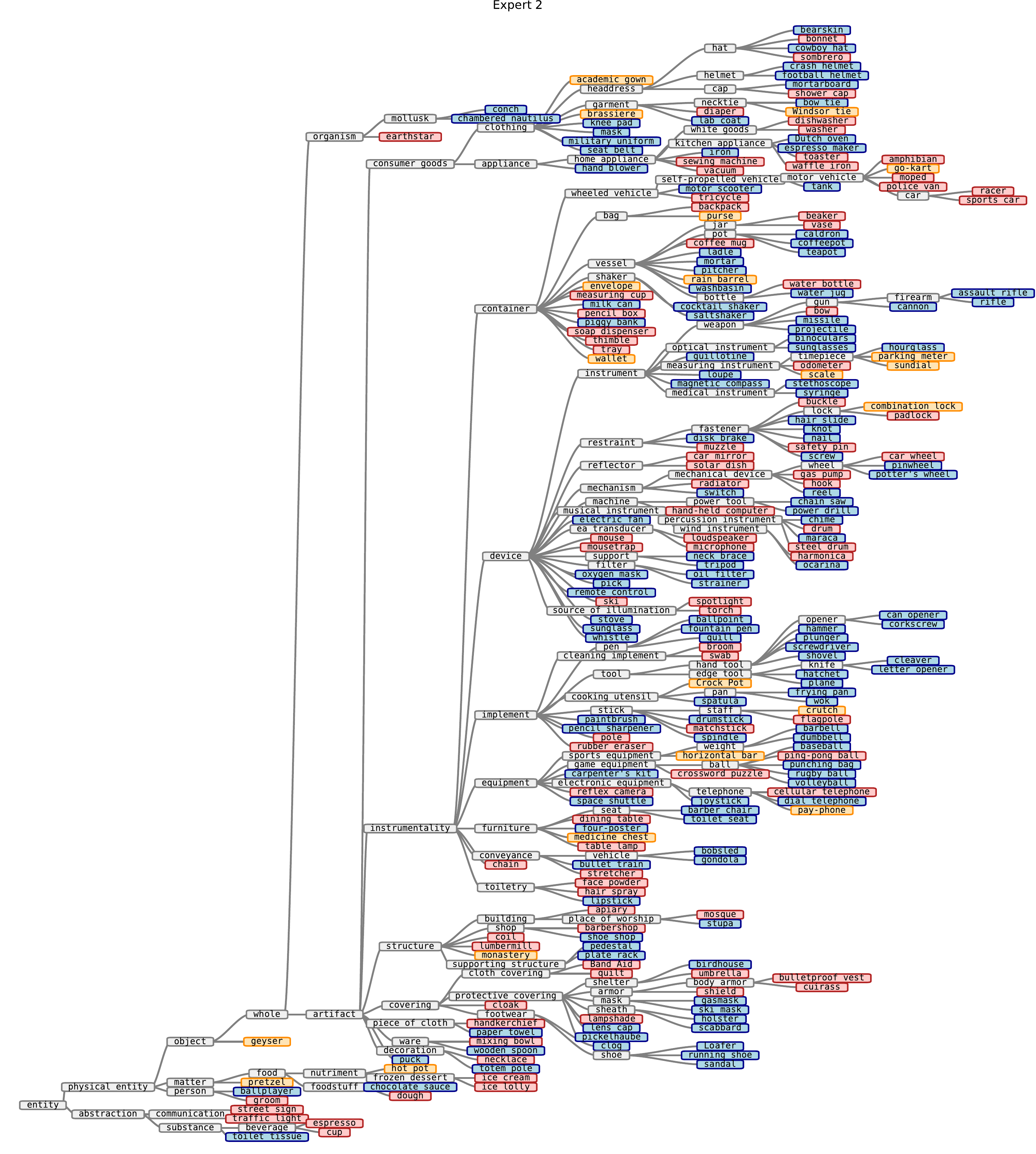}

\caption{Macro-class clustering for expert $2/8$ of HgVT-S on the ImageNet-1k validation set. Nodes are shaded using gray for intermediate nodes, and colored for leaf nodes as follows: (blue) grouped to a single edge, (red) split over two edges, (orange) split over three edges.}
\end{figure}   

\begin{figure}[ht]
    \centering
    \includegraphics[width=\textwidth]{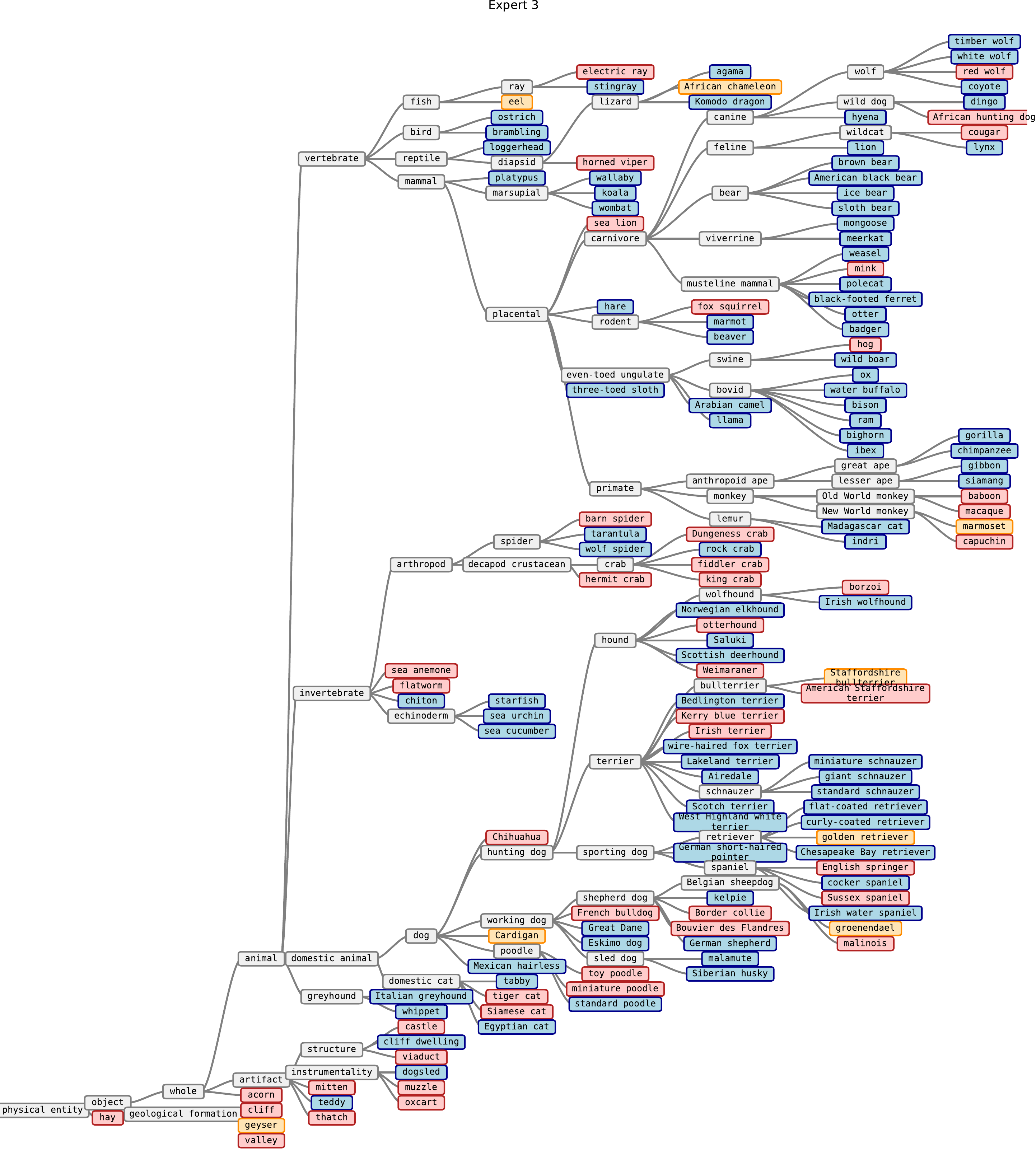}

\caption{Macro-class clustering for expert $3/8$ of HgVT-S on the ImageNet-1k validation set. Nodes are shaded using gray for intermediate nodes, and colored for leaf nodes as follows: (blue) grouped to a single edge, (red) split over two edges, (orange) split over three edges.}
\end{figure}   

\begin{figure}[ht]
    \centering
    \includegraphics[width=\textwidth]{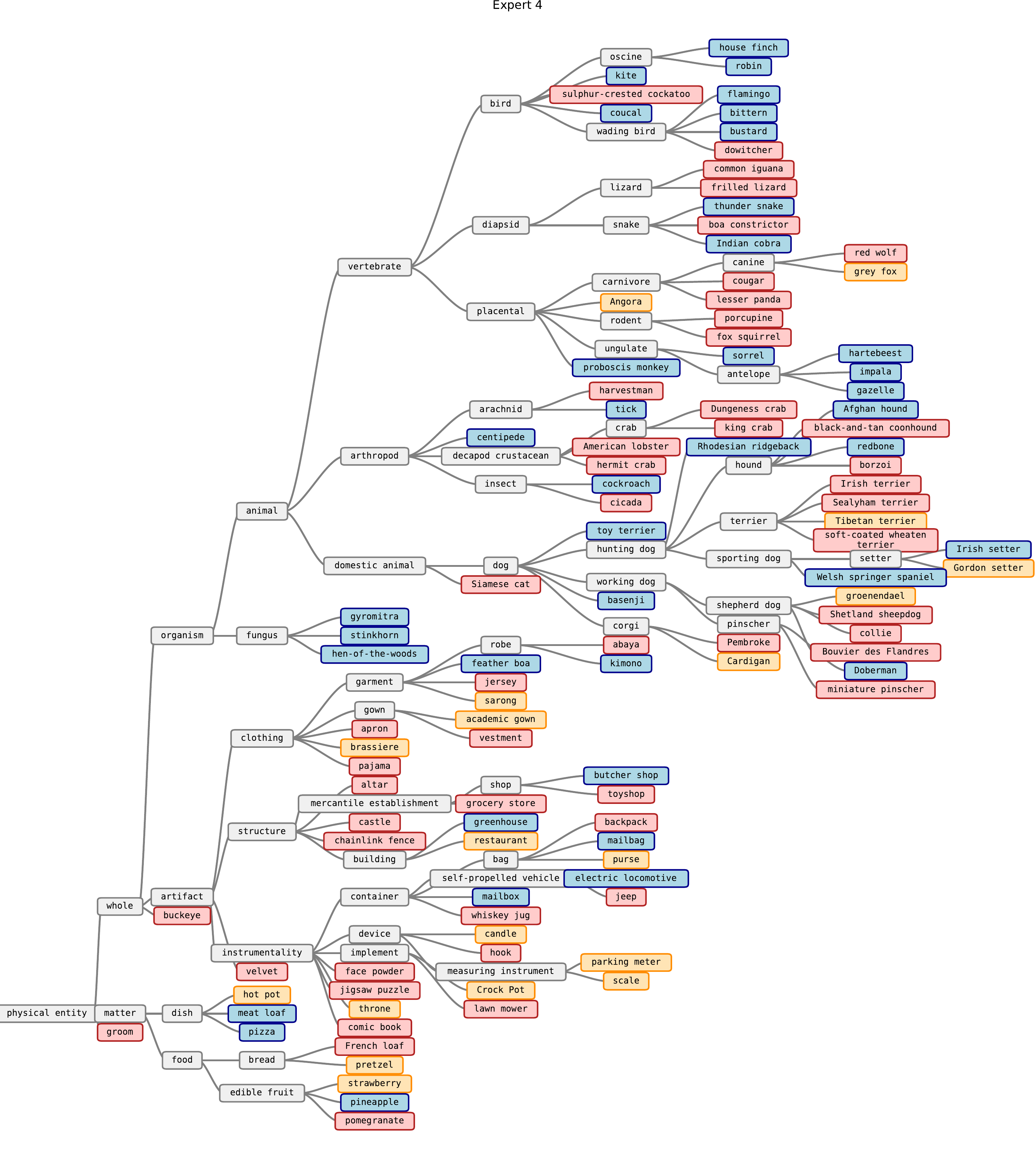}

\caption{Macro-class clustering for expert $4/8$ of HgVT-S on the ImageNet-1k validation set. Nodes are shaded using gray for intermediate nodes, and colored for leaf nodes as follows: (blue) grouped to a single edge, (red) split over two edges, (orange) split over three edges.}
\end{figure}   

\begin{figure}[ht]
    \centering
    \includegraphics[width=\textwidth]{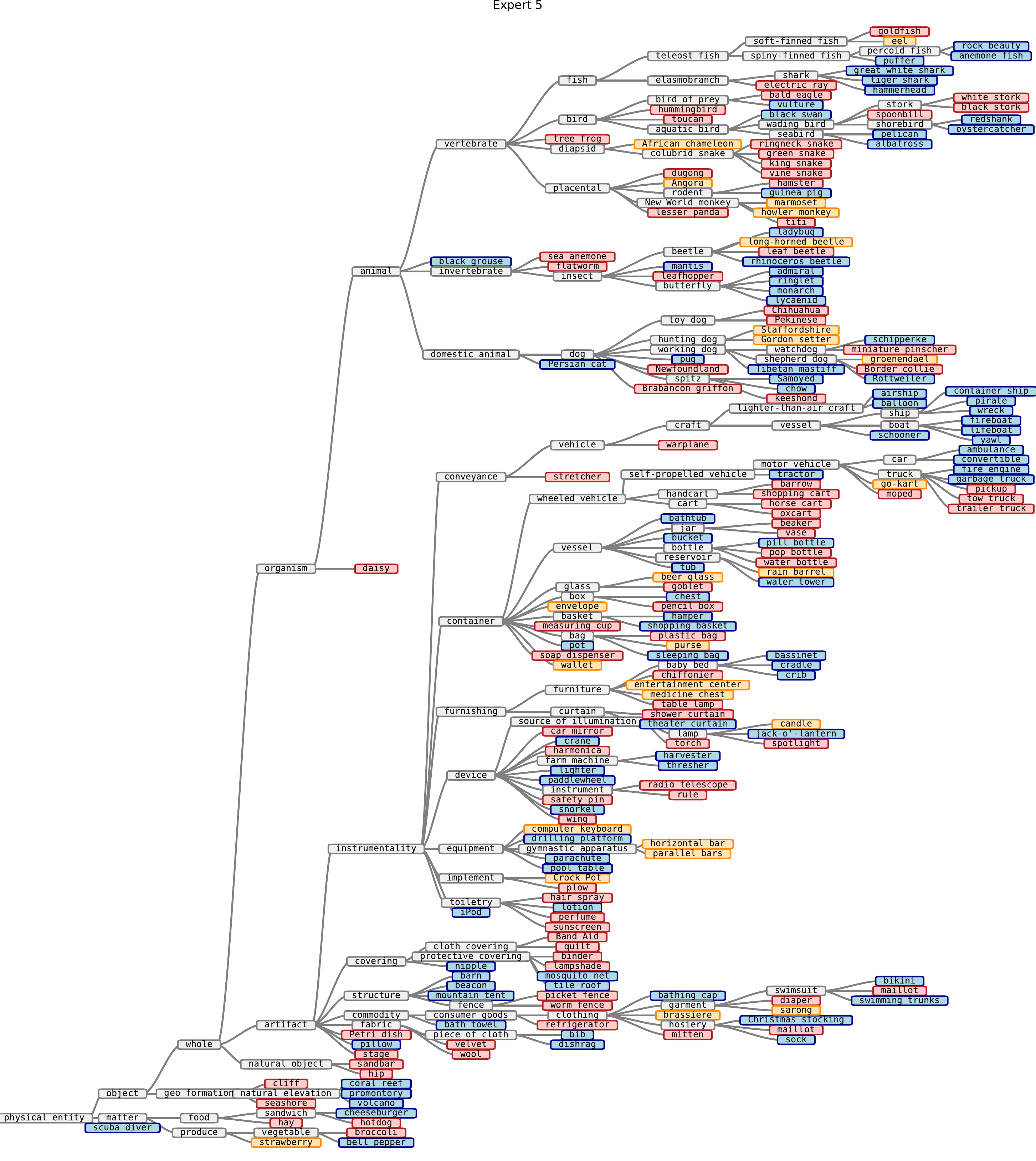}

\caption{Macro-class clustering for expert $5/8$ of HgVT-S on the ImageNet-1k validation set. Nodes are shaded using gray for intermediate nodes, and colored for leaf nodes as follows: (blue) grouped to a single edge, (red) split over two edges, (orange) split over three edges.}
\end{figure}

\begin{figure}[ht]
    \centering
    \includegraphics[width=\textwidth]{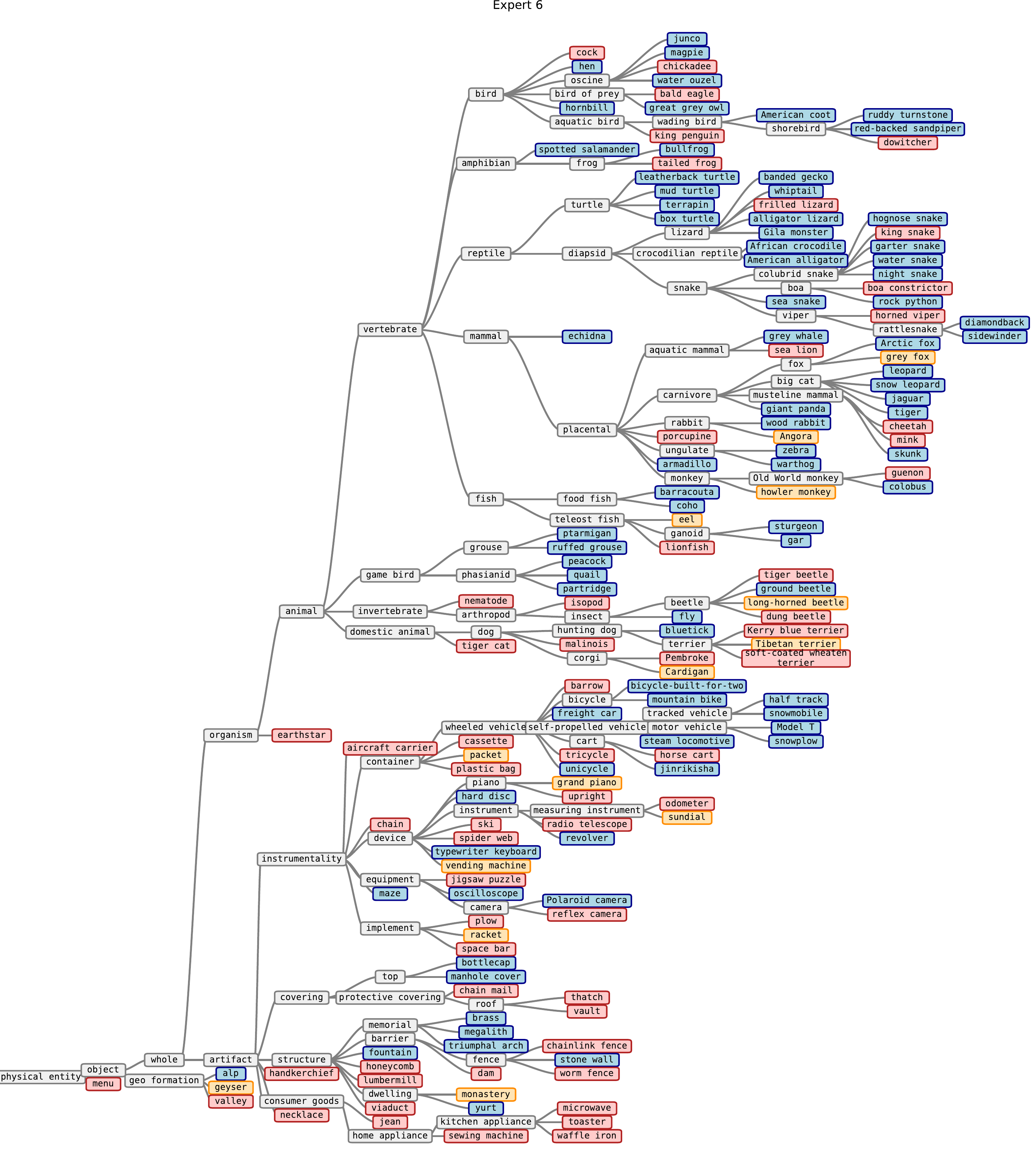}

\caption{Macro-class clustering for expert $6/8$ of HgVT-S on the ImageNet-1k validation set. Nodes are shaded using gray for intermediate nodes, and colored for leaf nodes as follows: (blue) grouped to a single edge, (red) split over two edges, (orange) split over three edges.}
\end{figure}

\begin{figure}[ht]
    \centering
    \includegraphics[width=\textwidth]{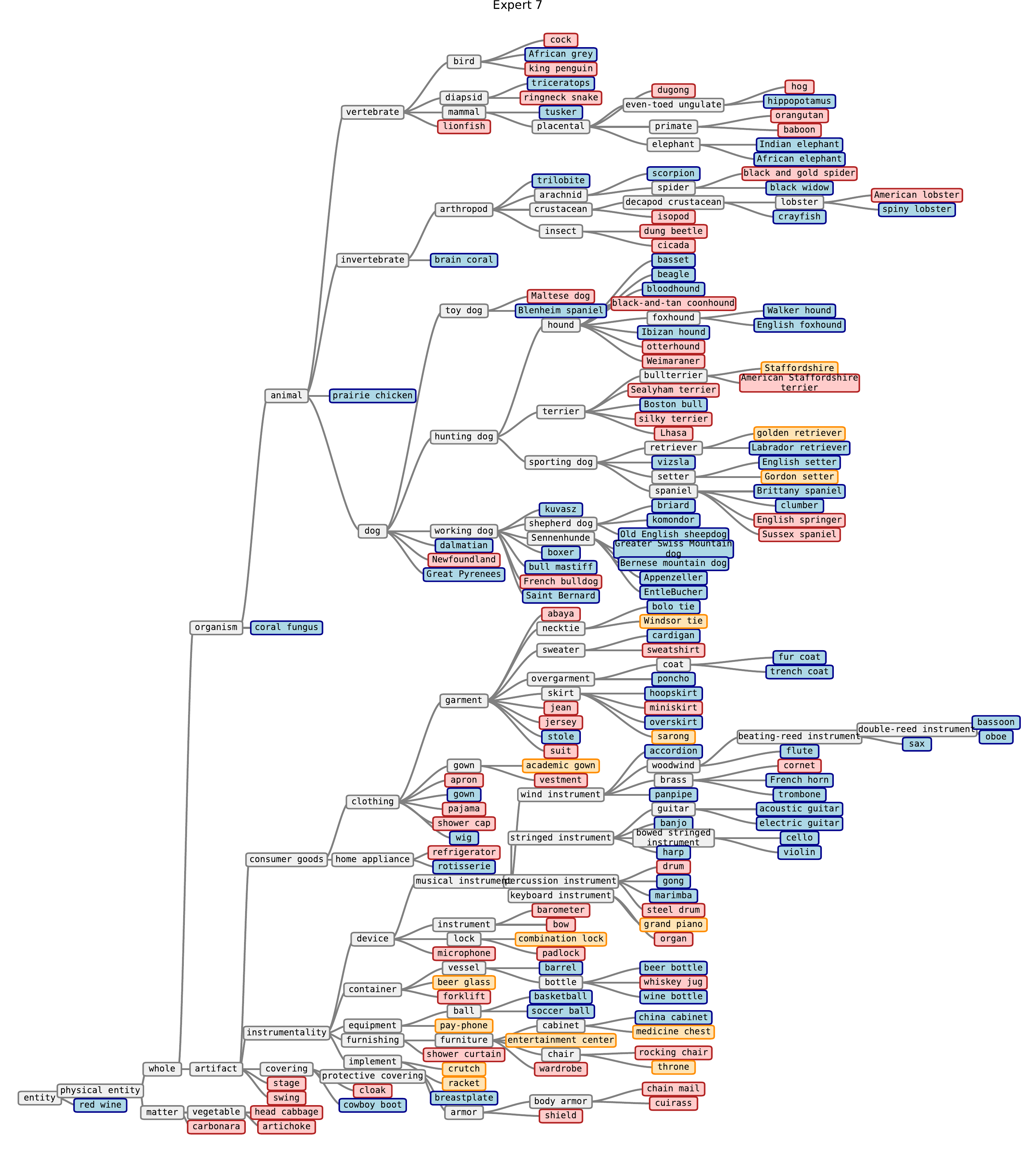}

\caption{Macro-class clustering for expert $7/8$ of HgVT-S on the ImageNet-1k validation set. Nodes are shaded using gray for intermediate nodes, and colored for leaf nodes as follows: (blue) grouped to a single edge, (red) split over two edges, (orange) split over three edges.}
\end{figure}   

\FloatBarrier

\end{document}